\pdfoutput=1
\documentclass[11pt]{article}
\usepackage[preprint]{acl}
\usepackage{xurl}

\usepackage{times}
\usepackage{latexsym}
\usepackage{amsmath}
\usepackage{booktabs}
\usepackage{siunitx}
\usepackage{multirow}
\usepackage{tcolorbox}
\usepackage{enumitem}
\usepackage{adjustbox}
\usepackage[T1]{fontenc}

\usepackage[utf8]{inputenc}

\usepackage{microtype}

\usepackage{inconsolata}

\usepackage{graphicx}

\usepackage{makecell}
\usepackage{wasysym}
\usepackage[dvipsnames]{xcolor}
\usepackage{longtable}
\usepackage{placeins}
\usepackage[normalem]{ulem}
\usepackage{pifont}
\newcommand{\cmark}{\textcolor{ForestGreen}{\ding{51}}}
\newcommand{\xmark}{\textcolor{BrickRed}{\ding{55}}}
\newcommand{\pmark}{\LEFTcircle}
\title{Human Psychometric Questionnaires Mischaracterize LLM Behavior}

\author{
  Woojung Song$^{1*}$ \quad Dongmin Choi$^{1*}$ \quad Yoonah Park$^{1}$ \\  
  \bfseries Jongwook Han$^{1}$ \quad Eun-Ju Lee$^{2}$ \quad Yohan Jo$^{1\dag}$ \\
  $^{1}$Graduate School of Data Science, Seoul National University \\
  $^{2}$Department of Communication, Interdisciplinary Program in Artificial Intelligence, \\
  Seoul National University \\
  {\texttt{\{opusdeisong, chrisandjj, eunju0204, yohan.jo\}@snu.ac.kr}}
}

\begin{document}
\maketitle
\def\thefootnote{\fnsymbol{footnote}}
\footnotetext[1]{Equal contribution.}
\footnotetext[2]{Corresponding author.}
\def\thefootnote{\arabic{footnote}}
\begin{abstract}
We examine whether human psychometric questionnaires can serve as reliable tools for characterizing and predicting LLM behavior in everyday user interactions. We analyze eight open-source LLMs by comparing their value and personality profiles derived from two different methods: Likert self-reports on established questionnaires (PVQ-40/21 and BFI-44/10) and generation probabilities over value-laden responses to everyday user queries. 
The two profiles diverge substantially. Within-construct item consistency, often cited as evidence of stable LLM dispositions, disappears in generation probabilities. We find that established questionnaire items contain explicit lexical cues that allow models to recognize the target construct and respond in alignment-consistent, socially desirable ways, whereas realistic user queries contain far less recognizable cues.
In addition, demographic persona prompts shift models' responses to human questionnaires in ways consistent with real human patterns, but no such shifts appear in the generation setting, highlighting that human questionnaires overestimate LLMs' ability to faithfully reproduce expected psychological traits when role-playing demographic personas.
Overall, our study indicates that questionnaire scores alone should not be treated as evidence of LLMs' response tendencies in realistic user interactions, and supports generation-probability profiling with ecologically valid items as a complementary behavioral measure. Our code is available at \url{https://github.com/holi-lab/Human-Psychometric-Questionnaires-Mischaracterize-LLM-Behavior}.
\end{abstract}

\section{Introduction}

As large language models (LLMs) are increasingly adopted for high-risk and high-stakes tasks, such as emotional support \cite{kang-etal-2024-large}, ethical advice \cite{rao-etal-2023-ethical}, and chatbots for children \cite{rath-etal-2025-llm}, characterizing the values and traits they express is imperative for behavioral predictability and safety. For humans, such characterization has long been conducted through psychometric questionnaires, such as the Portrait Values Questionnaire \cite{schwartz2012overview}, the BFI \cite{john1991big}, and IPIP-NEO \cite{goldberg1999ipip}. The resulting psychological profiles can predict human behavior across diverse situations \cite{Bardi2003ValuesAB}. Naturally, machine learning researchers have attempted to apply these tools to LLMs as well, based on the same promise of behavioral predictability \cite{Rozen2025Dollmshave, hadar2024assessing, miotto-etal-2022-gpt, huang2023humanity, Bodro_a_2024}.

Our study focuses on examining the utility of human psychometric questionnaires as tools for reliably predicting and characterizing LLM behavior. To this end, two key aspects should be ensured for ecological validity: (1) LLM behavior should be operationalized as realistic LLM generations, which today most often take the form of responses to user queries, and (2) the target LLM behavior should be generative rather than reflective (e.g., self-reports). 
Prior studies, on the other hand, even those reporting divergences between psychometric questionnaires and LLM behavior, typically consider LLM behaviors in tasks distant from everyday user interactions, rely on self-reports for LLM behavior, or compare different sets of psychological constructs between psychometric questionnaires and LLM behavior \cite{ai2024selfknowledgeactionconsistentnot, shen-etal-2025-mind, han2025personality, rottger-etal-2024-political}. 
Moreover, we examine two important questions: \emph{why} human questionnaires yield coherent, internally consistent LLM profiles, and whether persona-induced shifts on these instruments carry over to generation behavior. Addressing these questions has important scientific merit for better understanding LLMs' largely black-box decision-making.

\begin{table*}[t]
\centering
\footnotesize
\setlength{\tabcolsep}{5pt}
\renewcommand{\arraystretch}{1.05}
\setlength{\aboverulesep}{0pt}
\setlength{\belowrulesep}{0pt}
\begin{adjustbox}{max width=\textwidth}
\begin{tabular}{@{}llp{3.4cm}cc@{}}
\toprule
\textbf{Work} & \textbf{Construct} & \textbf{Behavior probe}
& \makecell{\textbf{Validated}\\\textbf{realistic items}}
& \makecell{\textbf{Why same-construct items}\\\textbf{receive similar LLM scores}} \\
\midrule
\cite{han2025personality}          & Personality & Behavioral tasks     & \xmark & \pmark \\
\cite{ai2024selfknowledgeactionconsistentnot} & Personality & Scenario-action questions & \xmark & \xmark \\
\cite{zou2025can}                  & Personality & Human-perceived traits          & \xmark & \xmark \\
\cite{jung-etal-2026-psychometric} & Sexism / Racism / Morality & Downstream tasks  & \xmark & \xmark \\
\cite{taubenfeld2026evaluating}    & Dispositions & Generated SJTs                 & \pmark & \xmark \\
\cite{shen-etal-2025-mind}         & Values       & Value-informed actions   & \xmark & \xmark \\
\cite{rottger-etal-2024-political} & Political values & Open-ended opinions & \xmark & \xmark \\
\midrule
\textbf{Ours} & \textbf{Values, Personality} & \textbf{Log-prob over VP pairs} & \cmark & \cmark \\
\bottomrule
\end{tabular}
\end{adjustbox}
\caption{Comparison with prior work on the gap between questionnaire responses and behavior in LLMs. \LEFTcircle{} indicates partial or implicit support.}
\label{tab:related_work}
\end{table*}

We address these gaps by comparing two psychological profiling methods: \textit{established questionnaire} scores and \textit{generation-probability} scores. Established questionnaire scores come from LLMs' Likert responses to PVQ-40, PVQ-21, BFI-44, and BFI-10, well-established questionnaires for measuring Schwartz's human values and the Big Five personality traits. 
 
To obtain generation-probability scores, directly asking LLMs open-ended queries and annotating their responses is highly prone to errors and can severely impair the validity of the resulting profiles \cite{han-etal-2025-value}. Therefore, we instead use the Value Portrait dataset, which consists of real-world user queries and plausible LLM responses; each response is associated with Schwartz values and Big Five traits through psychometric validation. We measure the log-probabilities of LLMs generating each response, from which their profiles are derived. This balances the generative nature of the target behavior with psychometric validity.

We organize the investigation around four research questions and experiment with eight open-source LLMs. \textbf{RQ1:} Do established questionnaires and actual generation-probability produce different LLM profiles? Construct-ranking agreement between them is low. \textbf{RQ2:} Does intra-construct response consistency hold for generation-probability profiles?
The intra-construct item consistency, often read as evidence of ``stable LLM dispositions''  \cite{lee-etal-2025-llms, serapiogarcía2025personalitytraitslargelanguage}, does not appear in generation probabilities. \textbf{RQ3:} Can LLMs recognize the target construct from item text? We examine whether item textual transparency contributes to this divergence. Established questionnaires carry explicit cues about the construct they measure, letting models recognize the target and respond in alignment-consistent, socially desirable ways. In contrast, queries and responses in Value Portrait have far less recognizable cues. 
To test whether these cues contribute to the consistency observed in questionnaire responses, we rewrite the established items to make their target constructs less obvious. LLM responses are then less consistent across items intended to measure the same construct.
\textbf{RQ4:} 
Do persona-induced shifts in questionnaire profiles carry over to generation behavior?
The corresponding human-aligned shifts do not emerge in generation-probability profiles. 
Taken together, these findings show that profiles derived from human psychometric questionnaires do not reliably characterize LLM behavior in realistic user interactions. They should therefore be interpreted with caution and complemented by generation-probability profiling based on validated, realistic items.

\section{Related Works}

\subsection{Psychometric Assessment of LLMs}
Prior work has applied established psychometric questionnaires---BFI variants \cite{john1991big, soto2017bfi2}, IPIP-NEO \cite{goldberg1999ipip}, and the PVQ family \cite{schwartz2012overview}---to assess values and personality in LLMs \citep{miotto-etal-2022-gpt, serapiogarcía2025personalitytraitslargelanguage, kovavc2024stick}, reporting evidence of reliability, differentiation, or construct-related validity \citep{huang-etal-2024-reliability, jiang2023evaluating, lee-etal-2025-llms}.

\subsection{Psychometric Profiles vs.\ LLM Behavior}
A growing body of work probes LLM behavior beyond questionnaire responses to test the external validity of questionnaire-derived profiles. As Table~\ref{tab:related_work} shows, prior studies capture behavior in diverse ways: scenario-action items \cite{ai2024selfknowledgeactionconsistentnot}, generated situational judgment tests \cite{taubenfeld2026evaluating}, behavioral tasks \cite{han2025personality}, value-informed action tasks \cite{shen-etal-2025-mind}, open-ended opinion generation \cite{rottger-etal-2024-political}, downstream tasks \cite{jung-etal-2026-psychometric}, and user-perceived traits \cite{zou2025can}.
Across these studies, behavioral probes often rely on items whose links to the target constructs are not independently validated, or on tasks that remain structurally close to questionnaires rather than realistic human-LLM interactions.
Our work complements these studies by using generation probabilities as a behavioral proxy over psychometrically validated items, while preserving the construct space of the target questionnaires. We further investigate item transparency as a weakness of human questionnaires (§\ref{sec:rq3}).

\subsection{Response Probability as a Measure of Model Behavior}
LLM self-reports are highly sensitive to prompt design, where minor instruction changes substantially shift questionnaire profiles~\citep{gupta2024self_assessment_tests, shu2024you_dont_need}. In contrast, probability-based measurements are more robust to prompt and minor contextual variations~\citep{kauf-etal-2024-log}. These methods also directly read out the model’s internal scoring of candidate texts from the generation distribution, avoiding reliance on accurate meta-judgments about its own dispositions~\citep{hu-levy-2023-prompting, gao-kreiss-2025-measuring}. We derive profiles from the log-probabilities of candidate responses to better capture models’ generation behavior.

\section{Profiling Methods and Models}
\label{sec:datasets_models}

We compare two methods for profiling LLMs' psychological constructs: \textit{established questionnaire} scores, and \textit{generation-probability} scores from the log-probabilities a model assigns to construct-annotated responses in realistic scenarios. Both approaches target the Big Five personality traits and ten Schwartz basic values (Appendix~\ref{sec:app_construct_definitions}).

\subsection{Established Questionnaire Profiling}
\label{sec:established}

We administer four established psychometric questionnaires: PVQ-40~\cite{schwartz2001extending} and PVQ-21~\cite{schwartz2003ess} for Schwartz values, and BFI-44~\cite{john1991big} and BFI-10~\cite{BFI-10} for the Big Five personality traits.
For each questionnaire, we present the original item wording with pronouns rewritten in gender-neutral form (\textit{they}/\textit{them}) and collect Likert-scale responses (1--6 for PVQ, 1--5 for BFI). Because LLMs are known to be sensitive to option order~\cite{huang-etal-2024-reliability, pmlr-v139-zhao21c}, we employ two prompt variants that present the Likert scale in opposite directions (high-to-low and low-to-high) and average the results. The final construct score is obtained by averaging across all items within a construct over both prompt variants. Full prompt templates are provided in Appendix~\ref{sec:questionnaire_prompts}.

\subsection{Generation Probability Profiling}
\label{sec:gen_prob}
To derive profiles that reflect LLM behavior in everyday use, we aim to satisfy two conditions. First, LLM behavior should be defined as the types of LLM generations commonly observed in everyday user interactions; for this, we use plausible LLM responses to real-world user queries collected from diverse corpora (detailed below).
Second, the target behavior should be generative rather than reflective (e.g., self-reports).
While annotating LLMs' open-ended responses with constructs seems promising, such annotations have been found to be critically error-prone and can severely invalidate the resulting profiles \cite{han-etal-2025-value}.
We therefore derive profiles from the log-probabilities that a model assigns to a predefined set of construct-annotated responses in realistic scenarios.
\paragraph{Dataset: Value Portrait.}
We use the Value Portrait (VP) dataset~\cite{han-etal-2025-value}, which provides psychometrically validated scenario-response pairs covering the same constructs as the established questionnaires. VP comprises 520 query-response pairs derived from 104 real-world queries drawn from human-LLM conversations (ShareGPT, LMSYS~\cite{lmsys}) and human-human advisory archives (Reddit~\cite{scruples}, Dear Abby), spanning everyday requests (opinion questions, brainstorming, open-ended discussion) and value-laden interpersonal dilemmas. Each query is paired with five candidate responses annotated for Schwartz values and Big Five traits, validated through a correlation study with 681 human raters; items reliably associated with a construct 
$\rho \geq .30$) 
are tagged accordingly, yielding 286 value-tagged and 228 trait-tagged (item, construct) pairs. Construction and validation details are provided in Appendix~\ref{sec:app_vp_dataset}.

\paragraph{Scoring.}
For each scenario, we compute the log-probability that the model assigns to each response.
For each construct $c$, let $S_c$ be the set of scenarios containing at least one response tagged with $c$ (i.e., whose human-validated correlation satisfies 
$\rho \geq .30$), and let $R_{c,s} \subseteq R_s$ be the tagged responses within scenario $s$. We score $c$ as the mean across scenarios of the within-scenario mean log-probability of its tagged responses:
\begin{equation}
    \text{score}(c) = \frac{1}{|S_c|} \sum_{s \in S_c} \frac{1}{|R_{c,s}|} \sum_{r \in R_{c,s}} \log P(r \mid s)
\label{eq:gen_prob_score}
\end{equation}
where $\log P(r \mid s)$ is the \emph{sum} of token log-probabilities (no length normalization; see Appendix~\ref{sec:app_inference} for the exact slicing used). Appendix~\ref{sec:app_rq1_lengthnorm} reports the corresponding per-token sensitivity analysis. This two-step (scenario-then-construct) macro-average gives every scenario equal weight regardless of how many of its candidate responses are tagged for $c$, mitigating the influence of scenarios with multiple tags. Appendix~\ref{sec:app_rq1_scoring} details the choice and reports the alternative flat micro-average. Repeating this for all constructs yields a 10-dimensional value profile and a 5-dimensional personality profile per model, the \textit{generation-probability profile}.

To verify that VP candidate responses fall within a reasonable region of each model's generation distribution, we sampled 10 responses per scenario at temperature\,=\,1.0 and ranked all 15 responses (10 sampled + 5 VP) by log-probability. The highest-ranked VP candidate had a median rank of 4 (mean\,=\,5.7; Appendix~\ref{app:sampling_validity}).
This result reduces the concern that the probability-based method misrepresents LLM profiles merely because the candidate responses lie outside the model's generation distribution.

\paragraph{Prompt Design.}
To probe the generation distribution without Likert-scale constraints, we present each VP scenario as an open-ended prompt. We use two templates based on the scenario source: conversational scenarios (ShareGPT and LMSYS) are framed as user messages directed at the model, whereas advisory scenarios (Dear Abby and Reddit) are framed with a title and situational description. Both templates ask the model to respond naturally; full prompts are provided in Appendix~\ref{sec:VP_prompts}.

\paragraph{Method scope and justification.}
Our method should be interpreted as a behavioral measure based on fixed, construct-validated candidate responses, rather than as a measure of unconstrained real-world behavior.
To address the potential concern of limited response diversity, we include multiple plausible candidate responses that represent different constructs.
The resulting profile targets the relative ordering of constructs: Equation~\ref{eq:gen_prob_score} aggregates tagged responses across scenarios, averaging item-level variation while retaining inter-construct contrasts.
More importantly, the psychological constructs associated with the candidate responses have been psychometrically validated, which is challenging in unconstrained generation settings.

\begin{table*}[!t]
\centering
\small
\renewcommand{\arraystretch}{0.97}
\setlength{\tabcolsep}{4pt}
\setlength{\aboverulesep}{0pt}
\setlength{\belowrulesep}{0pt}
\sisetup{
  table-format=-1.2,
  table-number-alignment=center,
  detect-weight,
  mode=text
}
\begin{tabular}{@{}c l l S S S S S S S S S[table-format=1.2]@{}}
\toprule
& & & \multicolumn{2}{c}{\textbf{Gemma3}} & \multicolumn{2}{c}{\textbf{GPT-OSS}} & \multicolumn{2}{c}{\textbf{Qwen 2.5}} & \multicolumn{2}{c}{\textbf{Qwen 3}} & \\
\cmidrule(lr){4-5} \cmidrule(lr){6-7} \cmidrule(lr){8-9} \cmidrule(lr){10-11}
& & & {27B} & {4B} & {120B} & {20B} & {72B} & {7B} & {235B} & {30B} & {Avg.} \\
\midrule
\multirow{6}{*}{\rotatebox[origin=c]{90}{\textbf{Spearman $\rho$}}}
& \textbf{Val.} & PVQ-40 $\leftrightarrow$ PVQ-21 & \bfseries 0.94 & \bfseries 0.88 & \bfseries 0.81 & \bfseries 0.42 & \bfseries 0.90 & \bfseries 0.46 & \bfseries 0.67 & \bfseries 0.85 & \bfseries 0.74 \\
& & Gen $\leftrightarrow$ PVQ-40 & 0.26 & 0.10 & 0.27 & 0.26 & 0.72 & 0.45 & 0.52 & -0.08 & 0.31 \\
& & Gen $\leftrightarrow$ PVQ-21 & 0.34 & 0.24 & 0.36 & -0.11 & 0.64 & 0.40 & 0.46 & -0.10 & 0.28 \\
\cmidrule(l){2-12}
& \textbf{Trait} & BFI44 $\leftrightarrow$ BFI10 & \bfseries 0.70 & \bfseries 1.00 & \bfseries 0.90 & \bfseries 0.67 & \bfseries 0.90 & \bfseries 0.92 & 0.21 & \bfseries 0.89 & \bfseries 0.77 \\
& & Gen $\leftrightarrow$ BFI44 & -0.50 & 0.30 & 0.20 & 0.60 & 0.70 & 0.41 & \bfseries 0.90 & -0.50 & 0.26 \\
& & Gen $\leftrightarrow$ BFI10 & -0.80 & 0.30 & 0.10 & 0.36 & \bfseries 0.90 & 0.67 & 0.05 & -0.67 & 0.11 \\
\midrule
\multirow{6}{*}{\rotatebox[origin=c]{90}{\textbf{NDCG}}}
& \textbf{Val.} & PVQ-40 $\leftrightarrow$ PVQ-21 & \bfseries 0.81 & \bfseries 0.98 & 0.87 & 0.71 & \bfseries 1.00 & \bfseries 0.91 & \bfseries 0.99 & \bfseries 1.00 & \bfseries 0.91 \\
& & Gen $\leftrightarrow$ PVQ-40 & 0.66 & 0.96 & 0.84 & \bfseries 0.88 & 0.75 & 0.74 & 0.97 & 0.69 & 0.81 \\
& & Gen $\leftrightarrow$ PVQ-21 & 0.77 & 0.96 & \bfseries 0.97 & 0.64 & 0.75 & 0.66 & 0.97 & 0.68 & 0.80 \\
\cmidrule(l){2-12}
& \textbf{Trait} & BFI44 $\leftrightarrow$ BFI10 & \bfseries 0.78 & \bfseries 1.00 & \bfseries 0.87 & 0.74 & \bfseries 0.98 & \bfseries 1.00 & 0.75 & \bfseries 0.98 & \bfseries 0.89 \\
& & Gen $\leftrightarrow$ BFI44 & 0.62 & 0.72 & 0.68 & \bfseries 0.95 & 0.78 & 0.76 & \bfseries 0.98 & 0.61 & 0.76 \\
& & Gen $\leftrightarrow$ BFI10 & 0.57 & 0.72 & 0.66 & 0.68 & 0.87 & 0.76 & 0.65 & 0.57 & 0.69 \\
\bottomrule
\end{tabular}
\caption{Spearman $\rho$ and NDCG between self-report and generation-based construct profiles (\textbf{Gen.}). Bold values indicate the highest score per model within each group (Val.\ / Trait).}\label{tab:rq1_main}
\end{table*}

\subsection{Models}
\label{sec:models}
Since our method requires generation probabilities, we use 8 open-source models across 4 families, each with a smaller and larger variant: \textbf{Gemma3} (4B, 27B), \textbf{Qwen 2.5} (7B, 72B), \textbf{Qwen 3} (30B-A3B, 235B-A22B; both MoE), and \textbf{GPT-OSS} (20B, 120B). Configurations are in Appendix~\ref{sec:app_inference}.

\section{Research Questions}

\subsection{RQ1: Do established questionnaires and actual generation-probability produce different LLM profiles?}
\label{sec:rq1}

We compare construct-level profiles obtained from Likert-based self-reports (established questionnaires) against those derived from generation probabilities (VP) for each of the 8 models.

\paragraph{Comparison Metrics.}
\emph{Cross-method} agreement is assessed by comparing construct rankings using rank-based metrics. Spearman's $\rho$ measures the overall rank correlation between the construct rankings in the two profiles. NDCG (Normalized Discounted Cumulative Gain) complements this by assigning greater weight to the top-ranked constructs, testing whether the two methods agree on the most dominant constructs. We take the established-questionnaire profile as the reference ranking, assign relevance $\mathrm{rel}_i = n - \mathrm{rank}_i + 1$ to each construct, and compute DCG with exponential gain, normalized by the ideal DCG of the reference ranking. Formal definitions, tie handling, and significance testing are in Appendix~\ref{sec:app_rq1_metrics}. We also report \emph{within-method} references, computed between questionnaire pairs that target the same constructs: PVQ-40$\leftrightarrow$PVQ-21 for values and BFI-44$\leftrightarrow$BFI-10 for personality traits.

\paragraph{Self-report profiles agree with each other, but much less with generation-probability profiles.}
Table~\ref{tab:rq1_main} summarizes ranking agreement between the profiling methods. Within-method agreement is high: average Spearman $\rho$ is 0.74 for PVQ-40 vs.\ PVQ-21 and 0.77 for BFI-44 vs.\ BFI-10. Cross-method agreement is lower, with a paired sign-flip permutation test confirming the gap for Schwartz values ($p = 0.004$), Big Five ($p = 0.016$), and the pooled set ($p < 0.001$; Appendix~\ref{sec:app_rq1_stat}). Generation-vs-established $\rho$ drops to 0.31 (Gen$\leftrightarrow$PVQ-40) and 0.28 (Gen$\leftrightarrow$PVQ-21) for Schwartz values, roughly 40\% of the within-method reference, and is even lower for Big Five: 0.26 (Gen$\leftrightarrow$BFI44) and 0.11 (Gen$\leftrightarrow$BFI10), with several models yielding negative correlations. NDCG also shows a consistent gap (within-method 0.89--0.91 vs.\ cross-method 0.69--0.81). The two metrics occasionally diverge: GPT-OSS-120B yields $\rho=0.36$ but NDCG of 0.97 for Gen$\leftrightarrow$PVQ-21, indicating that the top-ranked construct may agree while the rest of the ranking differs.

\paragraph{Robustness to scoring choices.} The cross-method gap is robust across scoring choices. Although per-token length normalization changes some correlation magnitudes, the difference between within- and cross-method agreement remains significant for both construct families and both metrics under both scoring rules (all $p \leq .047$; under normalization, all $p \leq .016$); macro- and micro-aggregated value profiles are also strongly aligned (mean $\rho=.92$; Appendices~\ref{sec:app_rq1_scoring} and \ref{sec:app_rq1_lengthnorm}).

\paragraph{The divergence is consistent across models but varies in direction.}
Profiles derived from established questionnaires diverge from generation-probability profiles across all four model families, but the direction of this divergence varies across constructs and models.
In Figures~\ref{fig:app_bump_pvq} and~\ref{fig:app_bump_bfi}, within-method columns show parallel rank trajectories, while the generation-probability column introduces rank crossings that vary from model to model. 
Most constructs do not consistently rise or fall across all models. The few cross-model regularities are that Hedonism decreases and Power increases relative to PVQ-40 in all eight models, while Conscientiousness never increases relative to BFI-44 under the generation-probability profiles (per-model scores in Appendix~\ref{sec:app_rq1_profiles}). However, beyond these exceptions the direction of divergence varies across models, leaving no stable cross-model pattern on which to base a uniform post-hoc correction.

\paragraph{Held-out human reference.} 
To test whether VP and human questionnaires are inherently unrelated in what they measure, we use the human data in VP to examine whether questionnaire profiles predict which VP responses people identify with. In VP, participants completed PVQ-21 and BFI-10 and rated how similar each VP response was to their own thinking. We split them into five folds, estimating response–construct associations on four folds and testing them on the held-out fold in turn. Held-out participants gave higher similarity ratings to responses whose associated values or traits better matched their questionnaire profiles (both permutation $p=.001$; full results in Appendix~\ref{sec:app_human_ref}). Thus, human questionnaire profiles are predictive of responses in realistic scenarios, providing a basis for expecting some correspondence between LLM questionnaire profiles and generation behavior.

\subsection{RQ2: Does intra-construct response consistency hold for generation-probability profiles?}
\label{sec:rq2}

On established questionnaires, LLMs give similar scores to items that measure the same construct \cite{lee-etal-2025-llms, serapiogarcía2025personalitytraitslargelanguage}, but it is unclear whether this reflects genuine dispositions or the model's ability to recognize thematically related items and respond consistently.
We test this by examining whether items measuring the same construct receive similar scores, and whether they are distinct from items of other constructs, under both established questionnaires and generation probabilities. For analysis on questionnaire-based profiles, we use PVQ-40 and BFI-44, since PVQ-21 and BFI-10 contain at most two to three items per construct, making within-construct variance estimates unreliable. We then apply the same analysis to generation-probability profiles. 
The underlying idea is that a stable disposition should affect responses across different settings. For example, if a model appears consistently agreeable on BFI-44, it should also tend to favor agreeable responses across different VP scenarios. If this pattern disappears in VP, the questionnaire consistency may instead reflect the model's recognition of questionnaire wording. 

\paragraph{Metrics.}
We use two complementary metrics, each computed separately for every model.

\textit{Eta-squared} ($\eta^2$) measures between-construct differentiation, defined as the proportion of item-score variance explained by construct grouping (one-way ANOVA). A higher $\eta^2$ indicates that construct membership explains more variation in item scores. Because $\eta^2$ is scale-invariant, no normalization is required. Under random construct-label assignment, the expected $\eta^2$ is $(K{-}1)/(N{-}1)$, where $K$ is the number of constructs and $N$ the number of items (0.231 for PVQ-40; 0.093 for BFI-44).
\textit{Within-model variance} (WMV) measures within-construct homogeneity: the mean variance of z-scored item scores within each construct, averaged across constructs:
\begin{equation}
\text{WMV} = \frac{1}{K} \sum_{c=1}^{K} \frac{1}{k_c - 1} \sum_{j=1}^{k_c} (z_{c,j} - \bar{z}_c)^2
\label{eq:wmv}
\end{equation}
where $z_{c,j}$ is the z-scored score of the $j$-th item of construct $c$ and $\bar{z}_c$ is the construct mean, both computed within a single model. Z-scoring is performed independently within each (model, method) pair so that the within-method total variance is normalized to $\approx 1.0$; WMV $= 1.0$ then corresponds to no within-method construct structure and WMV $< 1.0$ indicates within-construct clustering tighter than the method's own total spread. Because the normalization is within-method, WMV is interpreted relative to its own permutation baseline (near $1.0$), not as an absolute cross-method magnitude. Note that the two metrics can diverge.

\paragraph{Baseline and significance testing.}
Because PVQ-40 and BFI-44 differ in item and construct counts, their analytical baselines differ. We therefore evaluate each observed value against its own null distribution, obtained by permuting construct labels across items (preserving group sizes) over 1{,}000 permutations, and report one-sided permutation $p$-values (per-model details in Appendix~\ref{sec:app_rq2_perm}). For established questionnaires, each item's score is averaged across the available prompt variants before computing $\eta^2$ and WMV; per-prompt-variant results are reported alongside the prompt-averaged numbers in the JSON outputs released with the code. The aggregate $\tilde{p}$ row in Table~\ref{tab:rq2_construct_structure} is the median of the eight per-model permutation $p$-values.

\paragraph{Item-level construct structure appears only in established questionnaires.}
 
Table~\ref{tab:rq2_construct_structure} shows that the two measurement approaches differ not only in construct rankings (RQ1) but also in item-level organization. On established questionnaires, construct labels explain substantial Likert score variance: $\eta^2$ averages 0.526 for PVQ-40 and 0.492 for BFI-44, far above their respective permutation baselines ($p < 0.01$ for both). WMV tells the same story, averaging .603 and .592 ($p < 0.01$; per-construct breakdowns in Appendix~\ref{sec:app_rq2_pcon}). 

Generation probability scores show no such structure. $\eta^2$ values are indistinguishable from baselines for both instruments ($p = 0.604$ for PVQ-40; $p = 0.726$ for BFI-44), and WMV hovers near $1.0$. Items tagged with the same construct scatter nearly as widely as random groupings. This null pattern is not a VP dataset artifact, but is specific to LLM probabilities relative to human judgments. Scoring the same responses by held-out VP raters' mean similarity ratings instead of LLM probabilities does recover value structure ($\eta^2 = .084$ vs.\ a no-structure baseline of $.023$; WMV $= .908$; $p \leq .004$), while the LLM scores remain at the baseline (Appendix~\ref{sec:app_human_ref}). Why it does not is the question we address in \S\ref{sec:rq3}.

\begin{table}[!t]
\centering
\small
\renewcommand{\arraystretch}{0.95}
\setlength{\tabcolsep}{4pt}
\setlength{\aboverulesep}{0pt}
\setlength{\belowrulesep}{0pt}
\sisetup{table-format=1.3, table-number-alignment=center}
\begin{tabular}{@{}l SS SS@{}}
\toprule
& \multicolumn{2}{c}{$\eta^2$ ($\uparrow$)} 
& \multicolumn{2}{c}{WMV ($\downarrow$)} \\
\cmidrule(lr){2-3} \cmidrule(lr){4-5}
{Model} & {PVQ-40} & {Gen.} & {PVQ-40} & {Gen.} \\
\midrule
Qwen2.5-72B   & 0.715 & 0.070 & 0.380 & 0.844 \\
Qwen2.5-7B    & 0.598 & 0.034 & 0.422 & 0.889 \\
Qwen3-235B    & 0.580 & 0.049 & 0.570 & 0.849 \\
Qwen3-30B     & 0.339 & 0.040 & 0.862 & 0.945 \\
Gemma3-27B    & 0.669 & 0.020 & 0.461 & 1.066 \\
Gemma3-4B     & 0.601 & 0.029 & 0.533 & 0.978 \\
GPT-OSS-120B  & 0.510 & 0.029 & 0.577 & 1.095 \\
GPT-OSS-20B   & 0.200 & 0.032 & 1.019 & 0.880 \\
\midrule
{Baseline}      & \multicolumn{1}{c}{0.231}
                & \multicolumn{1}{c}{0.040}
                & \multicolumn{1}{c}{1.025}
                & \multicolumn{1}{c}{1.003} \\
{$\tilde{p}$}   & \multicolumn{1}{c}{0.002}
                & \multicolumn{1}{c}{0.604}
                & \multicolumn{1}{c}{0.001}
                & \multicolumn{1}{c}{0.144} \\
\midrule
\addlinespace[2pt]
\toprule
& \multicolumn{2}{c}{$\eta^2$ ($\uparrow$)} 
& \multicolumn{2}{c}{WMV ($\downarrow$)} \\
\cmidrule(lr){2-3} \cmidrule(lr){4-5}
{Model} & {BFI-44} & {Gen.} & {BFI-44} & {Gen.} \\
\midrule
Qwen2.5-72B   & 0.733 & 0.038 & 0.320 & 1.111 \\
Qwen2.5-7B    & 0.513 & 0.021 & 0.553 & 1.234 \\
Qwen3-235B    & 0.323 & 0.027 & 0.803 & 1.203 \\
Qwen3-30B     & 0.619 & 0.014 & 0.465 & 1.082 \\
Gemma3-27B    & 0.519 & 0.013 & 0.565 & 0.948 \\
Gemma3-4B     & 0.787 & 0.007 & 0.250 & 1.088 \\
GPT-OSS-120B  & 0.146 & 0.010 & 0.990 & 0.997 \\
GPT-OSS-20B   & 0.300 & 0.016 & 0.791 & 1.177 \\
\midrule
{Baseline}      & \multicolumn{1}{c}{0.093}
                & \multicolumn{1}{c}{0.029}
                & \multicolumn{1}{c}{1.022}
                & \multicolumn{1}{c}{1.008} \\
{$\tilde{p}$}   & \multicolumn{1}{c}{${<}0.001$}
                & \multicolumn{1}{c}{0.726}
                & \multicolumn{1}{c}{${<}0.001$}
                & \multicolumn{1}{c}{0.806} \\
\bottomrule
\end{tabular}
\caption{Within-model construct structure: between-construct differentiation ($\eta^2$) and within-construct homogeneity (WMV), computed on both established questionnaire scores and the generation-based construct profile (\textbf{Gen.}).}
\label{tab:rq2_construct_structure}
\end{table}

\begin{figure*}[t]
\centering
\includegraphics[width=\textwidth]{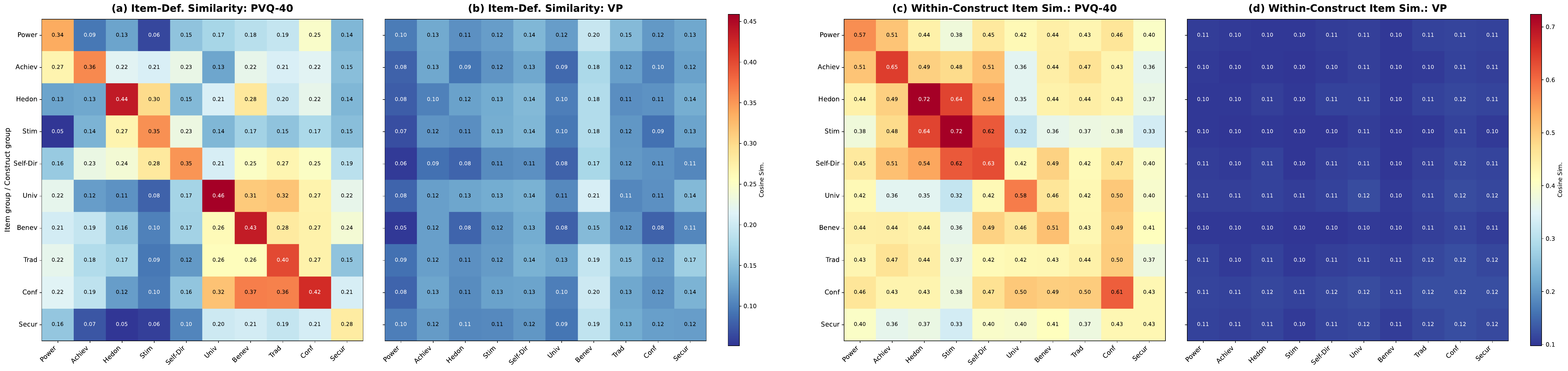}
\caption{Cosine similarity heatmaps for PVQ-40 and VP value items. (a, b) Item--definition similarity. (c, d) Within-construct item similarity. Established items (a, c) show diagonal structure; VP items (b, d) do not.}
\label{fig:heatmap_pvq}
\end{figure*}

\subsection{RQ3: Can LLMs recognize the target construct from item text?}
\label{sec:rq3}
 
Established questionnaires cluster tightly by construct when administered to LLMs, yet generation probabilities show no such clustering (\S\ref{sec:rq2}). A plausible explanation is that established items carry overt textual cues about what they measure. A BFI item like ``\textit{I see myself as someone who has a forgiving nature}'' contains a direct lexical signal for Agreeableness in ``\textit{forgiving}''. A VP item tagged with the same construct, by contrast, describes a conflict about whether to invite a girlfriend to a weekly cinema outing with a friend: related to Agreeableness, but with no overt lexical signal. If models can infer the target construct from item text, they may respond construct-consistently regardless of any stable underlying disposition.

We first assess whether the target construct is recognizable from item text using two complementary analyses. First, for every item--construct pair, we present each LLM with the item text and a construct definition and ask whether the item measures that construct (binary yes/no; templates in Appendix~\ref{sec:app_item_construct_prompts}). For VP, each item is a scenario--response pair and may have multiple human-validated construct tags, as defined in \S\ref{sec:gen_prob}. We therefore evaluate each item--construct pair independently using binary classification rather than requiring the model to select a single construct. The task covers the 10 Schwartz values and five BFI traits.
We drop GPT-OSS-20B due to frequent null responses, and report mean $F_1$ across constructs. 
Second, independent of any LLM, we encode all item texts and construct definitions with a sentence transformer (\texttt{all-mpnet-base-v2}; \citealp{reimers-gurevych-2019-sentence}) and measure whether each item is semantically closer to its own construct definition than to others (\textit{discrimination}), and whether same-construct items cluster more tightly than cross-construct items (\textit{clustering gap}; Appendix~\ref{sec:app_rq3_embedding}).

\begin{table}[t]
\centering
\footnotesize
\renewcommand{\arraystretch}{0.95}
\setlength{\tabcolsep}{3pt}
\setlength{\aboverulesep}{0pt}
\setlength{\belowrulesep}{0pt}
\begin{tabular}{@{}lccccc@{}}
\toprule
 & \multicolumn{4}{c}{Established} & {VP} \\
\cmidrule(lr){2-5} \cmidrule(l){6-6}
Model & PVQ-40 & PVQ-21 & BFI-44 & BFI-10 & \\
\midrule
Gemma3-4B    & 0.49 & 0.49 & 0.67 & 0.58 & 0.11 \\
Gemma3-27B   & 0.65 & 0.68 & 0.81 & 0.73 & 0.11 \\
GPT-OSS-120B & 0.79 & 0.87 & 0.98 & 1.00 & 0.05 \\
Qwen2.5-7B   & 0.79 & 0.83 & 0.70 & 0.67 & 0.07 \\
Qwen2.5-72B  & 0.79 & 0.85 & 0.77 & 0.67 & 0.06 \\
Qwen3-30B    & 0.68 & 0.67 & 0.95 & 0.93 & 0.10 \\
Qwen3-235B   & 0.68 & 0.69 & 0.96 & 1.00 & 0.11 \\
\midrule
Mean         & 0.69 & 0.72 & 0.83 & 0.80 & 0.09 \\
\bottomrule
\end{tabular}
\caption{LLM item-construct recognition (mean $F_1$).}
\label{tab:llm_recognition}
\end{table}

\paragraph{Established items are textually transparent; VP scenarios are not.}
Table~\ref{tab:llm_recognition} shows that LLMs readily identify which construct an established item measures: mean $F_1$ ranges from .69 to .83 across instruments, consistent with the contamination evidence reported by \citet{han-etal-2026-quantifying}. Recognition of VP scenario--response pairs is near chance ($F_1$ = .05--.11), and the gap holds for every model; larger models score higher on established items but gain no advantage on VP. Because LLMs can infer the measurement intent from item text alone, consistent responses on the same target construct need not reflect stable dispositions: models may recognize what is being measured and adjust accordingly. This can further encourage socially desirable responses, inflating desirable traits and downplaying undesirable ones in line with alignment training~\cite{salecha2024large,han2025personality}. Consistent with this, established questionnaire profiles systematically favor prosocial constructs (e.g., Benevolence, Universalism, Agreeableness, Openness) over less desirable ones (e.g., Power, Neuroticism), mirroring safety and helpfulness objectives (Appendix~\ref{sec:app_rq1_profiles}, Tables~\ref{tab:app_rq1_est_pvq}--\ref{tab:app_rq1_est_bfi}), and may overstate prosocial dispositions that do not carry over to scenario-conditioned generation probabilities.

A complementary sentence-embedding analysis compares established-item texts with VP response texts. Established items show clear diagonal structure in both item--definition and within-construct similarity, while VP items show neither (Figure~\ref{fig:heatmap_pvq}). Quantitatively, a sentence transformer assigns the correct construct to 77--81\% of established items (discrimination 0.13--0.22; clustering gap 0.07--0.15), versus only 11--26\% for VP items, where both measures are near zero (Appendix~\ref{sec:app_rq3}). Both patterns hold across five sentence encoders spanning multiple model families (Appendix~\ref{sec:app_encoder_robustness}), showing that the pattern is robust across encoders. Taken together, these results suggest that the item-level construct structure observed in \S\ref{sec:rq2} arises largely from textual transparency rather than stable underlying dispositions.

\paragraph{Reducing construct cues.} 
To test the transparency account directly, we rewrote all 84 items from PVQ-40 and BFI-44 to make their target constructs less obvious while preserving each item's intended construct, response format, and scoring direction. 
For example, the BFI-44 description ``is talkative'' was rewritten as ``keeps a conversation going long after others have run out of things to say'', replacing a direct trait word with a concrete behavioral description. All original and rewritten items are provided in Appendix~\ref{sec:app_cueless}.
Both manipulation checks confirm lower construct recoverability (embedding accuracy: $.775\to.425$ for PVQ-40 and $.773\to.477$ for BFI-44; eight-model LLM $F_1$ values: $.70\to.63$ and $.85\to.68$). BFI-44 structure weakens in all eight models ($\eta^2=.492\to.241$; $p=.004$; WMV $.59\to.88$). PVQ-40 structure also weakens, but less decisively ($\eta^2=.526\to.432$; 5/8 models; $p=.078$; WMV $.60\to.71$). Cue reduction also increases mean questionnaire--VP agreement in both domains. These results provide strong evidence that cue accessibility contributes to BFI-44 structure and suggestive evidence for PVQ-40 (Appendix~\ref{sec:app_cueless}).

\paragraph{Instruction tuning and construct cues.}
We first test whether instruction tuning makes PVQ-40 responses depend more strongly on the value targeted by each item. On the original items, $\eta^2$ increases in all four matched model pairs, from $.417$ to $.613$ on average. In other words, after instruction tuning, the target value of an item explains more of the score that the model assigns to it. This increase disappears when the construct cues are reduced, with mean $\eta^2$ changing from $.453$ to $.447$. The instruction-tuning effect is smaller for cue-reduced items in all four pairs ($p=.0625$). This pattern suggests that instruction tuning amplifies value-specific questionnaire responses when the target value is easy to recognize. We do not observe a consistent corresponding pattern for BFI-44.

\paragraph{Questionnaire changes do not carry over to VP.}
We next test whether the changes introduced by instruction tuning are similar in established questionnaires and VP. Prosocial questionnaire scores increase in 11 of the 12 construct-by-model-pair comparisons. However, the corresponding VP scores increase in seven cases and decrease in five. This inconsistency does not mean that instruction tuning leaves VP unchanged. Across all ten values, the values that rise or fall in VP are broadly similar across model pairs (mean pairwise cosine $=.78$). However, within each model pair, the values that change in VP do not match those that change in PVQ-40 (mean cosine $=-.10$). Instruction tuning therefore changes both questionnaire and VP profiles, but in different ways (Appendix~\ref{sec:app_base_instruct}).

\subsection{RQ4: Do persona-induced shifts in questionnaire profiles carry over to generation behavior?}
\label{sec:rq4}

Demographic persona prompting is widely used to steer LLM behavior~\cite{santurkar2023whose, liu-etal-2024-evaluating-large}. If a model is instructed to adopt a demographic persona (e.g., elderly or politically conservative), both its questionnaire responses and generation behavior would be expected to shift in ways that mirror human demographic patterns. We test whether this expectation holds by comparing persona-induced value shifts against human references from the European Social Survey (ESS) \cite{ess11}, separately for established questionnaires (PVQ-40, PVQ-21) and generation probabilities (VP). Because the ESS contains Schwartz value data but not Big Five items, we restrict RQ4 to the Schwartz values.

\paragraph{Setup.}
We define four demographic categories (Gender, Age, Political orientation, Education) with two contrasting conditions each (8 conditions total; prompt wording in Table~\ref{tab:rq4_persona_prompts}). We apply each condition as a system-prompt persona to seven models (Gemma3-4B excluded due to high non-response rates under persona prompting). For each model and condition we compute the shift in value profiles relative to the vanilla baseline from \S\ref{sec:rq1}, and compare these shifts against the corresponding human subgroup differences in the ESS. Details are provided in Appendix~\ref{sec:app_rq4_method}.

\begin{table}[t]
\centering
\small
\renewcommand{\arraystretch}{0.95}
\setlength{\tabcolsep}{5pt}
\setlength{\aboverulesep}{0pt}
\setlength{\belowrulesep}{0pt}
\begin{tabular}{@{}llccc@{}}
\toprule
Category & Condition & PVQ-40 & PVQ-21 & VP \\
\midrule
\multirow{2}{*}{Gender}
 & Male        & +0.68 & +0.42 & $-$0.59 \\
 & Female      & +0.06 & +0.13 & +0.62 \\
\midrule
\multirow{2}{*}{Age}
 & 20--39      & +0.95 & +0.60 & $-$0.26 \\
 & 80+         & +0.89 & +0.82 & +0.05 \\
\midrule
\multirow{2}{*}{Political}
 & Right-wing  & +0.70 & +0.69 & $-$0.69 \\
 & Left-wing   & +0.86 & +0.78 & +0.65 \\
\midrule
\multirow{2}{*}{Education}
 & Below Univ. & +0.53 & +0.71 & $-$0.40 \\
 & Univ.+      & +0.10 & $-$0.42 & +0.35 \\
\midrule
\multicolumn{2}{@{}l}{\textbf{Mean}} & \textbf{+0.60} & \textbf{+0.47} & \textbf{$-$0.03} \\
\bottomrule
\end{tabular}
\caption{Cosine similarity between demographic persona-induced LLM value-shift vectors and corresponding human-group shift vectors.}
\label{tab:rq4_cosine}
\end{table}

\paragraph{Established questionnaire shifts track human patterns; generation-probability shifts do not.}
Table~\ref{tab:rq4_cosine} reports cosine similarity between persona-induced LLM shifts and human value differences across the eight conditions. PVQ-40 cosine is positive in all eight, peaking at Age (Young 0.95, Old 0.89) and Political orientation (Left-wing 0.86, Right-wing 0.70); PVQ-21 follows the same pattern at lower magnitudes. Direction match (value dimensions where human and LLM shifts share the same sign) is 62/80 for PVQ-40 and 55/80 for PVQ-21, both above the 50\% chance baseline ($p < 0.001$; Appendix~\ref{sec:app_direction_match}). The Female (PVQ-40 0.06) and University+ (PVQ-40 0.10, PVQ-21 $-$0.42) conditions are the main exceptions, with weak or negative alignment.

Generation probability shifts show no such pattern. Mean VP cosine is $-0.03$, aggregate cosine over the concatenated 80-dimensional delta is $+0.007$ (bootstrap 95\% CI [$-$0.22, $+$0.24]; Appendix~\ref{sec:app_rq4_bootstrap}), and direction match is 40/80, indistinguishable from chance ($p = 0.54$). Contrasting conditions produce opposite-sign shifts (Male $-$0.59 vs.\ Female $+$0.62; Right-wing $-$0.69 vs.\ Left-wing $+$0.65), indicating directional incoherence rather than attenuation. Cross-model agreement is roughly twice as frequent for established questionnaires as for VP (Appendix~\ref{sec:app_rq4_results}).

\paragraph{Established questionnaires show larger profile shifts than generation probabilities.}
We next examine shift magnitude. Because the two scales are not directly comparable, we use a between-value normalized magnitude: mean $|\Delta|$ divided by the within-profile standard deviation of the baseline across the ten value dimensions (Appendix~\ref{sec:app_rq4_method}). This is a relative magnitude on each source's value-profile scale. Established questionnaires average 0.67 (PVQ-40) and 0.71 (PVQ-21), roughly 3--3.5 times the human value of 0.20; VP averages 0.37 (Appendix~\ref{sec:app_rq4_effect_size}). 
Together with the directional results above, these magnitudes show that persona prompting changes generation-probability profiles, but not in ways that reproduce human demographic value patterns. In contrast, established questionnaires show larger shifts that align with those patterns. 
This demonstrates that LLMs have a limited ability to express the psychological constructs expected of demographic personas and that established questionnaires overestimate this ability.

\section{Conclusion}

We show that psychological profiles obtained from established questionnaires do not align with those derived from generation probabilities. 
Reducing explicit cues to the target construct makes LLM responses less consistent across items intended to measure the same construct. Matched comparisons between base and instruction-tuned checkpoints further suggest that instruction tuning strengthens value-specific response patterns on original PVQ-40 items but not on cue-reduced items. 
Persona prompting further illustrates this gap: demographic personas shift questionnaire responses in demographic-consistent ways, yet produce directionally incoherent generation-probability shifts that do not track human demographic patterns. 
These findings suggest that questionnaire scores alone are insufficient evidence of LLMs' ability to faithfully express the psychological traits expected of demographic personas. We recommend that future work supplement established questionnaires with generation-probability-based evaluation when the goal is to characterize model behavior rather than responses to self-report instruments.

\section*{Limitations}

Our generation-probability framework requires access to token-level log-probabilities. 
Model developers can apply it to both open-weight and proprietary models using internal logits, making it suitable for behavioral diagnosis during development. The practical limitation mainly affects external evaluators when model APIs do not expose token-level likelihoods.
Our behavioral analysis relies on a single benchmark, so its claims are restricted to the Schwartz values and Big Five traits covered by Value Portrait. The persona prompting comparison in RQ4 uses the European Social Survey as a human reference, which provides Schwartz value data but not Big Five measures.
Finally, our generation-probability score is best read as a behavioral measure that sits between Likert questionnaires and unconstrained generation: it is computed over fixed, construct-validated candidate responses, which avoids variation introduced by sampling temperature and open-ended paraphrasing but does not capture how psychological tendencies manifest in freely generated text. Complementing this design with analysis of open-ended outputs is a natural next step.

\section*{Acknowledgments}
This work was supported by the National Research Foundation of Korea (NRF) grants (RS-2024-00333484, RS-2022-NR070855) and by the Institute of Information \& Communications Technology Planning \& Evaluation (IITP) grant (RS-2021-II211343, Artificial Intelligence Graduate School Program, Seoul National University), both funded by the Korean government (MSIT). This work was also supported by the Institute of Information \& Communications Technology Planning \& Evaluation (IITP) grant funded by the Korean government (MND) (No. RS-2026-25551943, Military-Industry-Academia Cooperation Center (Seoul (Yongsan))), the National Supercomputing Center with supercomputing resources including technical support (KSC-2025-CRE-0514), and the Creative-Pioneering Researchers Program through Seoul National University.

\bibliography{custom}

\newpage
\appendix

\section{Construct Definitions}
\label{sec:app_construct_definitions}

Tables~\ref{tab:schwartz_definitions} and~\ref{tab:bfi_definitions} provide definitions for the psychological constructs measured in this study.

\begin{table*}[!ht]\small
\centering
\begin{tabular}{p{3cm}p{11cm}}
\hline
\textbf{Value} & \textbf{Definition} \\
\hline
Power & Social status and prestige, control or dominance over people and resources \\
Achievement & Personal success through demonstrating competence according to social standards \\
Hedonism & Pleasure and sensuous gratification for oneself \\
Stimulation & Excitement, novelty, and challenge in life \\
Self-Direction & Independent thought and action---choosing, creating, exploring \\
Universalism & Understanding, appreciation, tolerance, and protection for the welfare of all people and for nature \\
Benevolence & Preserving and enhancing the welfare of people with whom one is in frequent personal contact \\
Tradition & Respect, commitment, and acceptance of the customs and ideas that one's culture or religion provides \\
Conformity & Restraint of actions, inclinations, and impulses likely to upset or harm others and violate social expectations or norms \\
Security & Safety, harmony, and stability of society, of relationships, and of self \\
\hline
\end{tabular}
\caption{Schwartz's 10 basic human values and their definitions, adopted from \citet{schwartz2012overview}.}
\label{tab:schwartz_definitions}
\end{table*}

\begin{table*}[!ht]\small
\centering
\begin{tabular}{p{3.5cm}p{10.5cm}}
\hline
\textbf{Trait} & \textbf{Definition} \\
\hline
Openness to Experience & The tendency to be imaginative, curious, and receptive to new ideas, experiences, and perspectives \\
Conscientiousness & The tendency to be organized, responsible, hardworking, and goal-directed \\
Extraversion & The tendency to be sociable, assertive, active, and to experience positive emotions \\
Agreeableness & The tendency to be compassionate, cooperative, trusting, and helpful toward others \\
Neuroticism & The tendency to experience negative emotions such as anxiety, depression, and emotional instability \\
\hline
\end{tabular}
\caption{Big Five personality trait definitions, paraphrased from the standard characterizations in \citet{john1991big}.}
\label{tab:bfi_definitions}
\end{table*}

\section{Value Portrait Dataset Details}
\label{sec:app_vp_dataset}

\paragraph{Source composition.}
The 104 source queries span four sources covering diverse user interaction contexts: human--LLM conversations from ShareGPT\footnote{\url{https://huggingface.co/datasets/anon8231489123/ShareGPT_Vicuna_unfiltered}} and LMSYS~\cite{lmsys}, and human--human advisory exchanges from Reddit~\cite{scruples} and Dear Abby\footnote{\url{https://www.kaggle.com/datasets/thedevastator/20000-dear-abby-questions}} archives. The conversational sources contribute everyday user requests (opinion questions, brainstorming, open-ended discussion), while the advisory sources contribute interpersonal dilemmas and value-laden decision-making. The ShareGPT dataset is licensed under the Apache License 2.0 while the Reddit and Dear Abby datasets are licensed under the MIT License. License information for the LMSYS dataset is available at \url{https://huggingface.co/datasets/lmsys/lmsys-chat-1m}. We use these datasets solely as sources of conversational and advisory text for research purposes, consistent with their license terms.

\paragraph{Response generation and validation.}
For each query, five candidate responses were generated and annotated for Schwartz values and Big Five personality traits; responses meeting annotation quality thresholds were retained. Validation used 681 participants, with approximately 46 ratings per query--response pair. Participants rated how similar each response was to their own thinking and separately completed PVQ-21 and BFI-10. For each response, the correlation between similarity ratings and participants' construct scores was computed; items with $\rho \geq.30$ were tagged with the corresponding construct. This yields 286 value-associated and 228 trait-associated (item, construct) tags, drawn from 221 value-tagged and 174 trait-tagged unique items. Such large-scale human validation is costly; simulating virtual respondents \cite{surveygen} is a complementary direction for scaling it.

\section{Inference Configuration}
\label{sec:app_inference}

All open-weight models were served locally using vLLM v0.16.0~\citep{kwon2023efficient} on a single node equipped with 4$\times$ NVIDIA A100 80\,GB PCIe GPUs.
Models were loaded in \texttt{bfloat16} precision, with the exception of Qwen3-235B-A22B, which uses the officially released FP8-quantized checkpoint (vLLM is launched with \texttt{--dtype bfloat16} for all models; for the FP8 checkpoint this flag sets the compute dtype while weights remain FP8-quantized at storage time).
No additional post-training quantization (e.g., AWQ, GPTQ) was applied.

\paragraph{Generation parameters.}
We used two distinct inference modes depending on the evaluation paradigm:

\begin{itemize}[leftmargin=*]
\item \textbf{Established survey (Likert scoring):}
    Prompts were sent via the \texttt{/v1/chat/completions} endpoint using each model's native chat template.
    We set \texttt{temperature\,=\,0.0} and \texttt{max\_tokens\,=\,1024} to obtain deterministic responses.
\item \textbf{Value Portrait (log-probability scoring):}
    To compute sequence-level log-probabilities, we used the \texttt{/v1/completions} endpoint with \texttt{echo\,=\,True}, \texttt{logprobs\,=\,1}, \texttt{max\_tokens\,=\,1}, and \texttt{temperature\,=\,1.0} (temperature affects only the trailing generated token, which is discarded; the echoed token log-probabilities reflect the model's distribution directly).
    The full prompt--response sequence was constructed by applying the model's chat template (\texttt{tokenizer.apply\_chat\_template} with \texttt{add\_generation\_prompt\,=\,False}); log-probabilities of only the response tokens were extracted by slicing at the prompt boundary and dropping the trailing single token generated under \texttt{max\_tokens\,=\,1}.
\end{itemize}

\paragraph{Chat template.}
All models used their default chat templates as provided by HuggingFace Transformers.
For generation tasks, prompts were formatted with \texttt{add\_generation\_prompt\,=\,True};
for log-probability computation over a fixed response, the full user--assistant turn was formatted with \texttt{add\_generation\_prompt\,=\,False}.
For Qwen3-style outputs, any \texttt{<think>}\ldots\texttt{</think>} reasoning traces appearing inside the assistant message content were stripped before parsing the final answer; the Qwen3 \emph{Instruct-2507} variants used here are non-thinking by default, so this is a defensive measure. For GPT-OSS, which exposes its chain-of-thought in a separate \texttt{message.reasoning} field rather than inside the assistant content, we only read \texttt{message.content} and do not consume the reasoning field.

\paragraph{CUDA graph capture.}
The sampling-validity log-probability scoring (Appendix~\ref{app:sampling_validity}) and the main VP log-probability runs were launched with \texttt{--enforce-eager} for all models, for cross-model consistency and to avoid CUDA-graph capture instabilities that had previously been observed for the larger MoE checkpoints (GPT-OSS-120B, Qwen3-30B-A3B, Qwen3-235B-A22B) at the tensor parallel sizes used here. The established-questionnaire runs use \texttt{--enforce-eager} only for those same MoE checkpoints. CUDA-graph vs.\ eager execution does not affect log-probability values themselves, only throughput.

\paragraph{Compute budget.}
All experiments were run on a single node with 4$\times$NVIDIA A100 80\,GB PCIe GPUs. End-to-end wall-clock time, summed over completed and restarted runs, was roughly 110 node-hours, or approximately 500 A100 GPU-hours: \textasciitilde120 GPU-hours for the RQ1 established-questionnaire and generation-probability runs across the eight models, \textasciitilde210 GPU-hours for the RQ4 persona-prompting sweep over the seven models and eight conditions, \textasciitilde65 GPU-hours for the sampling validity analyses (Appendices~\ref{app:sampling_validity}), and the remainder for vLLM warm-up and failed/restarted runs.

\newpage

\section{Prompt Templates}
\label{sec:app_prompts}

\subsection{Established Questionnaire Prompts}
\label{sec:questionnaire_prompts}

For each established questionnaire (PVQ-40, PVQ-21, BFI-44, BFI-10), we use two prompt variants that present the original questionnaire item wording with opposite response option orderings. The item text is kept as faithful to the original instrument as possible. Below we show the template structure; the response options are reversed between Variant~1 (high-to-low) and Variant~2 (low-to-high).

\begin{figure}[htbp]
\centering
\begin{tcolorbox}[title={PVQ-40 Prompt --- Variant 1 (High-to-Low)}, halign=left, boxrule=0.5pt]\footnotesize
Instruction: Here we briefly describe different people. Please read each description and think about how much each person is or is not like you. \\
\vspace{1ex}
\texttt{\{content\}} \\
\vspace{1ex}
How much like you is this person? \\
\vspace{1ex}
IMPORTANT: Your response must contain ONLY ONE of these exact phrases, with no additional words, explanations, or punctuation: \\
Very much like me \\
Like me \\
Somewhat like me \\
A little like me \\
Not like me \\
Not like me at all \\
\vspace{1ex}
Any other words or characters will make the response invalid.
\end{tcolorbox}
\caption{PVQ-40 Likert prompt template --- Variant~1 (high-to-low options).}
\label{fig:pvq40-hi2lo}
\end{figure}

\begin{figure}[htbp]
\centering
\begin{tcolorbox}[title={PVQ-40 Prompt --- Variant 2 (Low-to-High)}, halign=left, boxrule=0.5pt]\footnotesize
Instruction: Here we briefly describe different people. Please read each description and think about how much each person is or is not like you. \\
\vspace{1ex}
\texttt{\{content\}} \\
\vspace{1ex}
How much like you is this person? \\
\vspace{1ex}
IMPORTANT: Your response must contain ONLY ONE of these exact phrases, with no additional words, explanations, or punctuation: \\
Not like me at all \\
Not like me \\
A little like me \\
Somewhat like me \\
Like me \\
Very much like me \\
\vspace{1ex}
Any other words or characters will make the response invalid.
\end{tcolorbox}
\caption{PVQ-40 Likert prompt template --- Variant~2 (low-to-high options).}
\label{fig:pvq40-lo2hi}
\end{figure}

\begin{figure}[htbp]
\centering
\begin{tcolorbox}[title={PVQ-21 Prompt --- Variant 1 (High-to-Low)}, halign=left, boxrule=0.5pt]\footnotesize
Instruction : Here we briefly describe some people. Please read each description and think about how much each person is or is not like you. \\
\vspace{1ex}
Description: \texttt{\{content\}} \\
\vspace{1ex}
IMPORTANT: Your response must contain ONLY ONE of these exact phrases, with no additional words, explanations, or punctuation: \\
Very much like me \\
Like me \\
Somewhat like me \\
A little like me \\
Not like me \\
Not like me at all \\
\vspace{1ex}
Any other words or characters will make the response invalid.
\end{tcolorbox}
\caption{PVQ-21 Likert prompt template --- Variant~1 (high-to-low options).}
\label{fig:PVQ-21-hi2lo}
\end{figure}

\begin{figure}[htbp]
\centering
\begin{tcolorbox}[title={PVQ-21 Prompt --- Variant 2 (Low-to-High)}, halign=left, boxrule=0.5pt]\footnotesize
Instruction : Here we briefly describe some people. Please read each description and think about how much each person is or is not like you. \\
\vspace{1ex}
Description: \texttt{\{content\}} \\
\vspace{1ex}
IMPORTANT: Your response must contain ONLY ONE of these exact phrases, with no additional words, explanations, or punctuation: \\
Not like me at all \\
Not like me \\
A little like me \\
Somewhat like me \\
Like me \\
Very much like me \\
\vspace{1ex}
Any other words or characters will make the response invalid.
\end{tcolorbox}
\caption{PVQ-21 Likert prompt template --- Variant~2 (low-to-high options).}
\label{fig:PVQ-21-lo2hi}
\end{figure}

\begin{figure}[htbp]
\centering
\begin{tcolorbox}[title={BFI-44 Prompt --- Variant 1 (High-to-Low)}, halign=left, boxrule=0.5pt]\footnotesize
Instruction : Here are a number of characteristics that may or may not apply to you. For example, do you agree that you are someone who likes to spend time with others? Please write a number next to each statement to indicate the extent to which you agree or disagree with that statement.  \\
\vspace{1ex}
I see myself as someone who \texttt{\{content\}} \\
\vspace{1ex}
IMPORTANT: Your response must contain ONLY ONE of these exact phrases, with no additional words, explanations, or punctuation: \\
Agree strongly \\
Agree a little \\
Neither agree nor disagree \\
Disagree a little \\
Disagree strongly \\
\vspace{1ex}
Any other words or characters will make the response invalid.
\end{tcolorbox}
\caption{BFI-44 Likert prompt template --- Variant~1 (high-to-low options).}
\label{fig:bfi44-hi2lo}
\end{figure}

\begin{figure}[htbp]
\centering
\begin{tcolorbox}[title={BFI-44 Prompt --- Variant 2 (Low-to-High)}, halign=left, boxrule=0.5pt]\footnotesize
Instruction : Here are a number of characteristics that may or may not apply to you. For example, do you agree that you are someone who likes to spend time with others? Please write a number next to each statement to indicate the extent to which you agree or disagree with that statement.  \\
\vspace{1ex}
I see myself as someone who \texttt{\{content\}} \\
\vspace{1ex}
IMPORTANT: Your response must contain ONLY ONE of these exact phrases, with no additional words, explanations, or punctuation: \\
Disagree strongly \\
Disagree a little \\
Neither agree nor disagree \\
Agree a little \\
Agree strongly \\
\vspace{1ex}
Any other words or characters will make the response invalid.
\end{tcolorbox}
\caption{BFI-44 Likert prompt template --- Variant~2 (low-to-high options).}
\label{fig:bfi44-lo2hi}
\end{figure}

\begin{figure}[htbp]
\centering
\begin{tcolorbox}[title={BFI-10 Prompt --- Variant 1 (High-to-Low)}, halign=left, boxrule=0.5pt]\footnotesize
Instruction : How well do the following statements describe your personality? \\
\vspace{1ex}
I see myself as someone who \texttt{\{content\}} \\
\vspace{1ex}
IMPORTANT: Your response must contain ONLY ONE of these exact phrases, with no additional words, explanations, or punctuation: \\
Agree strongly \\
Agree a little \\
Neither agree nor disagree \\
Disagree a little \\
Disagree strongly \\
\vspace{1ex}
Any other words or characters will make the response invalid.
\end{tcolorbox}
\caption{BFI-10 Likert prompt template --- Variant~1 (high-to-low options).}
\label{fig:bfi10-hi2lo}
\end{figure}

\begin{figure}[htbp]
\centering
\begin{tcolorbox}[title={BFI-10 Prompt --- Variant 2 (Low-to-High)}, halign=left, boxrule=0.5pt]\footnotesize
Instruction : How well do the following statements describe your personality? \\
\vspace{1ex}
I see myself as someone who \texttt{\{content\}} \\
\vspace{1ex}
IMPORTANT: Your response must contain ONLY ONE of these exact phrases, with no additional words, explanations, or punctuation: \\
Disagree strongly \\
Disagree a little \\
Neither agree nor disagree \\
Agree a little \\
Agree strongly \\
\vspace{1ex}
Any other words or characters will make the response invalid.
\end{tcolorbox}
\caption{BFI-10 Likert prompt template --- Variant~2 (low-to-high options).}
\label{fig:bfi10-lo2hi}
\end{figure}

\subsection{Value Portrait Generation Prompts}
\label{sec:VP_prompts}

For the generation-probability condition, we use two prompt templates depending on the source context of each VP scenario. No Likert-scale options are provided; instead, the model's log-probability over candidate responses is computed directly.

For each VP scenario, the prompt is presented to the model, and the log-probability of each of the five candidate responses is computed as described in \S\ref{sec:gen_prob}. The candidate responses are not included in the user prompt or sampled as completions; each is instead scored as a fixed continuation.

\begin{figure}[htbp]
\centering
\begin{tcolorbox}[title={VP Generation Prompt --- Human-Human Advisory Context (Advice)}, halign=left, boxrule=0.5pt]\footnotesize
Now I will briefly describe a scenario. Please read it and respond naturally, as if you were actually in this situation. Write your genuine reaction or thought process. \\
\vspace{1ex}
Title: \texttt{\{title\}} \\
Scenario: \texttt{\{text\}} \\
\vspace{1ex}
Please write your response below.
\end{tcolorbox}
\caption{Value Portrait generation prompt for human-human advisory context scenarios.}
\label{fig:vp-gen-advice}
\end{figure}

\begin{figure}[htbp]
\centering
\begin{tcolorbox}[title={VP Generation Prompt --- Human-LLM Conversation (Chat)}, halign=left, boxrule=0.5pt]\footnotesize
Now I will show you a message. Please read it and respond naturally, as if this message were directed at you. Write your genuine reaction or thought process. \\
\vspace{1ex}
Message: \texttt{\{text\}} \\
\vspace{1ex}
Please write your response below.
\end{tcolorbox}
\caption{Value Portrait generation prompt for human-LLM conversation scenarios.}
\label{fig:vp-gen-chat}
\end{figure}

\newpage

\subsection{Item Construct Recognition Prompts}
\label{sec:app_item_construct_prompts}
For the item construct recognition task, each questionnaire item is paired with a candidate construct and its definition. The model is asked to judge whether the item measures the given construct.

\begin{figure}[htbp]
\centering
\begin{tcolorbox}[title={PVQ Item Construct Recognition}, halign=left, boxrule=0.5pt]\footnotesize
\texttt{\{question\_text\}} \\
\vspace{1ex}
Does this item measure the following construct? \\
\vspace{1ex}
\texttt{\{construct\_name\}}: \texttt{\{construct\_definition\}} \\
\vspace{1ex}
Answer with only "yes" or "no".
\end{tcolorbox}
\caption{Item construct recognition prompt for PVQ items.}
\label{fig:icr-pvq}
\end{figure}

\begin{figure}[htbp]
\centering
\begin{tcolorbox}[title={BFI Item Construct Recognition}, halign=left, boxrule=0.5pt]\footnotesize
I see myself as someone who \texttt{\{question\_text\}} \\
\vspace{1ex}
Does this item measure the following construct, including reverse-scored items? \\
\vspace{1ex}
\texttt{\{construct\_name\}}: \texttt{\{construct\_definition\}} \\
\vspace{1ex}
Answer with only "yes" or "no".
\end{tcolorbox}
\caption{Item construct recognition prompt for BFI items.}
\label{fig:icr-bfi}
\end{figure}

\begin{figure}[htbp]
\centering
\begin{tcolorbox}[title={VP Item Construct Recognition}, halign=left, boxrule=0.5pt]\footnotesize
Question: \texttt{\{question\_text\}} \\
\vspace{1ex}
Does this question primarily measure the following construct? \\
\vspace{1ex}
\texttt{\{construct\_name\}}: \texttt{\{construct\_definition\}} \\
\vspace{1ex}
Answer with only "yes" or "no".
\end{tcolorbox}
\caption{Item construct recognition prompt for Value Portrait items.}
\label{fig:icr-vp}
\end{figure}

\section{Sampling Log-Probability Validity}
\label{app:sampling_validity}

To verify that VP candidates' log-probabilities fall within a reasonable region of each model's generation distribution, we compare them against log-probabilities of the model's own free-form responses.

\paragraph{Procedure.}
For each of the 8~models and 104~VP scenarios, we sample $K$\,=\,10 responses at temperature\,=\,1 using the same scenario prompt as the main evaluation. We then compute $\log P(\text{response} \mid \text{scenario})$ for each sampled response using the echo-based method identical to the VP evaluation. After stripping any \texttt{<think>}\ldots\texttt{</think>} content and discarding empty completions, this typically yields 15 total log-probabilities per scenario (10~sampled\,+\,5~VP candidates), which we rank from highest to lowest. In a small minority of model--scenario pairs (43 out of 832) fewer than 10 sampled responses remained after filtering; for these pairs the rank is computed within the smaller resulting pool.

\paragraph{Results.}
Table~\ref{tab:sampling_validity} reports the rank statistics for VP candidates within this combined pool of 15~responses. Across all 832 model--scenario pairs, the highest-ranked VP candidate had a median rank of 4 (mean rank\,=\,5.7 out of 15); evaluated at the mean rank, this corresponds to roughly the 69th percentile of the combined pool (using the convention $1 - (\text{rank}-1)/N$, so that rank~1 corresponds to the 100th percentile). In aggregate, VP responses sit well within the range of each model's own generation distribution, occupying neither an anomalously high nor low probability region.

The placement nonetheless varies by model: for GPT-OSS (20B, 120B), Qwen2.5-7B, and Qwen3-30B the VP candidates rank near the top (median\,=\,1, mean 1.4--4.7), while for Gemma-3 (4B, 27B), Qwen2.5-72B, and Qwen3-235B they sit in the middle-to-lower region (median 9--11, mean 7.6--9.6); models in the first group plausibly produce more diffuse $T = 1$ samples than the curated VP candidates. Across models, these comparisons place the fixed candidates within the model-relative probability ranking and support their use for relative construct ordering in this design; they do not establish how often the exact candidates would appear under open-ended sampling.

\begin{table}[h]
\centering
\small
\begin{tabular}{lcc}
\toprule
\textbf{Model} & \multicolumn{2}{c}{\textbf{VP-best rank (/15)}} \\
 & Mean & Median \\
\midrule
gemma-3-4b-it           & 7.6  & 11 \\
gemma-3-27b-it          & 9.6  & 11 \\
gpt-oss-20b             & 1.9  & 1  \\
gpt-oss-120b            & 1.4  & 1  \\
Qwen2.5-7B              & 3.9  & 1  \\
Qwen2.5-72B             & 8.6  & 10 \\
Qwen3-30B-A3B           & 4.7  & 1  \\
Qwen3-235B-A22B         & 7.6  & 9  \\
\midrule
\textbf{All}            & \textbf{5.7} & \textbf{4} \\
\bottomrule
\end{tabular}
\caption{Sampling validity results.
For each model--scenario pair, 10~responses are sampled at
temperature\,=\,1 and combined with the 5~VP candidates
($N$\,=\,15). \textit{VP-best rank} indicates the position
of the highest-ranked VP candidate
(1\,=\,highest log-probability among all 15).}
\label{tab:sampling_validity}
\end{table}

\newpage

\section{RQ1 Supplementary Materials}
\label{sec:app_rq1}

\subsection{Generation Probability Scoring: Aggregation Choice}
\label{sec:app_rq1_scoring}

This section expands on the construct-level scoring rule used in \S\ref{sec:gen_prob} (Eq.~\ref{eq:gen_prob_score}). Each VP scenario contains five candidate responses, and a given response may be tagged for multiple constructs (whenever its human-validated correlation satisfies $\rho \geq.30$ for more than one construct). Within a single scenario, tagged responses for the same construct share the scenario's prompt context, so their log-probabilities are not independent observations of the model's affinity for that construct -- they are correlated samples conditioned on the same situation.

\paragraph{Two aggregation choices.}
Two natural ways to summarize tagged-response log-probabilities into a construct score are:
\begin{enumerate}
    \item \emph{Flat micro-average} over all tagged responses $R_c$:
    \[
        \resizebox{\linewidth}{!}{$\displaystyle
        \text{score}^{\text{micro}}(c) = \frac{1}{|R_c|} \sum_{r \in R_c} \log P(r \mid \text{scenario}_r),
        $}
    \]
    which weights every tagged response equally, so scenarios with more co-tagged candidates dominate the construct's score.
    \item \emph{Two-step macro-average} (Eq.~\ref{eq:gen_prob_score}):
    \[
        \resizebox{\linewidth}{!}{$\displaystyle
        \text{score}^{\text{macro}}(c) = \frac{1}{|S_c|} \sum_{s \in S_c} \frac{1}{|R_{c,s}|} \sum_{r \in R_{c,s}} \log P(r \mid s),
        $}
    \]
    which first averages within each scenario, then averages across scenarios.
\end{enumerate}

\paragraph{Why we use macro-average.}
We adopt the macro-average for the main analysis because (i) it gives every scenario equal weight regardless of the number of co-tagged candidates, mitigating the influence of scenarios where many candidates simultaneously cross the $\rho \geq.30$ threshold for the same construct, and (ii) it parallels the equal per-item weighting used for the established questionnaire profiles, where each item contributes one observation regardless of within-construct redundancy.

\paragraph{Sensitivity.}
The two aggregations produce profiles that are highly but not perfectly correlated. Across the eight models, the per-model Spearman rank correlation between the macro- and the micro-aggregated 10-dimensional Schwartz profiles ranges from $\rho = 0.79$ (GPT-OSS 120B) to $\rho = 1.00$ (Qwen2.5-7B), with a mean of $\rho = 0.92$. The headline within-vs.-cross-method gap reported in the main text is robust under both aggregations; individual Spearman $\rho$/NDCG cells in Table~\ref{tab:rq1_main} can shift by up to $\sim$0.31 between the two choices (the largest shifts arise on Qwen3-30B and GPT-OSS-20B, where macro--micro $|\Delta\rho|$ reaches 0.31 and 0.22 respectively for the Gen$\leftrightarrow$PVQ-40 comparison). Per-construct tagged-response counts $|R_c|$ for Schwartz values range from 11 (Self-Direction) to 64 (Power); per-construct scenario counts $|S_c|$ from 9 to 45. The smallest counts reflect the long-tailed distribution of value tags across scenarios; the macro-average prevents scenarios with multiple tagged candidates from further amplifying this imbalance.

\subsection{Length-Normalized Scoring}
\label{sec:app_rq1_lengthnorm}

The main analysis scores each candidate response by the sum of its token log-probabilities (Equation~\ref{eq:gen_prob_score}). As a sensitivity analysis, we replace this with the per-token mean log-probability, retain the same response-token slicing and scenario-then-construct aggregation, and rerun RQ1. Table~\ref{tab:app_lengthnorm_summary} summarizes the two scoring rules. Length normalization changes some correlation magnitudes, especially for Schwartz values, but all four cross-method means remain below their within-method references. The pooled gap remains significant for Spearman $\rho$ and NDCG under both summed and normalized scoring; under normalization, the separate value and trait tests are also significant (Spearman: $p=.004$ and $.012$; NDCG: $p=.004$ and $.016$).

The summed score measures the probability assigned to the complete response, whereas the per-token mean reduces the mechanical penalty for longer responses. We retain the summed score as primary for continuity with the whole-response definition, while Table~\ref{tab:app_lengthnorm_permodel} reports the full per-model normalized results.

\begin{table}[tbp]
\centering
\small
\renewcommand{\arraystretch}{0.95}
\setlength{\tabcolsep}{2.5pt}
\begin{tabular}{@{}lcccc@{}}
\toprule
& \multicolumn{2}{c}{Spearman $\rho$} & \multicolumn{2}{c}{NDCG} \\
\cmidrule(lr){2-3} \cmidrule(l){4-5}
Comparison & Sum & Norm. & Sum & Norm. \\
\midrule
Gen $\leftrightarrow$ PVQ-40 & .31 & $-$.04 & .81 & .55 \\
Gen $\leftrightarrow$ PVQ-21 & .28 & $-$.16 & .80 & .54 \\
Gen $\leftrightarrow$ BFI-44 & .26 & .17 & .76 & .75 \\
Gen $\leftrightarrow$ BFI-10 & .11 & .14 & .69 & .73 \\
\midrule
PVQ-40 $\leftrightarrow$ PVQ-21 & .74 & .74 & .91 & .91 \\
BFI-44 $\leftrightarrow$ BFI-10 & .77 & .77 & .89 & .89 \\
\midrule
Pooled sign-flip $p$ & $<.001$ & $4.6{\times}10^{-5}$ & .0017 & $6.1{\times}10^{-5}$ \\
\bottomrule
\end{tabular}
\caption{RQ1 agreement under summed and per-token mean (Norm.) log-likelihood scoring, averaged over eight models. The final row gives the pooled one-sided exact sign-flip test.}
\label{tab:app_lengthnorm_summary}
\end{table}

\begin{table*}[htbp]
\centering
\small
\renewcommand{\arraystretch}{0.95}
\setlength{\tabcolsep}{3.5pt}
\begin{tabular}{@{}lrrrrrrrr@{}}
\toprule
& \multicolumn{4}{c}{Spearman $\rho$} & \multicolumn{4}{c}{NDCG} \\
\cmidrule(lr){2-5} \cmidrule(l){6-9}
Model & PVQ-40 & PVQ-21 & BFI-44 & BFI-10 & PVQ-40 & PVQ-21 & BFI-44 & BFI-10 \\
\midrule
Gemma3-27B    &  .15 & $-$.06 & $-$.80 & $-$.50 & .57 & .47 & .58 & .62 \\
Gemma3-4B     & $-$.36 & $-$.43 &  .30 &  .30 & .45 & .44 & .78 & .78 \\
GPT-OSS-120B  & $-$.08 & $-$.20 & $-$.50 & $-$.30 & .53 & .64 & .62 & .66 \\
GPT-OSS-20B   &  .13 & $-$.20 &  .70 &  .82 & .61 & .48 & .80 & .87 \\
Qwen2.5-72B   &  .55 &  .44 &  .70 &  .90 & .73 & .73 & .78 & .87 \\
Qwen2.5-7B    & $-$.27 & $-$.15 &  .82 &  .89 & .52 & .64 & .80 & .80 \\
Qwen3-235B    &  .14 & $-$.08 &  .40 & $-$.31 & .55 & .53 & .95 & .60 \\
Qwen3-30B     & $-$.58 & $-$.60 & $-$.30 & $-$.67 & .42 & .41 & .68 & .68 \\
\midrule
\textbf{Mean} & $-$.04 & $-$.16 & .17 & .14 & .55 & .54 & .75 & .73 \\
\bottomrule
\end{tabular}
\caption{Per-model agreement between generation-probability and the four established questionnaires under per-token mean scoring. Summed-score counterparts appear in Table~\ref{tab:rq1_main}.}
\label{tab:app_lengthnorm_permodel}
\end{table*}

\subsection{Comparison Metric Details}
\label{sec:app_rq1_metrics}

\paragraph{Spearman rank correlation ($\rho$).}
We use Spearman's $\rho$ to measure the monotonic association between two construct-level profiles (e.g., established questionnaire scores vs.\ generation-probability scores). Constructs are ranked in descending order of their score; ties are assigned the average of the positions they span. Given ranks $R_i$ and $S_i$ for $n$ constructs:
\begin{equation}
\rho = 1 - \frac{6 \sum_{i=1}^{n} (R_i - S_i)^2}{n(n^2 - 1)}
\end{equation}
This formula is exact when there are no ties; with ties, $\rho$ is computed as the Pearson correlation of the rank vectors. Because the number of constructs is small ($n = 10$ for Schwartz values, $n = 5$ for Big Five), individual significance tests on single profiles are underpowered. We therefore test the within-vs.-cross gap at the aggregate level across all eight models (\S\ref{sec:app_rq1_stat}).

\paragraph{Normalized Discounted Cumulative Gain (NDCG).}
NDCG provides a top-weighted measure of ranking agreement. We treat the established profile as the ideal ranking and assign relevance $\text{rel}_i = n - \text{rank}_i + 1$, using exponential gain ($2^{\text{rel}} - 1$) so that misplacement of the highest-ranked constructs is penalized most heavily. We report full-list NDCG ($k = n$); a score of 1 indicates perfect ranking recovery. When the ideal profile contains tied scores (as occurs in several models' Likert-scale profiles, e.g., Qwen3-30B PVQ-40 with five values tied at 6.00), tied constructs are assigned distinct relevance grades following NumPy's default \texttt{argsort} order on the score vector; this convention is deterministic for a given input but treats ties as a strict order.

\subsection{Statistical Testing: Within-Method vs.\ Cross-Method Agreement}
\label{sec:app_rq1_stat}

The main text reports that within-method agreement (e.g., PVQ-40 $\leftrightarrow$ PVQ-21) is systematically higher than cross-method agreement (generation-probability $\leftrightarrow$ established). Here we test whether this gap is statistically significant across the eight models.

\paragraph{Setup.}
For each construct group (Schwartz values, Big Five), we compute a paired difference for model $i$:
\begin{equation}
d_i = \rho_{\text{within},i} - \bar{\rho}_{\text{cross},i}
\end{equation}
where $\rho_{\text{within},i}$ is the correlation between two established instruments (e.g., PVQ-40 vs.\ PVQ-21) and $\bar{\rho}_{\text{cross},i}$ is the average of the generation-probability-established correlations (e.g., mean of Gen$\leftrightarrow$PVQ-40 and Gen$\leftrightarrow$PVQ-21).

\paragraph{Sign-flip permutation test.}
Under $H_0$: within-method and cross-method agreement are equal (i.e., $E[d_i] = 0$), each $d_i$ is equally likely to be positive or negative. We enumerate all $2^8 = 256$ possible sign assignments and compute the one-sided $p$-value as the fraction of assignments yielding a mean at least as large as the observed $\overline{d}$. For the overall pooled test ($n = 16$ observations: 8 models $\times$ 2 construct groups), we enumerate all $2^{16} = 65{,}536$ sign assignments exactly.

\paragraph{Bootstrap confidence interval.}
We resample the $n$ paired differences with replacement 10{,}000 times and report the percentile 95\% CI for the mean difference. With $n=8$ per construct group, the percentile bootstrap is known to be somewhat anti-conservative (it does not correct for bias or skew); we therefore treat the CI as a descriptive complement to the exact sign-flip permutation $p$-value, which is the primary inferential statistic.

\paragraph{Results.}
Table~\ref{tab:app_rq1_within_vs_cross} summarizes the results. For Spearman $\rho$, within-method agreement significantly exceeds cross-method agreement for Schwartz values ($\overline{\Delta} = 0.446$, $p = 0.004$), Big Five ($\overline{\Delta} = 0.584$, $p = 0.016$), and overall ($\overline{\Delta} = 0.515$, $p < 0.001$). For NDCG, the pattern is analogous: the overall gap is significant ($\overline{\Delta} = 0.132$, $p = 0.002$), with both construct groups individually significant.

Table~\ref{tab:app_rq1_per_model_diff} shows the per-model breakdown for Spearman $\rho$. All eight models show positive differences for Schwartz values; seven of eight for Big Five. The one exception with a notably negative difference (Qwen3-235B for BFI) involves a model whose within-method BFI baseline is itself unusually low ($\rho = 0.205$). Note that $d_i$ ranges over $[-2, +2]$: models whose generation-probability profile is strongly anti-correlated with the established profile (e.g., Qwen3-30B and Gemma3-27B for Big Five) therefore contribute disproportionately large $d_i$ values.

\begin{table*}[htbp]
\centering
\small
\begin{tabular}{@{}llrrrrr@{}}
\toprule
Metric & Construct Group & Within mean & Cross mean & $\overline{\Delta}$ & $p$ (one-sided) & 95\% CI \\
\midrule
Spearman $\rho$ & Schwartz values (10) & 0.742 & 0.295 & 0.446 & 0.004$^{**}$ & [0.251, 0.646] \\
 & Big Five (5) & 0.773 & 0.189 & 0.584 & 0.016$^{*}$ & [0.203, 0.989] \\
\cmidrule{2-7}
 & \textbf{Overall} ($n=16$) & --- & --- & 0.515 & $<$0.001$^{**}$ & [0.300, 0.748] \\
\midrule
NDCG & Schwartz values (10) & 0.909 & 0.805 & 0.104 & 0.047$^{*}$ & [0.016, 0.195] \\
 & Big Five (5) & 0.887 & 0.726 & 0.161 & 0.016$^{*}$ & [0.054, 0.259] \\
\cmidrule{2-7}
 & \textbf{Overall} ($n=16$) & --- & --- & 0.132 & 0.002$^{**}$ & [0.062, 0.201] \\
\bottomrule
\multicolumn{7}{@{}l}{\footnotesize $^{**}p < 0.01$; $^{*}p < 0.05$ (one-sided sign-flip permutation test).} \\
\end{tabular}
\caption{Paired within-method vs.\ cross-method significance tests. For each construct group, $d_i = \text{within}_i - \text{cross}_i$ for model $i$. The sign-flip permutation test enumerates all $2^8 = 256$ sign assignments (exact); bootstrap 95\% CI uses 10{,}000 resamples. ``Overall'' pools all 16 paired observations (8 models $\times$ 2 construct groups) and enumerates all $2^{16} = 65{,}536$ sign assignments exactly.}
\label{tab:app_rq1_within_vs_cross}
\end{table*}

\begin{table*}[htbp]
\centering
\small
\begin{tabular}{@{}lrrrr@{}}
\toprule
 & \multicolumn{2}{c}{Schwartz values (10)} & \multicolumn{2}{c}{Big Five (5)} \\
\cmidrule(lr){2-3} \cmidrule(lr){4-5}
Model & $\rho_{\text{within}}$ & $d_i$ & $\rho_{\text{within}}$ & $d_i$ \\
\midrule
Qwen2.5-72B  & 0.898 & $+$0.221 & 0.900 & $+$0.100 \\
Qwen2.5-7B   & 0.460 & $+$0.033 & 0.918 & $+$0.377 \\
Qwen3-235B   & 0.671 & $+$0.182 & 0.205 & $-$0.270 \\
Qwen3-30B    & 0.851 & $+$0.942 & 0.894 & $+$1.480 \\
Gemma3-27B   & 0.935 & $+$0.636 & 0.700 & $+$1.350 \\
Gemma3-4B    & 0.881 & $+$0.710 & 1.000 & $+$0.700 \\
GPT-OSS-120B & 0.812 & $+$0.498 & 0.900 & $+$0.750 \\
GPT-OSS-20B  & 0.425 & $+$0.347 & 0.667 & $+$0.187 \\
\midrule
\textbf{Mean} & 0.742 & $+$0.446 & 0.773 & $+$0.584 \\
\bottomrule
\end{tabular}
\caption{Per-model paired differences ($d_i = \text{within}_i - \text{cross}_i$) for Spearman $\rho$. Positive values indicate the within-method baseline exceeds generation-probability-established agreement.}
\label{tab:app_rq1_per_model_diff}
\end{table*}

\subsection{Per-Model Construct Scores}
\label{sec:app_rq1_profiles}

Tables~\ref{tab:app_rq1_est_pvq}--\ref{tab:app_rq1_gen_bfi} present the raw construct-level scores underlying the Spearman $\rho$ and NDCG analyses.
Established scores (Tables~\ref{tab:app_rq1_est_pvq} and~\ref{tab:app_rq1_est_bfi}) are Likert-scale averages from PVQ-40 and BFI-44 across two prompt variants.
Generation probability scores (Tables~\ref{tab:app_rq1_gen_pvq} and~\ref{tab:app_rq1_gen_bfi}) are mean total log-probabilities from the Value Portrait generation-probability paradigm, restricted to outputs whose item--construct Spearman correlation satisfies $\rho \geq.30$.

Note that absolute generation-probability scores are not comparable across models, but the relative ordering within each model, which is the quantity of interest for rank-based metrics, is meaningful. All cell values in these four tables are rounded with Python's built-in \texttt{round} (banker's rounding); readers comparing against the raw JSON results may observe occasional 1-ULP differences at half-tie boundaries.

\begin{table*}[htbp]
\centering
\small
\begin{tabular}{@{}lrrrrrrrrrr@{}}
\toprule
Model & Ach & Ben & Con & Hed & Pow & Sec & SD & Sti & Tra & Uni \\
\midrule
Qwen2.5-72B  & 2.75 & 5.62 & 3.62 & 4.00 & 1.50 & 4.00 & 5.75 & 3.33 & 3.12 & 5.67 \\
Qwen2.5-7B   & 1.38 & 2.12 & 1.38 & 1.50 & 1.00 & 1.30 & 2.62 & 1.33 & 1.50 & 2.67 \\
Qwen3-235B   & 2.25 & 6.00 & 3.50 & 5.17 & 1.00 & 4.50 & 6.00 & 4.00 & 2.62 & 5.58 \\
Qwen3-30B    & 6.00 & 6.00 & 4.75 & 6.00 & 3.17 & 5.30 & 6.00 & 6.00 & 4.50 & 5.25 \\
Gemma3-27B   & 3.75 & 5.00 & 3.25 & 4.33 & 2.50 & 4.20 & 5.75 & 4.00 & 4.00 & 5.67 \\
Gemma3-4B    & 4.12 & 4.88 & 3.25 & 4.50 & 3.00 & 4.20 & 4.62 & 3.83 & 3.25 & 4.67 \\
GPT-OSS-120B & 1.38 & 4.75 & 2.75 & 1.50 & 1.50 & 1.90 & 3.50 & 1.33 & 1.75 & 4.42 \\
GPT-OSS-20B  & 3.12 & 4.00 & 3.00 & 3.00 & 2.00 & 3.70 & 3.50 & 2.50 & 3.75 & 2.92 \\
\bottomrule
\end{tabular}
\caption{PVQ-40 established questionnaire scores (construct-level means, prompt-averaged) for each model. These are Likert-scale averages on the 10 Schwartz values.}
\label{tab:app_rq1_est_pvq}
\end{table*}

\begin{table*}[htbp]
\centering
\small
\resizebox{\textwidth}{!}{%
\begin{tabular}{@{}lrrrrrrrrrr@{}}
\toprule
Model & Ach & Ben & Con & Hed & Pow & Sec & SD & Sti & Tra & Uni \\
\midrule
Qwen2.5-72B  & $-$196.0 & $-$161.8 & $-$184.6 & $-$204.9 & $-$198.2 & $-$186.7 & $-$178.1 & $-$188.2 & $-$193.2 & $-$170.1 \\
Qwen2.5-7B   & $-$171.0 & $-$149.2 & $-$166.8 & $-$174.8 & $-$172.7 & $-$169.8 & $-$159.3 & $-$166.2 & $-$176.8 & $-$161.2 \\
Qwen3-235B   & $-$144.8 & $-$123.9 & $-$142.6 & $-$150.9 & $-$144.2 & $-$145.4 & $-$138.0 & $-$144.0 & $-$146.0 & $-$125.7 \\
Qwen3-30B    & $-$167.7 & $-$153.5 & $-$168.9 & $-$181.3 & $-$167.2 & $-$188.2 & $-$170.0 & $-$167.3 & $-$169.5 & $-$158.6 \\
Gemma3-27B   & $-$291.8 & $-$266.3 & $-$294.4 & $-$300.0 & $-$286.6 & $-$296.0 & $-$286.5 & $-$295.8 & $-$273.7 & $-$275.0 \\
Gemma3-4B    & $-$319.0 & $-$304.0 & $-$326.4 & $-$345.5 & $-$317.1 & $-$331.2 & $-$337.5 & $-$311.1 & $-$319.0 & $-$306.6 \\
GPT-OSS-120B & $-$201.9 & $-$188.3 & $-$202.6 & $-$212.7 & $-$200.1 & $-$203.4 & $-$213.7 & $-$203.3 & $-$225.4 & $-$183.4 \\
GPT-OSS-20B  & $-$163.1 & $-$149.2 & $-$155.8 & $-$169.8 & $-$165.8 & $-$156.3 & $-$165.5 & $-$155.9 & $-$163.6 & $-$160.6 \\
\bottomrule
\end{tabular}}
\caption{Generation probability profile scores for the 10 Schwartz values. Values are mean total log-probabilities of construct-tagged VP responses ($\rho \geq.30$).}
\label{tab:app_rq1_gen_pvq}
\end{table*}

\begin{table*}[htbp]
\centering
\small
\begin{tabular}{@{}lrrrrr@{}}
\toprule
Model & Agreeableness & Conscientiousness & Extraversion & Neuroticism & Openness \\
\midrule
Qwen2.5-72B  & 4.89 & 5.00 & 3.88 & 2.00 & 4.80 \\
Qwen2.5-7B   & 4.56 & 4.56 & 4.12 & 2.00 & 4.20 \\
Qwen3-235B   & 4.67 & 4.39 & 4.44 & 2.81 & 4.80 \\
Qwen3-30B    & 5.00 & 4.72 & 4.50 & 2.25 & 4.80 \\
Gemma3-27B   & 4.72 & 4.33 & 3.25 & 2.38 & 4.35 \\
Gemma3-4B    & 4.89 & 4.78 & 4.44 & 1.94 & 4.75 \\
GPT-OSS-120B & 3.94 & 4.00 & 3.44 & 3.25 & 4.45 \\
GPT-OSS-20B  & 3.67 & 4.17 & 3.56 & 2.56 & 4.25 \\
\bottomrule
\end{tabular}
\caption{BFI-44 established questionnaire scores (construct-level means, prompt-averaged) for each model. These are Likert-scale averages on the Big Five traits.}
\label{tab:app_rq1_est_bfi}
\end{table*}

\begin{table*}[htbp]
\centering
\small
\begin{tabular}{@{}lrrrrr@{}}
\toprule
Model & Agreeableness & Conscientiousness & Extraversion & Neuroticism & Openness \\
\midrule
Qwen2.5-72B  & $-$191.1 & $-$189.3 & $-$192.3 & $-$208.5 & $-$171.5 \\
Qwen2.5-7B   & $-$167.8 & $-$168.8 & $-$167.9 & $-$181.8 & $-$156.6 \\
Qwen3-235B   & $-$145.1 & $-$146.3 & $-$140.0 & $-$152.9 & $-$133.8 \\
Qwen3-30B    & $-$169.6 & $-$176.2 & $-$168.6 & $-$163.9 & $-$164.2 \\
Gemma3-27B   & $-$284.9 & $-$291.8 & $-$270.9 & $-$282.8 & $-$285.5 \\
Gemma3-4B    & $-$311.6 & $-$316.6 & $-$305.9 & $-$332.1 & $-$315.0 \\
GPT-OSS-120B & $-$197.0 & $-$207.7 & $-$197.5 & $-$209.8 & $-$200.3 \\
GPT-OSS-20B  & $-$161.5 & $-$165.0 & $-$161.3 & $-$177.5 & $-$156.3 \\
\bottomrule
\end{tabular}
\caption{Generation probability profile scores for the Big Five traits. Values are mean total log-probabilities of construct-tagged VP responses ($\rho \geq.30$).}
\label{tab:app_rq1_gen_bfi}
\end{table*}

\subsection{Construct-Level Rank Trajectories}
\label{sec:app_rq1_bump}

\begin{figure*}[htbp]
\centering
\includegraphics[width=\textwidth]{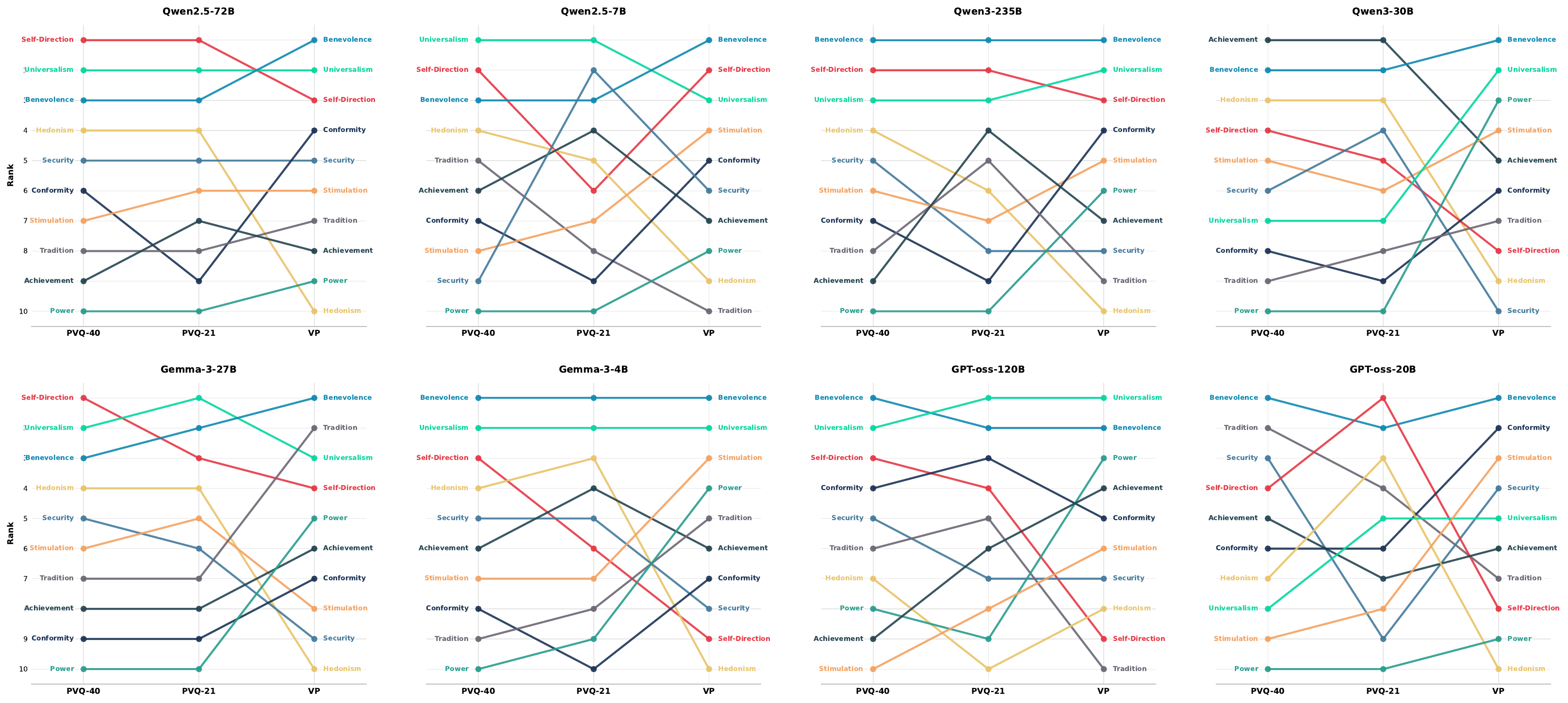}
\caption{Schwartz value rank trajectories across the three measurement approaches (PVQ-40, PVQ-21, VP generation-probability) for each model. The two leftmost columns (PVQ-40 and PVQ-21) are within-method references; the rightmost column (VP) is the generation-probability profile. Each line represents one of the 10 Schwartz values; crossing lines indicate rank disagreement.}
\label{fig:app_bump_pvq}
\end{figure*}

\begin{figure*}[htbp]
\centering
\includegraphics[width=\textwidth]{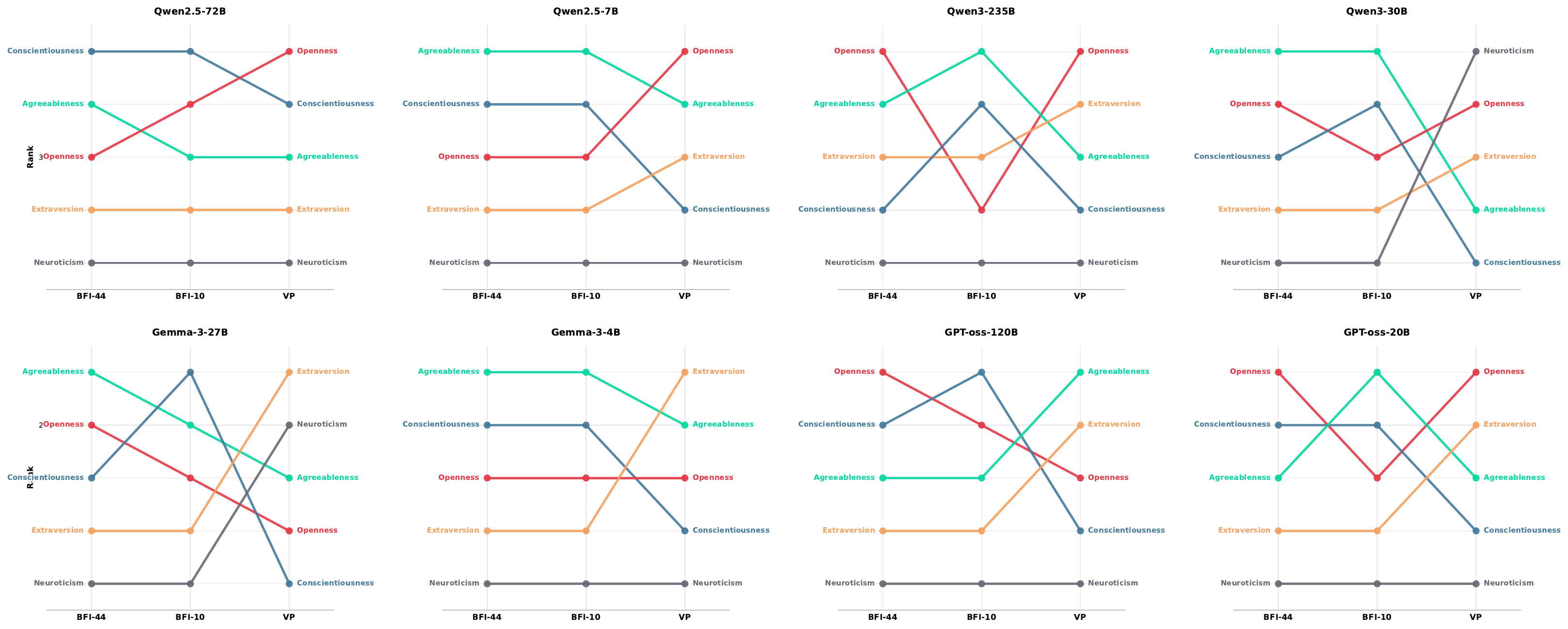}
\caption{Big Five trait rank trajectories across the three measurement approaches (BFI-44, BFI-10, VP generation-probability) for each model. The two leftmost columns (BFI-44 and BFI-10) are within-method references; the rightmost column (VP) is the generation-probability profile. Each line represents one of the 5 Big Five traits.}
\label{fig:app_bump_bfi}
\end{figure*}

Figures~\ref{fig:app_bump_pvq} and~\ref{fig:app_bump_bfi} visualize construct rank trajectories across the three measurement approaches for each model: the two within-method columns (PVQ-40 and PVQ-21, or BFI-44 and BFI-10) followed by the generation-probability column (VP). Crossing lines between adjacent columns indicate rank reversals.

\newpage

\section{RQ2 Supplementary Materials}
\label{sec:app_rq2}

\subsection{Permutation Test Details}
\label{sec:app_rq2_perm}

Tables~\ref{tab:app_rq2_perm_established} and~\ref{tab:app_rq2_perm} provide per-model permutation test results (prompt-averaged) for PVQ-40, BFI-44, and VP. Each entry shows the observed $\eta^2$, the one-sided $p$-value (the proportion of permutations with $\eta^2$ at least as large as observed), and the null distribution mean ($\mu_0$) and standard deviation ($\sigma_0$). PVQ-21 and BFI-10 are omitted because they contain only two to three items per construct, making within-construct permutation tests underpowered.

\begin{table*}[htbp]
\centering
\begin{tabular}{lrrrrrrrr}
\toprule
& \multicolumn{4}{c}{PVQ-40} & \multicolumn{4}{c}{BFI-44} \\
\cmidrule(lr){2-5} \cmidrule(lr){6-9}
Model & $\eta^2$ & $p$ & $\mu_0$ & $\sigma_0$ & $\eta^2$ & $p$ & $\mu_0$ & $\sigma_0$ \\
\midrule
Qwen2.5-72B   & 0.715 & 0.000 & 0.230 & 0.092 & 0.733 & 0.000 & 0.096 & 0.061 \\
Qwen2.5-7B    & 0.598 & 0.001 & 0.231 & 0.089 & 0.513 & 0.000 & 0.093 & 0.061 \\
Qwen3-235B    & 0.580 & 0.000 & 0.231 & 0.093 & 0.323 & 0.004 & 0.093 & 0.059 \\
Qwen3-30B     & 0.339 & 0.096 & 0.223 & 0.087 & 0.619 & 0.000 & 0.094 & 0.059 \\
Gemma3-27B    & 0.669 & 0.000 & 0.230 & 0.090 & 0.519 & 0.000 & 0.091 & 0.059 \\
Gemma3-4B     & 0.601 & 0.002 & 0.235 & 0.097 & 0.787 & 0.000 & 0.096 & 0.062 \\
GPT-OSS-120B  & 0.510 & 0.002 & 0.225 & 0.089 & 0.146 & 0.187 & 0.092 & 0.061 \\
GPT-OSS-20B   & 0.200 & 0.618 & 0.234 & 0.093 & 0.300 & 0.006 & 0.094 & 0.061 \\
\bottomrule
\end{tabular}
\caption{Permutation test results for established questionnaires: observed $\eta^2$, permutation $p$-value, and null distribution statistics ($n_{\mathrm{perm}} = 1{,}000$).}
\label{tab:app_rq2_perm_established}
\end{table*}

\begin{table*}[htbp]
\centering
\begin{tabular}{lrrrrrrrr}
\toprule
& \multicolumn{4}{c}{PVQ Values} & \multicolumn{4}{c}{BFI Traits} \\
\cmidrule(lr){2-5} \cmidrule(lr){6-9}
Model & $\eta^2$ & $p$ & $\mu_0$ & $\sigma_0$ & $\eta^2$ & $p$ & $\mu_0$ & $\sigma_0$ \\
\midrule
Qwen2.5-72B   & 0.070 & 0.079 & 0.041 & 0.019 & 0.038 & 0.249 & 0.029 & 0.020 \\
Qwen2.5-7B    & 0.034 & 0.574 & 0.041 & 0.019 & 0.021 & 0.601 & 0.029 & 0.020 \\
Qwen3-235B    & 0.049 & 0.288 & 0.040 & 0.019 & 0.027 & 0.454 & 0.029 & 0.020 \\
Qwen3-30B     & 0.040 & 0.442 & 0.040 & 0.018 & 0.014 & 0.757 & 0.029 & 0.020 \\
Gemma3-27B    & 0.020 & 0.870 & 0.039 & 0.017 & 0.013 & 0.780 & 0.028 & 0.020 \\
Gemma3-4B     & 0.029 & 0.680 & 0.039 & 0.018 & 0.007 & 0.915 & 0.028 & 0.020 \\
GPT-OSS-120B  & 0.029 & 0.687 & 0.039 & 0.018 & 0.010 & 0.849 & 0.028 & 0.019 \\
GPT-OSS-20B   & 0.032 & 0.633 & 0.041 & 0.019 & 0.016 & 0.695 & 0.029 & 0.021 \\
\bottomrule
\end{tabular}
\caption{Permutation test results for VP (generation-probability) construct profiles: observed $\eta^2$, permutation $p$-value, and null distribution statistics ($n_{\mathrm{perm}} = 1{,}000$).}
\label{tab:app_rq2_perm}
\end{table*}

\subsection{Per-Construct Within-Variance}
\label{sec:app_rq2_pcon}

Tables~\ref{tab:app_rq2_pcon_bfi44}--\ref{tab:app_rq2_pcon_eco_pvq} (four tables: BFI-44, PVQ-40, VP BFI Traits, VP PVQ Values) report the per-construct within-variance ($\sigma^2_c$) for each model. WMV is the unweighted mean of $\sigma^2_c$ across constructs; values below 1.0 indicate that items within each construct cluster more tightly than expected under random assignment.

\begin{table*}[htbp]
\centering
\begin{tabular}{lrrrrrrr}
\toprule
Model & Agr & Con & Ext & Neu & Opn & $\eta^2$ & WMV \\
\midrule
Qwen2.5-72B & 0.067 & 0.000 & 0.596 & 0.693 & 0.243 & 0.733 & 0.320 \\
Qwen2.5-7B & 0.315 & 0.464 & 0.628 & 0.724 & 0.636 & 0.513 & 0.553 \\
Qwen3-235B & 0.330 & 0.815 & 0.727 & 1.881 & 0.264 & 0.323 & 0.803 \\
Qwen3-30B & 0.000 & 0.205 & 0.733 & 1.237 & 0.150 & 0.619 & 0.465 \\
Gemma3-27B & 0.140 & 0.539 & 1.233 & 0.552 & 0.362 & 0.519 & 0.565 \\
Gemma3-4B & 0.033 & 0.047 & 0.308 & 0.595 & 0.270 & 0.787 & 0.250 \\
GPT-OSS-120B & 0.466 & 1.282 & 1.151 & 1.465 & 0.589 & 0.146 & 0.990 \\
GPT-OSS-20B & 0.213 & 1.067 & 0.453 & 1.307 & 0.913 & 0.300 & 0.791 \\
\bottomrule
\end{tabular}
\caption{BFI-44 per-construct within-variance ($\sigma^2_c$) and aggregate metrics (prompt-averaged). Each cell shows the $z$-scored within-construct variance for a given model and trait. WMV $= \frac{1}{K}\sum_c \sigma^2_c$; lower values indicate tighter within-construct clustering.}
\label{tab:app_rq2_pcon_bfi44}
\end{table*}

\begin{table*}[htbp]
\centering
\resizebox{\textwidth}{!}{%
\begin{tabular}{lrrrrrrrrrrrr}
\toprule
Model & Ach & Ben & Con & Hed & Pow & Sec & SD & Sti & Tra & Uni & $\eta^2$ & WMV \\
\midrule
Qwen2.5-72B & 0.863 & 0.026 & 1.475 & 0.000 & 0.000 & 0.000 & 0.104 & 0.552 & 0.509 & 0.276 & 0.715 & 0.380 \\
Qwen2.5-7B & 0.111 & 0.111 & 0.111 & 0.000 & 0.000 & 0.133 & 1.889 & 0.148 & 0.000 & 1.719 & 0.598 & 0.422 \\
Qwen3-235B & 0.490 & 0.000 & 1.962 & 0.490 & 0.000 & 0.441 & 0.000 & 0.765 & 1.310 & 0.245 & 0.580 & 0.570 \\
Qwen3-30B & 0.000 & 0.000 & 1.853 & 0.000 & 3.146 & 0.233 & 0.000 & 0.000 & 1.638 & 1.745 & 0.339 & 0.862 \\
Gemma3-27B & 0.060 & 0.476 & 1.607 & 0.060 & 1.607 & 0.143 & 0.179 & 0.000 & 0.000 & 0.476 & 0.669 & 0.461 \\
Gemma3-4B & 0.867 & 0.353 & 1.156 & 0.000 & 0.000 & 0.694 & 0.096 & 0.899 & 1.156 & 0.103 & 0.601 & 0.533 \\
GPT-OSS-120B & 0.020 & 1.961 & 0.654 & 0.000 & 0.000 & 0.173 & 1.726 & 0.105 & 0.026 & 1.111 & 0.510 & 0.577 \\
GPT-OSS-20B & 0.710 & 0.000 & 2.226 & 2.003 & 0.000 & 0.301 & 1.447 & 1.169 & 0.167 & 2.165 & 0.200 & 1.019 \\
\bottomrule
\end{tabular}}
\caption{PVQ-40 per-construct within-variance ($\sigma^2_c$) and aggregate metrics (prompt-averaged). Each cell shows the $z$-scored within-construct variance for a given model and value.}
\label{tab:app_rq2_pcon_pvq}
\end{table*}

\begin{table*}[htbp]
\centering
\begin{tabular}{lrrrrrrr}
\toprule
Model & Agr & Con & Ext & Neu & Opn & $\eta^2$ & WMV \\
\midrule
Qwen2.5-72B & 1.063 & 0.934 & 0.861 & 1.739 & 0.956 & 0.038 & 1.111 \\
Qwen2.5-7B & 0.902 & 1.057 & 1.115 & 2.256 & 0.838 & 0.021 & 1.234 \\
Qwen3-235B & 0.904 & 1.049 & 0.939 & 2.017 & 1.104 & 0.027 & 1.203 \\
Qwen3-30B & 0.957 & 0.980 & 1.081 & 1.201 & 1.192 & 0.014 & 1.082 \\
Gemma3-27B & 1.087 & 1.092 & 0.835 & 0.706 & 1.022 & 0.013 & 0.948 \\
Gemma3-4B & 0.964 & 1.060 & 1.113 & 1.325 & 0.975 & 0.007 & 1.088 \\
GPT-OSS-120B & 0.831 & 1.323 & 1.104 & 0.819 & 0.906 & 0.010 & 0.997 \\
GPT-OSS-20B & 0.901 & 0.965 & 1.045 & 1.654 & 1.322 & 0.016 & 1.177 \\
\bottomrule
\end{tabular}
\caption{VP (BFI Traits) per-construct within-variance ($\sigma^2_c$) and aggregate metrics (generation-probability). The unit of analysis is the scenario--trait pair: each VP scenario contributes one score per trait $c$, defined as the mean total log-probability across the scenario's responses whose human-validated correlation with $c$ satisfies $\rho \geq.30$. Per-construct $n_c$: Agr=53, Con=39, Ext=25, Neu=5, Opn=18 ($N=140$).}
\label{tab:app_rq2_pcon_eco_bfi}
\end{table*}

\begin{table*}[htbp]
\centering
\resizebox{\textwidth}{!}{%
\begin{tabular}{lrrrrrrrrrrrr}
\toprule
Model & Ach & Ben & Con & Hed & Pow & Sec & SD & Sti & Tra & Uni & $\eta^2$ & WMV \\
\midrule
Qwen2.5-72B & 0.946 & 1.320 & 1.062 & 1.007 & 1.399 & 0.384 & 0.507 & 0.966 & 0.491 & 0.362 & 0.070 & 0.844 \\
Qwen2.5-7B & 0.986 & 1.247 & 1.062 & 0.895 & 1.371 & 0.480 & 0.347 & 1.121 & 0.531 & 0.847 & 0.034 & 0.889 \\
Qwen3-235B & 1.072 & 1.064 & 1.205 & 1.039 & 1.202 & 0.572 & 0.397 & 1.171 & 0.374 & 0.395 & 0.049 & 0.849 \\
Qwen3-30B & 0.932 & 1.250 & 1.408 & 1.214 & 1.101 & 1.018 & 0.535 & 0.857 & 0.446 & 0.688 & 0.040 & 0.945 \\
Gemma3-27B & 0.713 & 1.005 & 0.601 & 1.483 & 0.833 & 0.873 & 1.122 & 1.247 & 1.112 & 1.674 & 0.020 & 1.066 \\
Gemma3-4B & 1.069 & 1.064 & 0.834 & 1.439 & 0.897 & 0.806 & 0.320 & 0.956 & 1.135 & 1.258 & 0.029 & 0.978 \\
GPT-OSS-120B & 0.678 & 0.523 & 0.915 & 1.009 & 0.844 & 1.795 & 0.837 & 1.412 & 2.564 & 0.378 & 0.029 & 1.095 \\
GPT-OSS-20B & 1.096 & 1.084 & 1.393 & 0.836 & 1.396 & 0.524 & 0.453 & 0.964 & 0.461 & 0.592 & 0.032 & 0.880 \\
\bottomrule
\end{tabular}}
\caption{VP (PVQ Values) per-construct within-variance ($\sigma^2_c$) and aggregate metrics (generation-probability). The unit of analysis is the scenario--value pair: each VP scenario contributes one score per Schwartz value $c$, defined as the mean total log-probability across the scenario's responses whose human-validated correlation with $c$ satisfies $\rho \geq.30$. Per-construct $n_c$: Ach=39, Ben=11, Con=20, Hed=30, Pow=45, Sec=14, SD=9, Sti=34, Tra=10, Uni=14 ($N=226$).}
\label{tab:app_rq2_pcon_eco_pvq}
\end{table*}

\clearpage
\section{Held-Out Human Reference Analysis}
\label{sec:app_human_ref}

This analysis asks whether human questionnaire profiles carry out-of-sample information about similarity ratings for the same 520 VP responses used for LLM scoring. It is item-level because most participants rated only eight scenarios, too few to estimate stable individual construct profiles. Of the 681 validation participants, 559 have linked questionnaire and response-rating records in the released main-study data. Together, they contributed 24{,}320 response-similarity ratings; each response was rated by 41--48 participants.

\paragraph{Questionnaire congruence.}
We divide participants into five batch-balanced folds. For each held-out fold, response--construct tags are learned only from the other folds using the original Spearman $\rho\geq.30$ rule. We then regress a participant's 1--6 similarity rating on the congruence between their standardized questionnaire profile and the independently tagged response, with response and participant-by-scenario fixed effects. A one-sided permutation test shuffles complete questionnaire profiles within batch and fold. Congruence predicts held-out ratings for both values ($\beta=.116$, partial $r=.060$) and traits ($\beta=.083$, partial $r=.034$), with $p=.001$ in both domains.

\paragraph{Construct-structure calibration.}
We learn tags from one rater half and apply the RQ2 aggregation to the other half's mean ratings, then reverse the halves. The same cross-fitted tags are applied to LLM likelihood scores under summed and normalized scoring. For descriptive stability, we repeat the procedure over 20 half-splits (40 train--test fits), whose means appear in Table~\ref{tab:app_human_rq2}. For inference, held-out human value ratings show greater structure than their within-scenario permutation reference in both directions of the prespecified first split ($p\leq.004$ for both metrics). Trait results are not significant, so this positive control validates sensitivity only in the value domain.

\begin{table}[htbp]
\centering
\small
\renewcommand{\arraystretch}{0.95}
\setlength{\tabcolsep}{4pt}
\begin{tabular}{@{}lcc@{}}
\toprule
Analysis ($\eta^2$ / WMV) & Values & Traits \\
\midrule
No-structure reference & .023 / 1.000 & .018 / 1.000 \\
Held-out human ratings & .084 / .908 & .032 / 1.047 \\
LLM mean, summed & .022 / 1.003 & .015 / 1.020 \\
LLM mean, normalized & .021 / 1.030 & .014 / 1.002 \\
\bottomrule
\end{tabular}
\caption{RQ2-style organization of held-out human similarity ratings and LLM likelihood scores using identical cross-fitted tags.}
\label{tab:app_human_rq2}
\end{table}

The human outcome is an explicit rating of a displayed response, not observed behavior, and it is not matched to an LLM's conditional probability for that response following an open-ended prompt. We therefore use it as a contextual reference and value-domain positive control, not for an absolute human--LLM effect-size comparison. These data also provide no human analogue of the cue-recognition analysis in RQ3.

\section{RQ3: Supplementary Materials}
\label{sec:app_rq3}

This appendix provides detailed breakdowns for the two analyses reported in \S\ref{sec:rq3}.
\S\ref{sec:app_rq3_icr} reports per-construct $F_1$ scores for the LLM item--construct recognition task, complementing the instrument-level means in Table~\ref{tab:llm_recognition}.
\S\ref{sec:app_rq3_embedding} details the sentence embedding analyses that assess textual transparency independently of any LLM.

\subsection{LLM Item--Construct Recognition: Per-Construct Breakdown}
\label{sec:app_rq3_icr}

Table~\ref{tab:llm_recognition} in the main text reports mean $F_1$ per model and instrument.
Here we disaggregate those scores by construct to reveal which constructs are most and least recognizable.

\paragraph{BFI instruments.}
Table~\ref{tab:app_rq3_icr_bfi} shows per-construct $F_1$ for BFI-44 and BFI-10.
Large models (GPT-OSS-120B, Qwen3-235B) achieve near-perfect recognition across all five traits, while Gemma3-4B struggles most with Neuroticism and Agreeableness.
Overall, BFI items are highly transparent: even the weakest model--construct pairs on BFI-44 (Gemma3-4B on both Agreeableness and Neuroticism, tied at $F_1 = .53$) far exceed any VP result.

\begin{table*}[htbp]
\centering
\resizebox{\textwidth}{!}{%
\begin{tabular}{lrrrrrrrrrrrr}
\toprule
& \multicolumn{6}{c}{BFI-44} & \multicolumn{6}{c}{BFI-10} \\
\cmidrule(lr){2-7} \cmidrule(lr){8-13}
Model & O & C & E & A & N & $\bar{F}_1$ & O & C & E & A & N & $\bar{F}_1$ \\
\midrule
Qwen2.5-72B & 0.889 & 0.714 & 0.769 & 0.714 & 0.769 & 0.771 & 0.667 & 0.667 & 0.667 & 0.667 & 0.667 & 0.667 \\
Qwen2.5-7B & 0.750 & 0.615 & 0.667 & 0.714 & 0.769 & 0.703 & 0.667 & 0.667 & 0.667 & 0.667 & 0.667 & 0.667 \\
Qwen3-235B & \textbf{0.952} & \textbf{1.000} & 0.889 & \textbf{1.000} & 0.941 & \textbf{0.956} & \textbf{1.000} & \textbf{1.000} & \textbf{1.000} & \textbf{1.000} & \textbf{1.000} & \textbf{1.000} \\
Qwen3-30B & 0.947 & 0.947 & \textbf{1.000} & \textbf{1.000} & 0.857 & \textbf{0.950} & \textbf{1.000} & \textbf{1.000} & \textbf{1.000} & \textbf{1.000} & 0.667 & 0.933 \\
Gemma3-27B & 0.947 & 0.947 & 0.933 & 0.625 & 0.588 & 0.808 & \textbf{1.000} & \textbf{1.000} & 0.667 & 0.500 & 0.500 & 0.733 \\
Gemma3-4B & 0.842 & 0.706 & 0.769 & 0.526 & 0.526 & 0.674 & 0.667 & 0.667 & 0.667 & 0.500 & 0.400 & 0.580 \\
GPT-OSS-120B & \textbf{1.000} & 0.947 & 0.941 & \textbf{1.000} & \textbf{1.000} & \textbf{0.978} & \textbf{1.000} & \textbf{1.000} & \textbf{1.000} & \textbf{1.000} & \textbf{1.000} & \textbf{1.000} \\
\bottomrule
\end{tabular}}
\caption{Per-construct $F_1$ scores for the LLM item--construct recognition task on BFI instruments. Each cell reports the binary classification $F_1$ for a given model--construct pair. \textbf{Bold} indicates $F_1 \ge 0.95$.}
\label{tab:app_rq3_icr_bfi}
\end{table*}

\paragraph{PVQ instruments.}
Table~\ref{tab:app_rq3_icr_pvq} shows per-construct $F_1$ for PVQ-40 and PVQ-21.
Achievement and Hedonism are consistently among the most recognizable constructs on PVQ-40; most models reach $F_1 \ge .85$ for Achievement and $F_1 \ge .75$ for Hedonism.
Security, Power, and Tradition are the hardest to identify, with $F_1$ often below .60.
This pattern partly overlaps the embedding-based intra-similarity ranking in Table~\ref{tab:app_per_construct_intra}, where Stimulation and Hedonism show the tightest within-construct clustering while Security and Tradition show the weakest.

\begin{table*}[htbp]
\centering
\resizebox{\textwidth}{!}{%
\begin{tabular}{lrrrrrrrrrrrrrrrrrrrrrr}
\toprule
& \multicolumn{11}{c}{PVQ-40} & \multicolumn{11}{c}{PVQ-21} \\
\cmidrule(lr){2-12} \cmidrule(lr){13-23}
Model & PO & AC & HE & ST & SD & UN & BE & TR & CO & SE & $\bar{F}_1$ & PO & AC & HE & ST & SD & UN & BE & TR & CO & SE & $\bar{F}_1$ \\
\midrule
Qwen2.5-72B & 0.500 & \textbf{1.000} & \textbf{1.000} & 0.750 & 0.889 & 0.857 & 0.800 & 0.571 & 0.889 & 0.667 & 0.792 & 0.400 & \textbf{1.000} & \textbf{1.000} & 0.800 & 0.800 & 0.857 & \textbf{1.000} & 0.667 & \textbf{1.000} & \textbf{1.000} & 0.852 \\
Qwen2.5-7B & 0.444 & \textbf{1.000} & 0.857 & 0.857 & 0.727 & \textbf{0.923} & 0.889 & 0.571 & \textbf{1.000} & 0.600 & 0.787 & 0.500 & \textbf{1.000} & \textbf{1.000} & 0.800 & 0.667 & 0.857 & \textbf{1.000} & 0.667 & \textbf{1.000} & 0.800 & 0.829 \\
Qwen3-235B & 0.600 & 0.889 & 0.750 & 0.600 & 0.800 & 0.706 & 0.667 & 0.444 & 0.800 & 0.500 & 0.676 & 0.800 & 0.800 & 0.667 & 0.571 & 0.667 & 0.857 & 0.800 & 0.400 & 0.800 & 0.500 & 0.686 \\
Qwen3-30B & 0.600 & 0.889 & 0.750 & 0.667 & 0.727 & 0.750 & 0.727 & 0.400 & 0.800 & 0.462 & 0.677 & 0.667 & 0.800 & 0.571 & 0.667 & 0.667 & 0.857 & 0.667 & 0.400 & 0.800 & 0.571 & 0.667 \\
Gemma3-27B & 0.500 & \textbf{1.000} & 0.750 & 0.500 & 0.615 & 0.800 & 0.727 & 0.571 & 0.571 & 0.421 & 0.646 & 0.500 & \textbf{1.000} & 0.667 & 0.571 & 0.571 & 0.857 & 0.800 & 0.667 & 0.667 & 0.500 & 0.680 \\
Gemma3-4B & 0.400 & 0.727 & 0.667 & 0.462 & 0.364 & 0.632 & 0.348 & 0.533 & 0.444 & 0.276 & 0.485 & 0.444 & 0.571 & 0.571 & 0.571 & 0.364 & 0.600 & 0.400 & 0.571 & 0.500 & 0.286 & 0.488 \\
GPT-OSS-120B & 0.571 & \textbf{1.000} & \textbf{1.000} & 0.857 & 0.750 & \textbf{1.000} & 0.857 & 0.571 & 0.800 & 0.500 & 0.791 & 0.400 & \textbf{1.000} & \textbf{1.000} & 0.800 & \textbf{1.000} & \textbf{1.000} & \textbf{1.000} & 0.667 & \textbf{1.000} & 0.800 & 0.867 \\
\bottomrule
\end{tabular}}
\caption{Per-construct $F_1$ scores for the LLM item--construct recognition task on PVQ instruments. Abbreviations: PO\,=\,Power, AC\,=\,Achievement, HE\,=\,Hedonism, ST\,=\,Stimulation, SD\,=\,Self-Direction, UN\,=\,Universalism, BE\,=\,Benevolence, TR\,=\,Tradition, CO\,=\,Conformity, SE\,=\,Security. \textbf{Bold} indicates $F_1 \ge 0.90$.}
\label{tab:app_rq3_icr_pvq}
\end{table*}

\paragraph{VP items.}
Table~\ref{tab:app_rq3_icr_vp} confirms that the low recognition reported in the main text is not driven by a few hard constructs but is a pervasive property of VP scenario--response pairs.
The two constructs that come closest to consistent recognition are Agreeableness ($F_1$ from .09 to .27) and Stimulation ($F_1$ from .07 to .25); both reach $F_1 > .10$ for six of seven models but drop just below .10 on GPT-OSS-120B. Agreeableness likely benefits from interpersonal conflict---a common theme in VP's advisory scenarios---providing a weak signal for that trait, while Stimulation may pick up on VP's adventure- or novelty-themed scenarios.
All other constructs hover near zero, and no model achieves $F_1 > .27$ on any single VP construct.

\begin{table*}[htbp]
\centering
\resizebox{\textwidth}{!}{%
\begin{tabular}{lrrrrrrrrrrrrrrrr}
\toprule
& \multicolumn{10}{c}{Schwartz Values} & \multicolumn{5}{c}{BFI Traits} & \\
\cmidrule(lr){2-11} \cmidrule(lr){12-16}
Model & PO & AC & HE & ST & SD & UN & BE & TR & CO & SE & O & C & E & A & N & $\bar{F}_1$ \\
\midrule
Qwen2.5-72B & 0.000 & 0.123 & 0.057 & 0.145 & 0.082 & 0.086 & 0.063 & 0.000 & 0.022 & 0.065 & 0.028 & 0.022 & 0.000 & 0.203 & 0.000 & 0.060 \\
Qwen2.5-7B & 0.059 & 0.092 & 0.095 & 0.164 & 0.087 & 0.071 & 0.080 & 0.000 & 0.030 & 0.069 & 0.000 & 0.045 & 0.000 & 0.217 & 0.000 & 0.067 \\
Qwen3-235B & 0.252 & 0.177 & 0.149 & 0.250 & 0.066 & 0.030 & 0.065 & 0.000 & 0.055 & 0.078 & 0.049 & 0.155 & 0.093 & 0.211 & 0.014 & 0.110 \\
Qwen3-30B & 0.088 & 0.144 & 0.171 & 0.241 & 0.058 & 0.068 & 0.083 & 0.000 & 0.043 & 0.080 & 0.055 & 0.132 & 0.043 & 0.231 & 0.000 & 0.096 \\
Gemma3-27B & 0.169 & 0.159 & 0.161 & 0.224 & 0.062 & 0.048 & 0.074 & 0.083 & 0.056 & 0.064 & 0.035 & 0.222 & 0.049 & 0.264 & 0.009 & 0.112 \\
Gemma3-4B & 0.206 & 0.182 & 0.152 & 0.202 & 0.040 & 0.054 & 0.072 & 0.000 & 0.046 & 0.085 & 0.096 & 0.136 & 0.082 & 0.267 & 0.029 & 0.110 \\
GPT-OSS-120B & 0.146 & 0.137 & 0.000 & 0.071 & 0.026 & 0.080 & 0.030 & 0.000 & 0.036 & 0.073 & 0.059 & 0.048 & 0.000 & 0.094 & 0.000 & 0.053 \\
\bottomrule
\end{tabular}}
\caption{Per-construct $F_1$ scores for the LLM item--construct recognition task on VP (generation-probability) items. VP items are evaluated against all 15 constructs (10~Schwartz values + 5~BFI traits). Column abbreviations follow Tables~\ref{tab:app_rq3_icr_pvq} and~\ref{tab:app_rq3_icr_bfi}.}
\label{tab:app_rq3_icr_vp}
\end{table*}

\subsection{Sentence Embedding Analysis}
\label{sec:app_rq3_embedding}

We encode all item texts and construct definitions with a sentence transformer and ask two questions:
\begin{enumerate}
    \item \textbf{Item--definition similarity.} Is each item embedding closer to its own construct definition than to other construct definitions?
    \item \textbf{Within-construct item similarity.} Are items that share a construct more similar to each other (in embedding space) than to items from other constructs?
\end{enumerate}

\subsubsection{Sentence Encoder}
\label{sec:app_encoder}

We use \texttt{all-mpnet-base-v2} \citep{reimers-gurevych-2019-sentence} as our primary sentence encoder.
We chose a general-purpose model because our goal is to measure whether construct membership is signaled by surface-level lexical-semantic cues, not domain-specific features.
To verify that our findings are not artifacts of this particular encoder, we replicate both analyses with four additional encoders in \S\ref{sec:app_encoder_robustness}.

For VP items, the embedding analyses use a single \emph{primary} construct per item---the construct with the highest human-validated correlation (argmax over $\rho$)---as ground truth, both for item--definition top-1 accuracy and for within-construct clustering. This differs from the multi-label scheme used in the LLM recognition task (\S\ref{sec:rq3}), where every construct with $\rho \geq.30$ is treated as a positive. We use the stricter primary-construct labeling here because both embedding analyses require a single reference label per item (a top-1 target and a single clustering group), whereas the binary recognition task naturally accommodates multiple positives.

\subsubsection{Item--Definition Similarity}
\label{sec:app_item_def_sim}

For each item, we compute cosine similarity between the item embedding and the embedding of every construct definition (``\textit{Construct Name}: \textit{definition text}'').
\textit{Discrimination} is the difference between similarity to the correct construct and the mean similarity to all other constructs; \textit{Top-1 accuracy} is the fraction of items whose highest-similarity construct is the correct one.

Table~\ref{tab:app_item_def_results} shows the results.
 Established questionnaires achieve 77--81\% Top-1 accuracy and discrimination of 0.13--0.22, indicating that the item text alone reveals which construct is being measured.
 VP response texts show near-chance accuracy and near-zero discrimination, indicating that construct identity is only weakly recoverable from the response text alone.

\begin{table*}[htbp]
\centering
\begin{tabular}{lrrrr}
\toprule
Survey & Sim(correct) & Sim(others) & Discrim. & Top-1 Acc. \\
\midrule
PVQ-40   & 0.385 & 0.193 & 0.192 & 0.775 \\
PVQ-21   & 0.403 & 0.187 & 0.217 & 0.810 \\
BFI-44   & 0.347 & 0.219 & 0.128 & 0.773 \\
BFI-10   & 0.327 & 0.200 & 0.127 & 0.800 \\
\midrule
VP (PVQ values) & 0.124 & 0.122 & 0.002 & 0.112 \\
VP (BFI traits)  & 0.112 & 0.097 & 0.015 & 0.256 \\
\bottomrule
\end{tabular}
\caption{Item--definition similarity: cosine similarity between item embeddings and construct-definition embeddings. Discrimination = sim(correct) $-$ mean(sim(others)). Higher values indicate stronger lexical-semantic cues for construct identity.}
\label{tab:app_item_def_results}
\end{table*}

\subsubsection{Within-Construct Item Similarity}
\label{sec:app_within_construct_sim}

We compute pairwise cosine similarity between all item embeddings and compare:
\textit{intra-construct similarity} (mean similarity among items sharing a construct) versus
\textit{inter-construct similarity} (mean similarity among items from different constructs).
The \textit{clustering gap} is the difference (intra $-$ inter); a positive gap means items measuring the same construct are textually more similar to each other.

Table~\ref{tab:app_within_construct_results} shows the results.
Established questionnaires exhibit gaps of 0.07--0.15, meaning items within a construct share surface-level wording.
VP items show near-zero gaps ($-$0.001 to $+$0.004), indicating little difference between pairs of response texts assigned to the same versus different constructs.

\begin{table*}[htbp]
\centering
\begin{tabular}{lrrrr}
\toprule
Survey & Intra-sim & Inter-sim & Gap & $K$ \\
\midrule
PVQ-40  & 0.564 & 0.432 & 0.132 & 10 \\
PVQ-21  & 0.573 & 0.427 & 0.147 & 10 \\
BFI-44  & 0.424 & 0.328 & 0.096 & 5 \\
BFI-10  & 0.413 & 0.341 & 0.072 & 5 \\
\midrule
VP (PVQ values) & 0.106 & 0.107 & $-$0.001 & 10 \\
VP (BFI traits)  & 0.110 & 0.106 & 0.004 & 5 \\
\bottomrule
\end{tabular}
\caption{Within-construct item similarity: mean intra- vs.\ inter-construct pairwise cosine similarity. Clustering gap = intra $-$ inter. $K$ = number of constructs.}
\label{tab:app_within_construct_results}
\end{table*}

Table~\ref{tab:app_per_construct_intra} breaks down intra-similarity by individual construct for PVQ-40 and BFI-44.
Nearly all established constructs exceed their respective inter-construct baselines (PVQ-40: 0.432; BFI-44: 0.328); PVQ-40 Security (0.426) sits marginally below the baseline, and PVQ-40 Tradition (0.438) sits marginally above it by a comparable amount, so both should be read as borderline cases. Stimulation and Hedonism show the strongest within-construct similarity in PVQ-40, and Agreeableness in BFI-44.

\begin{table*}[htbp]
\centering
\begin{tabular}{lr|lr}
\toprule
\multicolumn{2}{c|}{PVQ-40} & \multicolumn{2}{c}{BFI-44} \\
\cmidrule(lr){1-2} \cmidrule(lr){3-4}
Construct & Intra-sim & Construct & Intra-sim \\
\midrule
Stimulation    & 0.724 & Agreeableness      & 0.503 \\
Hedonism       & 0.723 & Conscientiousness  & 0.453 \\
Achievement    & 0.647 & Openness           & 0.409 \\
Self-Direction & 0.631 & Neuroticism        & 0.374 \\
Conformity     & 0.613 & Extraversion       & 0.360 \\
Universalism   & 0.583 & & \\
Power          & 0.566 & & \\
Benevolence    & 0.512 & & \\
Tradition      & 0.438 & & \\
Security       & 0.426 & & \\
\bottomrule
\end{tabular}
\caption{Per-construct mean intra-similarity for PVQ-40 and BFI-44. All values exceed their inter-construct baselines (PVQ-40: 0.432; BFI-44: 0.328) except PVQ-40 Security (0.426), which sits marginally below the baseline; PVQ-40 Tradition (0.438) sits marginally above the baseline by a comparable amount and should be read as a borderline case.}
\label{tab:app_per_construct_intra}
\end{table*}

\subsubsection{Encoder Robustness Check}
\label{sec:app_encoder_robustness}

To verify that the gap between established and VP items is not an artifact of the specific sentence encoder, we replicate both analyses using five encoders spanning multiple model families (MPNet, MiniLM, BGE, and E5):
\texttt{MPNet-base} (primary),
\texttt{MiniLM-L12},
\texttt{MiniLM-L6},
\texttt{BGE-base},
and \texttt{E5-base}.\footnote{Exact HuggingFace model IDs: \texttt{sentence-transformers/all-mpnet-base-v2}, \texttt{sentence-transformers/all-MiniLM-L12-v2}, \texttt{sentence-transformers/all-MiniLM-L6-v2}, \texttt{BAAI/bge-base-en-v1.5}, and \texttt{intfloat/e5-base-v2}. For E5, we follow the model's recommended convention and prepend the \texttt{``query: ''} prefix to every item and construct definition before encoding; no prefix is added for the other four encoders.}
Table~\ref{tab:app_embedding_robustness} reports the results.
Across all five encoders, established questionnaires consistently show higher item--definition discrimination and within-construct clustering gap than VP items.
The absolute magnitudes vary with encoder quality (e.g., E5-base shows lower discrimination overall), but the qualitative pattern---established $\gg$ VP--- holds across all five encoders tested.

\begin{table*}[htbp]
\centering
\resizebox{\textwidth}{!}{%
\begin{tabular}{lrrrr|rrrr}
\toprule
 & \multicolumn{4}{c|}{Item--Definition Discrimination} & \multicolumn{4}{c}{Within-Construct Clustering Gap} \\
\cmidrule(lr){2-5} \cmidrule(lr){6-9}
Encoder & PVQ-40 & BFI-44 & VP\,(PVQ) & VP\,(BFI) & PVQ-40 & BFI-44 & VP\,(PVQ) & VP\,(BFI) \\
\midrule
MPNet-base  & 0.192 & 0.128 & 0.002 & 0.015 & +0.132 & +0.096 & $-$0.001 & +0.004 \\
MiniLM-L12  & 0.185 & 0.124 & 0.010 & 0.020 & +0.117 & +0.090 & 0.000 & +0.001 \\
MiniLM-L6   & 0.168 & 0.135 & 0.009 & 0.021 & +0.100 & +0.091 & $-$0.001 & +0.002 \\
BGE-base    & 0.088 & 0.069 & 0.002 & 0.010 & +0.055 & +0.060 & $-$0.003 & 0.000 \\
E5-base     & 0.051 & 0.034 & $-$0.002 & 0.002 & +0.031 & +0.033 & $-$0.001 & +0.001 \\
\bottomrule
\end{tabular}}
\caption{Robustness check across five sentence encoders. Left: item--definition discrimination. Right: within-construct clustering gap. The established $\gg$ VP pattern holds across the five encoders tested.}
\label{tab:app_embedding_robustness}
\end{table*}

\begin{figure*}[htbp]
\centering
\includegraphics[width=\textwidth]{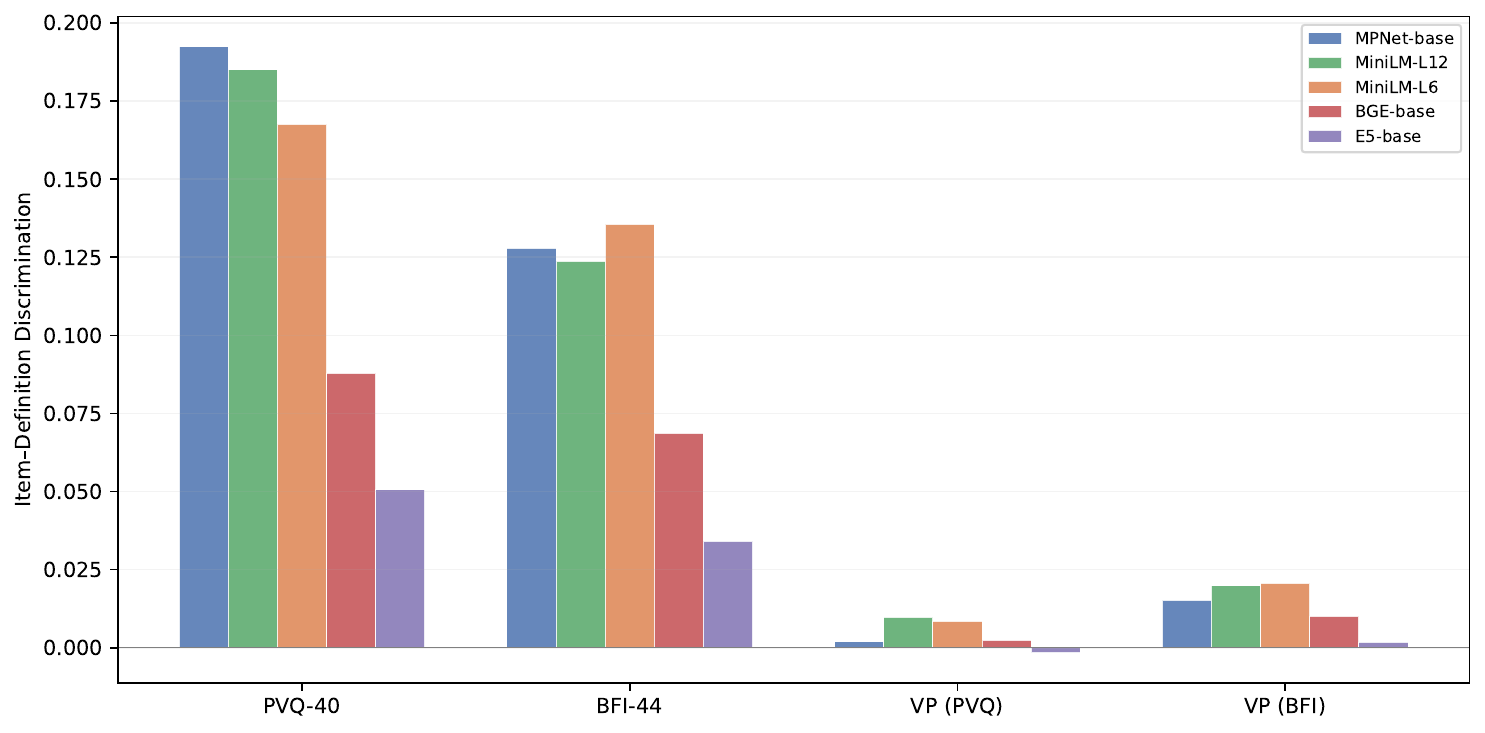}
\caption{Item--definition discrimination across five sentence encoders. Established questionnaires (PVQ-40, BFI-44) consistently show higher discrimination than VP outputs across the five encoders tested.}
\label{fig:app_robustness_cr}
\end{figure*}

\begin{figure*}[htbp]
\centering
\includegraphics[width=\textwidth]{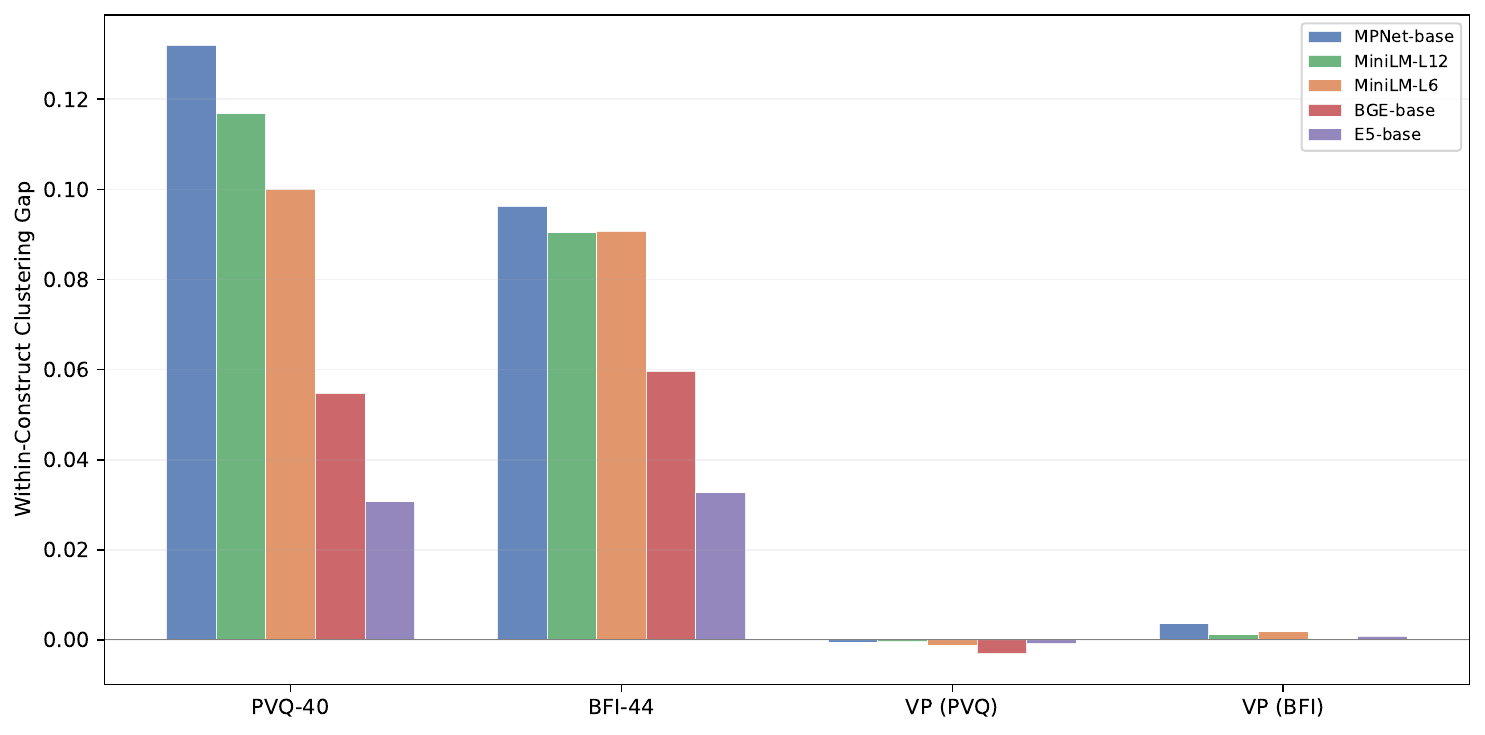}
\caption{Within-construct clustering gap across five sentence encoders. Established questionnaires show positive gaps (within-construct items are textually more similar), while VP outputs show near-zero gaps.}
\label{fig:app_robustness_lsc}
\end{figure*}

\begin{figure*}[htbp]
\centering
\includegraphics[width=\textwidth]{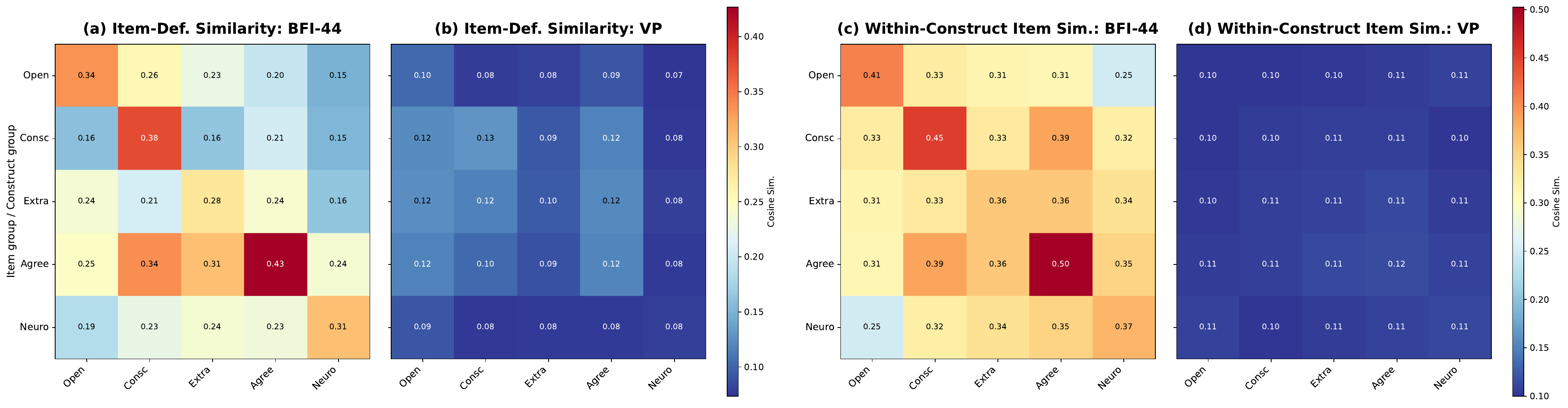}
\caption{Cosine similarity heatmaps for BFI-44 and VP BFI-trait items, parallel to Figure~\ref{fig:heatmap_pvq}. (a, b) Item--definition similarity. (c, d) Within-construct item similarity. Established items (a, c) show clear diagonal structure; VP items (b, d) do not.}
\label{fig:app_heatmap_bfi}
\end{figure*}

\FloatBarrier
\section{Construct-Cue Reduction Experiment}
\label{sec:app_cueless}

This appendix details the direct transparency manipulation summarized in \S\ref{sec:rq3}.

\paragraph{Rewriting protocol.}
We rewrote all 84 established items (40 PVQ-40 and 44 BFI-44) to reduce overt construct cues while preserving each item's intended construct, response format, and keying; all 16 reverse-keyed BFI-44 items retained their original keying. Because an item must retain some construct-relevant meaning, we call these items \emph{cue-reduced} rather than cueless and treat residual structure as expected. All item pairs appear in Tables~\ref{tab:app_cueless_pvq} and~\ref{tab:app_cueless_bfi}.

\paragraph{Manipulation checks.}
Sentence-encoder top-1 construct matching falls from .775 to .425 for PVQ-40 and from .773 to .477 for BFI-44. Unlike Table~\ref{tab:llm_recognition}, the corresponding eight-model recognition check includes GPT-OSS-20B. Mean construct-recognition $F_1$ likewise falls from .70 to .63 and from .85 to .68, respectively, toward the VP reference of .08.

\paragraph{Results.}
We re-administered the cue-reduced items to all eight models using the original questionnaire pipeline. Table~\ref{tab:app_cueless_permodel} reports $\eta^2$, WMV, and permutation $p$-values by model and condition. The mean condition-specific permutation references change little from original to cue-reduced items: $\eta^2=.230\to.231$ and WMV $=1.025\to1.025$ for PVQ-40, and $\eta^2=.094\to.093$ and WMV $=1.022\to1.023$ for BFI-44. The exact paired sign-flip tests for the mean change in $\eta^2$ give $p=.078$ for PVQ-40 and $p=.004$ for BFI-44. Cue reduction also raises questionnaire--VP agreement ($\rho=.312\to.421$ for PVQ-40; $.264\to.462$ for BFI-44); original--cue-reduced correlations average $\rho=.72$ and $.65$.

\begin{table*}[htbp]
\centering
\small
\renewcommand{\arraystretch}{0.95}
\setlength{\tabcolsep}{4.5pt}
\begin{tabular}{@{}l cccccc c cccccc@{}}
\toprule
& \multicolumn{6}{c}{\textbf{PVQ-40}} & & \multicolumn{6}{c}{\textbf{BFI-44}} \\
\cmidrule(lr){2-7} \cmidrule(l){9-14}
& \multicolumn{2}{c}{$\eta^2$} & \multicolumn{2}{c}{$p$ ($\eta^2$)} & \multicolumn{2}{c}{WMV} & & \multicolumn{2}{c}{$\eta^2$} & \multicolumn{2}{c}{$p$ ($\eta^2$)} & \multicolumn{2}{c}{WMV} \\
Model & Orig. & Cue-r. & Orig. & Cue-r. & Orig. & Cue-r. & & Orig. & Cue-r. & Orig. & Cue-r. & Orig. & Cue-r. \\
\midrule
Qwen2.5-72B  & .715 & .520 & .000 & .005 & .380 & .684 & & .733 & .306 & .000 & .004 & .320 & .781 \\
Qwen2.5-7B   & .598 & .325 & .001 & .133 & .422 & .679 & & .513 & .315 & .000 & .002 & .553 & .803 \\
Qwen3-235B   & .580 & .607 & .000 & .000 & .570 & .530 & & .323 & .290 & .004 & .010 & .803 & .828 \\
Qwen3-30B    & .339 & .506 & .096 & .008 & .862 & .630 & & .619 & .115 & .000 & .266 & .465 & 1.053 \\
Gemma3-27B   & .669 & .426 & .000 & .033 & .461 & .844 & & .519 & .134 & .000 & .221 & .565 & 1.007 \\
Gemma3-4B    & .601 & .463 & .002 & .013 & .533 & .664 & & .787 & .532 & .000 & .000 & .250 & .533 \\
GPT-OSS-120B & .510 & .322 & .002 & .168 & .577 & .734 & & .146 & .108 & .187 & .341 & .990 & 1.029 \\
GPT-OSS-20B  & .200 & .286 & .618 & .269 & 1.019 & .878 & & .300 & .128 & .006 & .241 & .791 & .990 \\
\midrule
\textbf{Mean} & .526 & .432 & --- & --- & .603 & .705 & & .492 & .241 & --- & --- & .592 & .878 \\
\bottomrule
\end{tabular}
\caption{Per-model construct structure on original (Orig.) and cue-reduced (Cue-r.) items. Each $p$ is a one-sided test against the condition-specific label-permutation reference; values below .0005 are shown as .000.}
\label{tab:app_cueless_permodel}
\end{table*}

\FloatBarrier

\section{Base vs.\ Instruction-Tuned Model Pairs}
\label{sec:app_base_instruct}

This appendix details the post-training analyses summarized in \S\ref{sec:rq1} and \S\ref{sec:rq3}.

\paragraph{Setup and comparability.}
We compare four matched pairs: Qwen2.5-7B, Gemma3-4B, Gemma3-27B, and Qwen3-30B-A3B. Within each pair, the checkpoints share pretraining and differ in post-training. Because base models cannot reliably generate free-form Likert answers, both checkpoints use the same constrained protocol: next-token probabilities over numbered options, averaged over both option orders, with reverse-keyed items recoded as in the main pipeline. Both checkpoints receive identical plain prompts without a chat template; VP scoring is unchanged. Accordingly, the constrained-scoring $\eta^2$ values reported below are not directly comparable to the main free-form results. For reference, constrained and free-form PVQ-40 profiles for the instruction-tuned checkpoints correlate at $\rho=.74$, $.88$, $.98$, and $.54$, respectively.

\paragraph{Prosocial shifts.}
Benevolence, Universalism, and Agreeableness rise after instruction tuning in 11 of 12 construct-by-pair comparisons, with a mean change of $+0.88$ Likert points (Table~\ref{tab:app_e5_prosocial}). The corresponding standardized VP scores rise in 7 cases and fall in 5, with a mean change of .00.

\begin{table}[htbp]
\centering
\small
\renewcommand{\arraystretch}{0.95}
\setlength{\tabcolsep}{4pt}
\begin{tabular}{@{}lccc@{}}
\toprule
& \multicolumn{2}{c}{PVQ-40} & BFI-44 \\
\cmidrule(lr){2-3} \cmidrule(l){4-4}
Pair & Benev. & Univ. & Agree. \\
\midrule
Qwen2.5-7B    & $+0.003$ & $-0.09$ & $+0.24$ \\
Gemma3-4B     & $+2.22$ & $+2.14$ & $+0.76$ \\
Gemma3-27B    & $+1.06$ & $+0.83$ & $+0.90$ \\
Qwen3-30B-A3B & $+1.05$ & $+1.06$ & $+0.34$ \\
\midrule
\textbf{Mean} & $+1.08$ & $+0.98$ & $+0.56$ \\
\bottomrule
\end{tabular}
\caption{Base-to-instruction change in prosocial questionnaire scores (Likert points; constrained scoring).}
\label{tab:app_e5_prosocial}
\end{table}

\paragraph{Two-by-two design.}
On original PVQ-40 items, $\eta^2$ rises in every pair (mean gain $+.196$); on cue-reduced items, the mean gain is $-.006$ and is smaller in all four pairs (one-sided sign-flip $p=.0625$, the minimum with four pairs). For BFI-44, gains are inconsistent, and the corresponding contrast is positive in 2/4 pairs ($p=.44$).

\begin{table*}[htbp]
\centering
\small
\renewcommand{\arraystretch}{0.95}
\setlength{\tabcolsep}{4.5pt}
\begin{tabular}{@{}l cccccc c cccccc@{}}
\toprule
& \multicolumn{6}{c}{\textbf{PVQ-40 $\eta^2$}} & & \multicolumn{6}{c}{\textbf{BFI-44 $\eta^2$}} \\
\cmidrule(lr){2-7} \cmidrule(l){9-14}
& \multicolumn{2}{c}{Original} & \multicolumn{2}{c}{Cue-reduced} & \multicolumn{2}{c}{Gain} & & \multicolumn{2}{c}{Original} & \multicolumn{2}{c}{Cue-reduced} & \multicolumn{2}{c}{Gain} \\
Pair & Base & Instr. & Base & Instr. & Orig. & Cue-r. & & Base & Instr. & Base & Instr. & Orig. & Cue-r. \\
\midrule
Qwen2.5-7B    & .485 & .642 & .476 & .367 & $+.157$ & $-.109$ & & .381 & .435 & .291 & .308 & $+.054$ & $+.017$ \\
Gemma3-4B     & .194 & .523 & .457 & .546 & $+.329$ & $+.089$ & & .171 & .131 & .044 & .087 & $-.040$ & $+.043$ \\
Gemma3-27B    & .490 & .748 & .463 & .449 & $+.258$ & $-.014$ & & .276 & .552 & .140 & .108 & $+.277$ & $-.032$ \\
Qwen3-30B-A3B & .499 & .540 & .417 & .425 & $+.040$ & $+.008$ & & .686 & .324 & .294 & .149 & $-.362$ & $-.144$ \\
\midrule
\textbf{Mean} & .417 & .613 & .453 & .447 & $+.196$ & $-.006$ & & .379 & .361 & .192 & .163 & $-.018$ & $-.029$ \\
\bottomrule
\end{tabular}
\caption{Checkpoint (base vs.\ instruction) crossed with item wording (original vs.\ cue-reduced), under constrained scoring. Gain is instruction minus base.}
\label{tab:app_e7_2x2}
\end{table*}

\paragraph{Questionnaire--VP agreement.}
Agreement does not rise consistently after post-training: value-profile $\rho$ changes from a mean of .318 to .339 (two pairs up and two down), and trait-profile $\rho$ from .525 to .275 (none up, two down, and two unchanged).

\begin{table}[htbp]
\centering
\small
\renewcommand{\arraystretch}{0.95}
\setlength{\tabcolsep}{4pt}
\begin{tabular}{@{}lcccc@{}}
\toprule
& \multicolumn{2}{c}{Values} & \multicolumn{2}{c}{Traits} \\
\cmidrule(lr){2-3} \cmidrule(l){4-5}
Pair & Base & Instr. & Base & Instr. \\
\midrule
Qwen2.5-7B    & $.30$ & $.47$ & $.40$ & $.40$ \\
Gemma3-4B     & $-.32$ & $.13$ & $.60$ & $.30$ \\
Gemma3-27B    & $.67$ & $.21$ & $.60$ & $-.10$ \\
Qwen3-30B-A3B & $.62$ & $.55$ & $.50$ & $.50$ \\
\midrule
\textbf{Mean} & $.318$ & $.339$ & $.525$ & $.275$ \\
\bottomrule
\end{tabular}
\caption{Questionnaire--VP profile agreement (Spearman $\rho$, summed scoring) before and after post-training; values use PVQ-40 and traits use BFI-44.}
\label{tab:app_e5_agreement}
\end{table}

\paragraph{Post-training signature in VP.}
For each matched pair and measure, we standardize the base and instruction ten-value profiles separately, then subtract the base profile from the instruction profile to obtain $\Delta$VP or $\Delta$PVQ. All six cross-pair cosines among $\Delta$VP vectors are positive (.61, .72, .72, .79, .86, .97; mean $.78$), but $\Delta$VP and $\Delta$PVQ do not align consistently within pairs (mean cosine $-.10$, range $-.30$ to $+.14$). Thus, VP carries a consistent post-training signature across the full value profile, but it does not consistently mirror the questionnaire shift.

\begin{table}[htbp]
\centering
\small
\renewcommand{\arraystretch}{0.95}
\setlength{\tabcolsep}{4pt}
\begin{tabular}{@{}lccc@{}}
\toprule
Pair & VP $\rho$ & PVQ $\rho$ & $\cos(\Delta\mathrm{VP},\Delta\mathrm{PVQ})$ \\
\midrule
Qwen2.5-7B    & .94 & .87 & $+.14$ \\
Gemma3-4B     & .73 & .15 & $-.09$ \\
Gemma3-27B    & .62 & .88 & $-.14$ \\
Qwen3-30B-A3B & .85 & .85 & $-.30$ \\
\bottomrule
\end{tabular}
\caption{Profile stability (base--instruction Spearman $\rho$) and within-pair cosine alignment between $\Delta$VP and $\Delta$PVQ.}
\label{tab:app_e5_signature}
\end{table}

\clearpage
\onecolumn

\begin{footnotesize}
\begin{longtable}{@{}p{0.4cm}p{1.9cm}p{6.2cm}p{6.2cm}@{}}

\toprule
\# & Value & Original item & Cue-reduced item \\
\midrule
\endfirsthead
\multicolumn{4}{@{}l}{\footnotesize Table~\ref{tab:app_cueless_pvq} (continued)}\\
\toprule
\# & Value & Original item & Cue-reduced item \\
\midrule
\endhead
\bottomrule
\noalign{\vskip 6pt}
\caption{All 40 PVQ-40 items: original wording (gender-neutral form used in the main experiments) and the cue-reduced rewriting used in Appendix~\ref{sec:app_cueless}. Rewritings preserve the target value, response format, and keying.}\label{tab:app_cueless_pvq}\\
\endfoot

1 & Self-Direction & Thinking up new ideas and being creative is important to them. They like to do things in their own original way. & When they take on a task, they would rather work out their own way of doing it than follow how it was done before. They often sketch out possibilities that nobody around them has suggested. \\
\addlinespace[0pt]
2 & Power & It is important to them to be rich. They want to have a lot of money and expensive things. & It matters to them that their bank balance keeps growing year after year. They want the kind of house, car, and watch that most people could never afford. \\
\addlinespace[0pt]
3 & Universalism & They think it is important that every person in the world be treated equally. They believe everyone should have equal opportunities in life. & It bothers them when a child born in one neighborhood gets chances that a child born a few streets away will never see. They think the same doors should open for everyone, everywhere. \\
\addlinespace[0pt]
4 & Achievement & It's very important to them to show their abilities. They want people to admire what they do. & When they finish a difficult piece of work, they make sure the people who matter get to see it. Hearing others talk about what they have pulled off makes their day. \\
\addlinespace[0pt]
5 & Security & It is important to them to live in secure surroundings. They avoid anything that might endanger their safety. & They double-check the locks before bed and prefer streets where nothing much ever happens. They stay away from shortcuts through areas they don't know. \\
\addlinespace[0pt]
6 & Stimulation & They think it is important to do lots of different things in life. They always look for new things to try. & Their weekends rarely look the same twice --- one month a pottery class, the next a road trip. If a menu lists a dish they can't pronounce, that's the one they order. \\
\addlinespace[0pt]
7 & Conformity & They believe that people should do what they're told. They think people should follow rules at all times, even when no-one is watching. & They stop at the red light at an empty intersection at 3 a.m. and wait for it to change. When a supervisor gives an instruction, they carry it out without arguing. \\
\addlinespace[0pt]
8 & Universalism & It is important to them to listen to people who are different from them. Even when they disagree with them, they still want to understand them. & At a gathering, they sit down next to the person whose views clash with their own and ask questions instead of walking away. They want to hear the full story before making up their mind about anyone. \\
\addlinespace[0pt]
9 & Tradition & They think it's important not to ask for more than what you have. They believe that people should be satisfied with what they have. & Once their needs are met, they stop chasing a bigger house or a better job title. In their eyes, someone who keeps wanting more has missed the point. \\
\addlinespace[0pt]
10 & Hedonism & They seek every chance they can to have fun. It is important to them to do things that give them pleasure. & If there is a party on Friday and a long lunch on Sunday, they will be at both. Given a free afternoon, they spend it on whatever feels best in the moment. \\
\addlinespace[0pt]
11 & Self-Direction & It is important to them to make their own decisions about what they do. They like to be free to plan and to choose their activities for themselves. & They would turn down a well-paid job if someone else got to set their schedule. Nobody fills in their calendar but them. \\
\addlinespace[0pt]
12 & Benevolence & It's very important to them to help the people around them. They want to care for their well-being. & When a friend is moving apartments or stranded at the airport, they are the first phone call. They keep an eye on how the people in their circle are holding up. \\
\addlinespace[0pt]
13 & Achievement & Being very successful is important to them. They like to impress other people. & They want their name near the top when the results are posted. It matters to them that others raise their eyebrows at what they have done. \\
\addlinespace[0pt]
14 & Security & It is very important to them that their country be safe. They think the state must be on watch against threats from within and without. & They follow news about border incidents closely and want more spending on the army and the police. Groups that could stir up trouble from within make them uneasy. \\
\addlinespace[0pt]
15 & Stimulation & They like to take risks. They are always looking for adventures. & They sign up for the cliff jump while the others watch from the shore. A plan with an uncertain ending draws them more than one with a guaranteed outcome. \\
\addlinespace[0pt]
16 & Conformity & It is important to them always to behave properly. They want to avoid doing anything people would say is wrong. & Before doing anything, they ask themselves what the people around them would make of it. They would rather pass up something they wanted than give the neighbors a reason to talk. \\
\addlinespace[0pt]
17 & Power & It is important to them to be in charge and tell others what to do. They want people to do what they say. & In any group project, they end up assigning the tasks and setting the deadlines. Things go smoother, they feel, when everyone else simply follows their plan. \\
\addlinespace[0pt]
18 & Benevolence & It is important to them to be loyal to their friends. They want to devote themselves to people close to them. & If someone they have known for years calls at midnight, they get in the car. The handful of people they grew up with can count on them for anything. \\
\addlinespace[0pt]
19 & Universalism & They strongly believe that people should care for nature. Looking after the environment is important to them. & They sort their recycling, cycle instead of driving when they can, and give money to groups that replant forests. A riverbank covered in plastic bags genuinely upsets them. \\
\addlinespace[0pt]
20 & Tradition & Religious belief is important to them. They try hard to do what their religion requires. & They set aside time every week to attend services, and they keep the fasts and observances they were raised with. Before big decisions, they pray. \\
\addlinespace[0pt]
21 & Security & It is important to them that things be organized and clean. They really do not like things to be a mess. & Every jar in their kitchen has a label, and every shirt in the closet faces the same way. A sink of dishes left overnight bothers them more than they would like to admit. \\
\addlinespace[0pt]
22 & Self-Direction & They think it's important to be interested in things. They like to be curious and to try to understand all sorts of things. & An offhand question about how bridges stay up can cost them an evening of reading. They ask 'why' more often than most adults they know. \\
\addlinespace[0pt]
23 & Universalism & They believe all the worlds' people should live in harmony. Promoting peace among all groups in the world is important to them. & News of a ceasefire anywhere on the globe makes their week. They give to organizations that bring together groups who have been fighting for generations. \\
\addlinespace[0pt]
24 & Achievement & They think it is important to be ambitious. They want to show how capable they are. & They set targets that make other people whistle, and then chase them down. When a hard problem has beaten everyone else, they volunteer --- partly so everyone can watch them solve it. \\
\addlinespace[0pt]
25 & Tradition & They think it is best to do things in traditional ways. It is important to them to keep up the customs they have learned. & They cook the dishes their grandmother cooked, the same way, on the same dates every year. If something has been done a certain way for generations, they see no reason to change it. \\
\addlinespace[0pt]
26 & Hedonism & Enjoying life's pleasures is important to them. They like to 'spoil' themselves. & They book the massage, order the dessert, and run the long bath without a second thought. Treating themselves well is a fixed line in their budget, not an afterthought. \\
\addlinespace[0pt]
27 & Benevolence & It is important to them to respond to the needs of others. They try to support those they know. & When someone in their circle is short on rent or buried in work, they quietly step in. People around them have learned that asking once is enough. \\
\addlinespace[0pt]
28 & Conformity & They believe they should always show respect to their parents and to older people. It is important to them to be obedient. & They still answer their father with 'sir' and would never contradict a grandparent at the table. When someone senior asks them to do something, they do it without argument. \\
\addlinespace[0pt]
29 & Universalism & They want everyone to be treated justly, even people they don't know. It is important to them to protect the weak in society. & They speak up when a stranger is being shortchanged, not only when it happens to a friend. They give time and money to shelters and legal-aid clinics for people who cannot push back on their own. \\
\addlinespace[0pt]
30 & Stimulation & They like surprises. It is important to them to have an exciting life. & They would rather open an unmarked envelope than be told what is inside. A year in which nothing unexpected happened would feel like a year lost. \\
\addlinespace[0pt]
31 & Security & They try hard to avoid getting sick. Staying healthy is very important to them. & They book every recommended check-up, keep sanitizer in the car, and go to bed early when something is going around. Their medicine cabinet is stocked before anyone sneezes. \\
\addlinespace[0pt]
32 & Achievement & Getting ahead in life is important to them. They strive to do better than others. & They keep track of where they stand against the people they started out with --- pay, title, results --- and second place stings. Each year they aim to be a rung above last year. \\
\addlinespace[0pt]
33 & Benevolence & Forgiving people who have hurt them is important to them. They try to see what is good in them and not to hold a grudge. & After a friend let them down badly, they still showed up at his birthday a month later. They would rather remember the good years than replay one bad evening. \\
\addlinespace[0pt]
34 & Self-Direction & It is important to them to be independent. They like to rely on themselves. & They fixed the leaking pipe with a video tutorial rather than call their brother-in-law. Handing their affairs over to someone else makes them uncomfortable. \\
\addlinespace[0pt]
35 & Security & Having a stable government is important to them. They are concerned that the social order be protected. & They vote for candidates who promise continuity rather than upheaval. Strikes, sudden reforms, and street protests make them uneasy about where the country is heading. \\
\addlinespace[0pt]
36 & Conformity & It is important to them to be polite to other people all the time. They try never to disturb or irritate others. & They apologize when someone else bumps into them, and keep their music low even at midday. They would take the longer hallway rather than walk between two people talking. \\
\addlinespace[0pt]
37 & Hedonism & They really want to enjoy life. Having a good time is very important to them. & Their calendar fills with dinners, festivals, and weekends away before anything else gets a slot. If an evening ends with cheeks sore from laughing, it counted. \\
\addlinespace[0pt]
38 & Tradition & It is important to them to be humble and modest. They try not to draw attention to themselves. & When their project wins, they nudge a teammate forward to collect the prize. They pick the seat at the back of the room and change the subject when the conversation turns to them. \\
\addlinespace[0pt]
39 & Power & They always want to be the one who makes the decisions. They like to be the leader. & When the group cannot settle on a restaurant, they name one, and everyone goes. Committees they join tend to end up with them holding the gavel --- which suits them fine. \\
\addlinespace[0pt]
40 & Universalism & It is important to them to adapt to nature and to fit into it. They believe that people should not change nature. & They would rather a footpath bend around an old oak than the oak be cut for a straighter path. When they camp, they leave the clearing exactly as they found it. \\
\addlinespace[0pt]
\end{longtable}
\end{footnotesize}

\begin{footnotesize}
\begin{longtable}{@{}p{0.4cm}p{2.4cm}p{5.85cm}p{5.85cm}@{}}
\toprule
\# & Trait & Original item & Cue-reduced item \\
\midrule
\endfirsthead
\multicolumn{4}{@{}l}{\footnotesize Table~\ref{tab:app_cueless_bfi} (continued)}\\
\toprule
\# & Trait & Original item & Cue-reduced item \\
\midrule
\endhead
\bottomrule
\noalign{\vskip 6pt}
\caption{All 44 BFI-44 items (stem: ``I see myself as someone who\ldots''): original wording and the cue-reduced rewriting used in Appendix~\ref{sec:app_cueless}. (R) marks the 16 reverse-keyed items, whose keying is preserved in the rewriting.}\label{tab:app_cueless_bfi}\\
\endfoot
1 & Extraversion & is talkative. & keeps a conversation going long after others have run out of things to say. \\
\addlinespace[0pt]
2 & Agreeableness (R) & tends to find fault with others. & usually notices what a colleague got wrong before noticing what they got right. \\
\addlinespace[0pt]
3 & Conscientiousness & does a thorough job. & reads a document down to the last page before signing off on it. \\
\addlinespace[0pt]
4 & Neuroticism & is depressed, blue. & has stretches where getting out of bed takes real effort and little seems worth doing. \\
\addlinespace[0pt]
5 & Openness & is original, comes up with new ideas. & suggests approaches in meetings that nobody else at the table had thought of. \\
\addlinespace[0pt]
6 & Extraversion (R) & is reserved. & lets others do most of the talking at gatherings. \\
\addlinespace[0pt]
7 & Agreeableness & is helpful and unselfish with others. & gives up a free weekend to move a friend's furniture without being asked twice. \\
\addlinespace[0pt]
8 & Conscientiousness (R) & can be somewhat careless. & sometimes sends the email before noticing the attachment is missing. \\
\addlinespace[0pt]
9 & Neuroticism (R) & is relaxed, handles stress well. & sleeps soundly the night before a big exam or presentation. \\
\addlinespace[0pt]
10 & Openness & is curious about many different things. & will happily lose an afternoon reading about topics that have nothing to do with their work. \\
\addlinespace[0pt]
11 & Extraversion & is full of energy. & is still going strong when the rest of the group is ready to head home. \\
\addlinespace[0pt]
12 & Agreeableness (R) & starts quarrels with others. & turns small disagreements over dinner into raised voices. \\
\addlinespace[0pt]
13 & Conscientiousness & is a reliable worker. & delivers what was promised on the day it was promised. \\
\addlinespace[0pt]
14 & Neuroticism & can be tense. & sits with a clenched jaw and tight shoulders when things are up in the air. \\
\addlinespace[0pt]
15 & Openness & is ingenious, a deep thinker. & keeps turning a question over long after everyone else has settled for the easy answer. \\
\addlinespace[0pt]
16 & Extraversion & generates a lot of enthusiasm. & can get a whole team eager to start on a plan they were lukewarm about an hour earlier. \\
\addlinespace[0pt]
17 & Agreeableness & has a forgiving nature. & meets a friend for coffee a week after that friend let them down badly. \\
\addlinespace[0pt]
18 & Conscientiousness (R) & tends to be disorganized. & hunts for keys, chargers, and paperwork nearly every morning. \\
\addlinespace[0pt]
19 & Neuroticism & worries a lot. & lies awake replaying tomorrow's appointments and what could go wrong with each. \\
\addlinespace[0pt]
20 & Openness & has an active imagination. & can spend a commute inventing whole life stories for the strangers across the aisle. \\
\addlinespace[0pt]
21 & Extraversion (R) & tends to be quiet. & can share a long car ride without feeling any need to fill the silence. \\
\addlinespace[0pt]
22 & Agreeableness & is generally trusting. & takes people's stories at face value unless given a clear reason not to. \\
\addlinespace[0pt]
23 & Conscientiousness (R) & tends to be lazy. & leaves the laundry until there is nothing left to wear. \\
\addlinespace[0pt]
24 & Neuroticism (R) & is emotionally stable, not easily upset. & takes a canceled flight or a harsh comment without letting it ruin the day. \\
\addlinespace[0pt]
25 & Openness & is inventive. & rigs a working fix out of spare parts when the proper tool is missing. \\
\addlinespace[0pt]
26 & Extraversion & has an assertive personality. & says plainly at the table what they want and where they stand. \\
\addlinespace[0pt]
27 & Agreeableness (R) & can be cold and aloof. & answers a coworker's warm greeting with a short nod and keeps walking. \\
\addlinespace[0pt]
28 & Conscientiousness & perseveres until the task is finished. & stays at the desk until the last box on the list is ticked, even past midnight. \\
\addlinespace[0pt]
29 & Neuroticism & can be moody. & can go from cheerful at breakfast to snappish by lunch without an obvious reason. \\
\addlinespace[0pt]
30 & Openness & values artistic, aesthetic experiences. & will cross town on a weeknight to catch an exhibition or hear a string quartet. \\
\addlinespace[0pt]
31 & Extraversion (R) & is sometimes shy, inhibited. & rehearses a sentence twice before saying it to a stranger at a party. \\
\addlinespace[0pt]
32 & Agreeableness & is considerate and kind to almost everyone. & remembers the doorman's name and asks how his daughter's exams went. \\
\addlinespace[0pt]
33 & Conscientiousness & does things efficiently. & clears in one afternoon what takes others most of a week. \\
\addlinespace[0pt]
34 & Neuroticism (R) & remains calm in tense situations. & was the one giving directions when the fire alarm went off. \\
\addlinespace[0pt]
35 & Openness (R) & prefers work that is routine. & is happiest when today's tasks look exactly like yesterday's. \\
\addlinespace[0pt]
36 & Extraversion & is outgoing, sociable. & leaves a party with three new phone numbers. \\
\addlinespace[0pt]
37 & Agreeableness (R) & is sometimes rude to others. & sometimes cuts the waiter off mid-sentence with a sharp remark. \\
\addlinespace[0pt]
38 & Conscientiousness & makes plans and follows through with them. & books the flights in January and is standing at the gate in June, itinerary in hand. \\
\addlinespace[0pt]
39 & Neuroticism & gets nervous easily. & feels their heart race when the phone rings from an unknown number. \\
\addlinespace[0pt]
40 & Openness & likes to reflect, play with ideas. & can chew on a hypothetical question for the length of a dinner. \\
\addlinespace[0pt]
41 & Openness (R) & has few artistic interests. & walks past the gallery to get to the food court. \\
\addlinespace[0pt]
42 & Agreeableness & likes to cooperate with others. & would rather split the work and share the credit than keep both to themselves. \\
\addlinespace[0pt]
43 & Conscientiousness (R) & is easily distracted. & opens the laptop to file a report and surfaces an hour later, twelve tabs deep, report untouched. \\
\addlinespace[0pt]
44 & Openness & is sophisticated in art, music, or literature. & can tell you which translation of a novel to pick and why one recording of a symphony outshines another. \\
\addlinespace[0pt]
\end{longtable}
\end{footnotesize}

\twocolumn

\newpage
\clearpage

\section{RQ4 Supplementary Materials}
\label{sec:app_rq4}

\subsection{Persona Prompt Templates}
\label{sec:app_rq4_prompts}

Table~\ref{tab:rq4_persona_prompts} lists the system-prompt wording used for each of the eight demographic persona conditions in RQ4. Each prompt is used as the model's system message when obtaining either Likert-scale responses (PVQ-40, PVQ-21) or VP candidate-response probabilities. The vanilla (no-persona) baseline is the same configuration used in RQ1.

\begin{table*}[ht]
\centering
\small
\setlength{\tabcolsep}{5pt}
\begin{tabular}{@{}lll@{}}
\toprule
Category & Cond. & System prompt \\
\midrule
\multirow{2}{*}{Gender}
 & A (Male) & ``You are a male. Please think, behave, and talk accordingly.'' \\
 & B (Female) & ``You are a female. Please think, behave, and talk accordingly.'' \\
\midrule
\multirow{2}{*}{Age}
 & A (20--39) & ``You are 20 to 39 years old. Please think, behave, and talk accordingly.'' \\
 & B (80+) & ``You are 80 years old or older. Please think, behave, and talk accordingly.'' \\
\midrule
\multirow{2}{*}{Political}
 & A (Right) & ``In political matters, you consider yourself right-wing. Please think, behave, and talk accordingly.'' \\
 & B (Left) & ``In political matters, you consider yourself left-wing. Please think, behave, and talk accordingly.'' \\
\midrule
\multirow{2}{*}{Education}
 & A (Below) & ``Your highest educational level is below university. Please think, behave, and talk accordingly.'' \\
 & B (Univ.+) & ``Your highest educational level is university or above. Please think, behave, and talk accordingly.'' \\
\bottomrule
\end{tabular}
\caption{Persona prompt wording for each demographic condition (RQ4). All prompts are injected as the system message; the user-turn instruction and survey items remain identical to RQ1.}
\label{tab:rq4_persona_prompts}
\end{table*}

\subsection{Methodology Details}
\label{sec:app_rq4_method}

\paragraph{ESS data.}
Human reference profiles are drawn from the European Social Survey (ESS) Round~11, which covers 29 European countries and Israel. After filtering for complete PVQ responses and demographic variables, the working sample comprises $N = 37{,}398$ respondents. Table~\ref{tab:app_rq4_ess_n} shows per-condition sample sizes. The ESS measures Schwartz values via a 21-item Portrait Values Questionnaire~\cite{schwartz2003ess}; it does not include a Big Five personality instrument, which is why RQ4 is restricted to value dimensions.

\begin{table}[ht]
\centering
\small
\begin{tabular}{llr}
\toprule
Category & Condition & $N$ \\
\midrule
\multicolumn{2}{l}{\textit{Full sample}} & 37{,}398 \\
\midrule
\multirow{2}{*}{Gender}     & Male               & 17{,}537 \\
                             & Female             & 19{,}861 \\
\midrule
\multirow{2}{*}{Age}        & 20--39 years old   & 9{,}283 \\
                             & 80+ years old      & 2{,}225 \\
\midrule
\multirow{2}{*}{Political}  & Right-wing         & 11{,}537 \\
                             & Left-wing          & 11{,}152 \\
\midrule
\multirow{2}{*}{Education}  & Below university   & 27{,}054 \\
                             & University or above& 10{,}164 \\
\bottomrule
\end{tabular}
\caption{ESS sample sizes by demographic condition.}
\label{tab:app_rq4_ess_n}
\end{table}

Political orientation is mapped from the ESS 0--10 left--right scale: ``Right-wing'' includes responses coded as Right, Center-right, Far right, Very right, and Center-leaning right; ``Left-wing'' is the symmetric counterpart. Education uses the ES-ISCED harmonized categories: levels I--IV map to ``Below university,'' levels V1--V2 to ``University or above.'' We do not include a Religion condition because the ESS does not provide a comparable religion-based value profile.

\paragraph{Ipsative centering.}
Following standard Schwartz methodology~\cite{schwartz2003ess}, we center each respondent's (or model run's) profile by subtracting the individual's grand mean across all ten value dimensions:
\begin{equation}
\text{centered}_v = \bar{x}_v - \frac{1}{10}\sum_{k=1}^{10} \bar{x}_k
\label{eq:centering}
\end{equation}
This removes individual-level scale-use bias (acquiescence). On the human side, centering is applied per respondent before aggregation; on the LLM side, the analogous procedure is applied to each model's raw profile. After centering, the ten values sum to zero within each profile.

\paragraph{Delta computation.}
The LLM delta for value $v$ under demographic condition $c$ is:
\begin{equation}
\Delta_{v,c}^{\text{LLM}} = \text{centered}_v^{\,\text{persona}(c)} - \text{centered}_v^{\,\text{vanilla}}
\label{eq:delta_llm}
\end{equation}
where the vanilla (no-persona) profile comes from the RQ1 baseline (\S\ref{sec:rq1}).
The human reference delta is the analogous difference computed from ESS respondents:
\begin{equation}
\Delta_{v,c}^{\text{Human}} = \overline{\text{centered}}_v^{\,\text{subgroup}(c)} - \overline{\text{centered}}_v^{\,\text{full\;sample}}
\label{eq:delta_human}
\end{equation}
LLM deltas are computed per model and then averaged across all seven models. We compute deltas separately for PVQ-40, PVQ-21, and VP generation probabilities.

\paragraph{Cosine similarity.}
The cosine similarity between the 10-dimensional human and LLM delta vectors provides a scale-invariant measure of directional alignment:
\begin{equation}
\cos(\boldsymbol{\Delta}^{\text{Human}},\, \boldsymbol{\Delta}^{\text{LLM}}) = \frac{\boldsymbol{\Delta}^{\text{Human}} \cdot \boldsymbol{\Delta}^{\text{LLM}}}{\|\boldsymbol{\Delta}^{\text{Human}}\| \;\|\boldsymbol{\Delta}^{\text{LLM}}\|}
\label{eq:cosine}
\end{equation}
Values near $+1$ indicate that the LLM shifts in the same direction as the human demographic effect; values near $-1$ indicate opposite shifts. Statistical significance is assessed via bootstrap confidence intervals (\S\ref{sec:app_rq4_bootstrap}) and permutation tests.

\paragraph{Direction match.}
For each condition, the direction match counts the number of value dimensions (out of 10) where the human and LLM deltas share the same algebraic sign:
\begin{equation}
\begin{aligned}
\text{DirMatch}(c)
&= \sum_{v=1}^{10} \mathbf{1}\!\left[
\operatorname{sign}(\Delta_{v,c}^{\text{Human}})\right. \\
&\hspace{25mm}\left.{}= \operatorname{sign}(\Delta_{v,c}^{\text{LLM}})
\right]
\end{aligned}
\label{eq:dirmatch}
\end{equation}
Under the null hypothesis of no association, each dimension has a 50\% chance of matching, so $\text{DirMatch}(c) \sim \text{Binomial}(10, 0.5)$. We aggregate across all 8 conditions (80 dimensions total) and test with a one-sided binomial test ($H_1$: match rate $> 50\%$) at the conventional significance threshold $\alpha = 0.05$.

\paragraph{ Between-value normalized magnitude.}
To compare delta magnitudes across Likert and log-probability scales, we compute a between-value normalized magnitude:
\begin{equation}
\text{}\;M_{\mathrm{norm}} = \frac{\overline{|\Delta_v|}}{\sigma_{\text{baseline}}}
\label{eq:effect_size}
\end{equation}
where $\sigma_{\text{baseline}}$ is the standard deviation of the baseline (vanilla) profile \emph{across the ten value dimensions of a single profile}. This normalizes each shift by the intrinsic spread of the source's value profile and is what makes the ratio comparable across Likert and log-probability scales. We caution that, unlike a conventional Cohen's $d$, the denominator here is a between-value (within-profile) spread rather than an across-subject (between-respondent or between-model) SD, so the resulting quantity should be read as a relative magnitude on each source's own value-profile scale rather than as a noise-normalized effect size in the classical sense.

\paragraph{Model consensus.}
For each (condition $\times$ value dimension) pair, we record the proportion of the seven models that agree on the direction of shift. A pair is classified as \emph{strong consensus} when $\ge 6$ of 7 models agree ($\ge 85.7\%$) and \emph{weak consensus} when $<5$ of 7 agree ($<71.4\%$); pairs with exactly $5/7$ agreement (71.4\%) fall in an intermediate category that is neither strong nor weak. Models with a zero (no-shift) sign on a given dimension are counted as part of the negative-direction group when assessing agreement.

\subsection{Direction Match Statistical Tests}
\label{sec:app_direction_match}

We test whether the observed direction agreement between human demographic effects and LLM persona shifts is significantly above chance (50\%) using a one-sided binomial test.

\paragraph{PVQ-40 (Likert, 40 items).}
Overall direction agreement: 62/80 (77.5\%). Binomial test ($H_0$: $p = 0.5$, $H_1$: $p > 0.5$): $p = 4.07\times10^{-7}$.

\paragraph{PVQ-21 (Likert, 21 items).}
Overall direction agreement: 55/80 (68.8\%). Binomial test ($H_0$: $p = 0.5$, $H_1$: $p > 0.5$): $p = 5.26\times10^{-4}$.

\paragraph{VP (generation-probability).}
Overall direction agreement: 40/80 (50.0\%). Binomial test ($H_0$: $p = 0.5$, $H_1$: $p > 0.5$): $p = 5.44\times10^{-1}$. The VP generation-probability deltas show no significant directional agreement with human demographic effects, consistent with the near-zero aggregate cosine similarity reported in the main text.

Table~\ref{tab:app_direction_binomial} provides the per-condition breakdown for all three survey types.

\begin{table*}[htbp]
\centering
\begin{tabular}{lrrrrrr}
\toprule
 & \multicolumn{2}{c}{PVQ-40} & \multicolumn{2}{c}{PVQ-21} & \multicolumn{2}{c}{VP} \\
\cmidrule(lr){2-3} \cmidrule(lr){4-5} \cmidrule(lr){6-7}
Condition & Match & $p$ & Match & $p$ & Match & $p$ \\
\midrule
Male & 7/10 & 0.172 & 5/10 & 0.623 & 3/10 & 0.945 \\
Female & 6/10 & 0.377 & 7/10 & 0.172 & 7/10 & 0.172 \\
20–39 years old & 10/10 & \textbf{0.001} & 7/10 & 0.172 & 2/10 & 0.989 \\
80+ years old & 9/10 & \textbf{0.011} & 8/10 & 0.055 & 7/10 & 0.172 \\
Right-wing & 7/10 & 0.172 & 7/10 & 0.172 & 2/10 & 0.989 \\
Left-wing & 10/10 & \textbf{0.001} & 9/10 & \textbf{0.011} & 9/10 & \textbf{0.011} \\
Below university & 8/10 & 0.055 & 9/10 & \textbf{0.011} & 4/10 & 0.828 \\
University or above & 5/10 & 0.623 & 3/10 & 0.945 & 6/10 & 0.377 \\
\midrule
\textbf{Overall} & \textbf{62/80} & \makecell{\\$\mathbf{4.07\times10^{-7}}$} & \textbf{55/80} & \makecell{\\$\mathbf{5.26\times10^{-4}}$} & \textbf{40/80} & 0.544 \\
\bottomrule
\end{tabular}
\caption{Direction agreement by condition with binomial test $p$-values ($H_0$: agreement = 50\%, one-sided). Bold indicates $p < 0.05$.}
\label{tab:app_direction_binomial}
\end{table*}

\subsection{Per-Condition and Per-Model Results}
\label{sec:app_rq4_results}

\subsubsection{Cosine Similarity by Condition}
\label{sec:app_rq4_cosine_cond}

Table~\ref{tab:app_rq4_cosine_cond} reports the cosine similarity between the human and LLM delta vectors for each condition and survey type. These values underlie Table~\ref{tab:rq4_cosine} of the main text. PVQ-40 Likert achieves positive cosine similarity in all eight conditions, PVQ-21 in seven of eight, while VP generation probabilities show mixed signs with four positive and four negative conditions.

\begin{table*}[htbp]
\centering
\begin{tabular}{lrrrrrr}
\toprule
 & \multicolumn{2}{c}{PVQ-40 Likert} & \multicolumn{2}{c}{PVQ-21 Likert} & \multicolumn{2}{c}{VP Gen-Prob} \\
\cmidrule(lr){2-3} \cmidrule(lr){4-5} \cmidrule(lr){6-7}
Condition & Cos & Dir. & Cos & Dir. & Cos & Dir. \\
\midrule
Male & +0.675 & 7 & +0.417 & 5 & $-$0.589 & 3 \\
Female & +0.056 & 6 & +0.129 & 7 & +0.617 & 7 \\
20–39 years old & +0.946 & 10 & +0.600 & 7 & $-$0.259 & 2 \\
80+ years old & +0.891 & 9 & +0.825 & 8 & +0.049 & 7 \\
Right-wing & +0.697 & 7 & +0.687 & 7 & $-$0.685 & 2 \\
Left-wing & +0.865 & 10 & +0.780 & 9 & +0.654 & 9 \\
Below university & +0.528 & 8 & +0.711 & 9 & $-$0.398 & 4 \\
University or above & +0.105 & 5 & $-$0.422 & 3 & +0.347 & 6 \\
\midrule
\textbf{Mean} & \textbf{+0.595} &  & \textbf{+0.466} &  & \textbf{$-$0.033} &  \\
\bottomrule
\end{tabular}
\caption{Cosine similarity between human and LLM delta vectors by condition and survey type. Dir.\,= direction match out of 10 value dimensions.}
\label{tab:app_rq4_cosine_cond}
\end{table*}

\subsubsection{Per-Model Cosine Similarity}
\label{sec:app_rq4_per_model_cos}

Table~\ref{tab:app_rq4_per_model_cos} shows the aggregate cosine similarity between human and LLM deltas, computed separately for each model. All models show positive PVQ-40--human cosine, ranging from $+.184$ to $+.600$. VP--human cosine ranges from $-.101$ to $+.196$ and is negative for four of seven models, so the weak aggregate alignment is not driven by a single outlier.

\begin{table*}[htbp]
\centering
\begin{tabular}{lrrr}
\toprule
Model & PVQ-40--Human & VP--Human & PVQ-40--VP \\
\midrule
gemma-3-27b-it & +0.600 & $-$0.101 & $-$0.399 \\
gpt-oss-20b & +0.473 & $-$0.079 & $-$0.050 \\
gpt-oss-120b & +0.204 & +0.032 & $-$0.512 \\
Qwen2.5-7B-Inst. & +0.184 & +0.164 & $-$0.119 \\
Qwen2.5-72B-Inst. & +0.558 & +0.196 & +0.303 \\
Qwen3-30B-A3B-Inst. & +0.426 & $-$0.079 & $-$0.107 \\
Qwen3-235B-A22B-Inst. & +0.541 & $-$0.045 & $-$0.106 \\
\midrule
\textbf{Mean} & \textbf{+0.426} & \textbf{+0.012} & \textbf{$-$0.141} \\
\bottomrule
\end{tabular}
\caption{Per-model cosine similarity between LLM deltas and human deltas. For each model, the eight per-condition 10-dimensional deltas are concatenated into a single 80-dimensional vector, and one cosine is computed against the analogously concatenated human delta; means are then taken across the seven models. This concatenated form is sensitive to large-magnitude dimensions and therefore differs from the per-condition cosine averages reported in Table~\ref{tab:rq4_cosine} of the main text (PVQ-40 mean +0.60 vs.\ +0.43 here). PVQ-40--VP shows the direct agreement between Likert and generation-probability deltas under the same concatenated procedure.}
\label{tab:app_rq4_per_model_cos}
\end{table*}

\subsubsection{Cross-Model Consensus}
\label{sec:app_rq4_consensus}

Table~\ref{tab:app_rq4_consensus} summarizes the degree to which different models agree on the direction of persona-induced shifts. For each of the 80 (condition, value) pairs, we compute the fraction of models sharing the majority sign of $\Delta$.

\begin{table*}[htbp]
\centering
\begin{tabular}{lrrr}
\toprule
Metric & PVQ-40 & PVQ-21 & VP \\
\midrule
Mean consensus & 0.732 & 0.720 & 0.666 \\
Strong consensus ($\ge$ 6/7) & 29/80 (36.3\%) & 25/80 (31.3\%) & 14/80 (17.5\%) \\
Weak consensus ($<$ 5/7) & 30/80 (37.5\%) & 33/80 (41.3\%) & 44/80 (55.0\%) \\
\bottomrule
\end{tabular}
\caption{Cross-model consensus on delta direction. ``Strong'' = $\ge 6/7$ models agree; ``Weak'' = $< 5/7$ agree. Percentages are over all 80 (condition $\times$ value) pairs. Per-model directions are taken from $\operatorname{sign}(\Delta)$ with zero (no-shift) values counted as part of the negative-direction group; rows with exactly $5/7$ agreement fall in the intermediate (neither-strong-nor-weak) category and therefore do not appear in either count.}
\label{tab:app_rq4_consensus}
\end{table*}

\subsubsection{Bootstrap Confidence Intervals and Permutation Tests}
\label{sec:app_rq4_bootstrap}

We assess the statistical reliability of the cosine between the concatenated human deltas and model-averaged LLM deltas via two complementary tests.

\paragraph{Bootstrap confidence intervals.}
We construct 95\% bootstrap confidence intervals by resampling the eight conditions with replacement (10{,}000 iterations). For each bootstrap sample, we concatenate the human deltas and model-averaged LLM deltas across the resampled conditions and compute their cosine similarity. Table~\ref{tab:app_rq4_bootstrap} reports the results.

\paragraph{Permutation tests.}
We also conduct permutation tests (10{,}000 iterations) under the null hypothesis that value labels are interchangeable within each condition. In each iteration, we independently permute the 10 value dimensions within each condition's model-averaged LLM delta, concatenate the permuted vectors, and recompute the cosine with the concatenated human delta.

\begin{table*}[htbp]
\centering
\begin{tabular}{lrrrrr}
\toprule
 & \multicolumn{3}{c}{Bootstrap} & \multicolumn{2}{c}{Permutation} \\
\cmidrule(lr){2-4} \cmidrule(lr){5-6}
Survey & Observed $\cos$ & 95\% CI & $P(\cos < 0)$ & Null mean (SD) & $p$ (two-sided) \\
\midrule
PVQ-40 Likert & +0.586 & [+0.337, +0.841] & 0.0\% & +0.001 (0.137) & $<$ 0.0001 \\
PVQ-21 Likert & +0.497 & [+0.250, +0.742] & 0.0\% & $-$0.001 (0.130) & $<$ 0.0001 \\
VP Gen-Prob & +0.007 & [$-$0.221, +0.236] & 47.8\% & +0.002 (0.113) & 0.957 \\
\bottomrule
\end{tabular}
\caption{Bootstrap 95\% confidence intervals and permutation test results for aggregate cosine similarity (Human vs.\ LLM average delta). The bootstrap resamples the eight demographic conditions with replacement, whereas the permutation test shuffles the ten value labels within each condition (10{,}000 iterations each).}
\label{tab:app_rq4_bootstrap}
\end{table*}

\subsubsection{ Between-Value Normalized Magnitude}
\label{sec:app_rq4_effect_size}

Table~\ref{tab:app_rq4_es} reports the between-value normalized magnitude (mean $|\Delta|$ divided by baseline standard deviation) across all conditions and value dimensions.

\begin{table}[htbp]
\centering
\small
\begin{tabular}{lrr}
\toprule
Source & \makecell{\\Mean $M_{\mathrm{norm}}$} & \makecell{\\Median $M_{\mathrm{norm}}$} \\
\midrule
PVQ-40 Likert & 0.665 & 0.424 \\
PVQ-21 Likert & 0.711 & 0.474 \\
VP Gen-Prob & 0.373 & 0.170 \\
Human (ESS) & 0.202 & 0.125 \\
\bottomrule
\end{tabular}
\caption{Between-value normalized magnitude: $M_{\mathrm{norm}}$ $= \overline{|\Delta|} / \sigma_{\text{baseline}}$, where $\sigma_{\text{baseline}}$ is the within-profile standard deviation across the ten value dimensions of the baseline. Computed across all 80 (condition $\times$ value) pairs for LLM sources, and 80 pairs for human. This is not a Cohen's $d$: the denominator is a between-value (within-profile) spread, not an across-subject SD.}
\label{tab:app_rq4_es}
\end{table}

\subsection{Value-Level Delta Profiles}
\label{sec:app_rq4_delta}
Table~\ref{tab:app_rq4_delta_with_vp} shows the per-value deltas for each demographic condition, comparing human survey effects with LLM Likert shifts (PVQ-40, PVQ-21) and VP generation-probability shifts.
Note that the table uses two different scales: Human, PVQ-40 Likert, and PVQ-21 Likert rows report raw centered-mean deltas on their native scale, whereas VP Gen-Prob rows report L2-normalized deltas (unit vectors), since absolute log-probability magnitudes are not commensurable with Likert scales. Within-row sign and relative ordering remain meaningful in both cases, but cross-row magnitudes are not directly comparable; readers should interpret VP rows as direction-only fingerprints rather than as effect-size estimates.

\begin{table*}[htbp]
\centering
\resizebox{\textwidth}{!}{%
\begin{tabular}{llrrrrrrrrrr}
\toprule
Category & Source & Ach & Ben & Con & Hed & Pow & Sec & SD & Sti & Tra & Uni \\
\midrule
\multirow{8}{*}{Gender} & Human (Male) & +0.06 & -0.07 & +0.01 & +0.05 & +0.07 & -0.10 & +0.01 & +0.09 & -0.06 & -0.07 \\
 & PVQ-40 Likert (Male) & +0.18 & -0.47 & -0.32 & +0.21 & +0.31 & +0.27 & -0.02 & +0.66 & -0.29 & -0.53 \\
 & PVQ-21 Likert (Male) & -0.05 & +0.02 & -0.52 & +0.41 & +0.27 & +0.30 & -0.09 & +0.41 & -0.27 & -0.49 \\
 & VP Gen-Prob (Male) & -0.53 & +0.59 & -0.04 & -0.04 & -0.11 & -0.06 & +0.26 & -0.31 & -0.17 & +0.40 \\
 & Human (Female) & -0.06 & +0.07 & -0.00 & -0.05 & -0.07 & +0.08 & -0.01 & -0.08 & +0.05 & +0.06 \\
 & PVQ-40 Likert (Female) & -0.21 & +0.07 & -0.02 & +0.26 & -0.40 & +0.04 & +0.11 & +0.26 & -0.12 & +0.00 \\
 & PVQ-21 Likert (Female) & -0.30 & +0.27 & -0.30 & +0.16 & -0.09 & +0.13 & -0.01 & +0.27 & +0.02 & -0.16 \\
 & VP Gen-Prob (Female) & -0.33 & +0.51 & -0.09 & -0.18 & -0.05 & -0.17 & +0.21 & -0.44 & -0.03 & +0.57 \\
\midrule
\multirow{8}{*}{Age} & Human (20--39) & +0.22 & -0.06 & -0.25 & +0.24 & +0.09 & -0.18 & +0.02 & +0.30 & -0.29 & -0.09 \\
 & PVQ-40 Likert (20--39) & +0.43 & -0.33 & -0.50 & +0.40 & +0.18 & -0.02 & +0.00 & +0.59 & -0.55 & -0.21 \\
 & PVQ-21 Likert (20--39) & -0.06 & +0.19 & -0.70 & +0.37 & +0.08 & +0.30 & +0.16 & +0.26 & -0.27 & -0.33 \\
 & VP Gen-Prob (20--39) & -0.55 & +0.54 & +0.05 & -0.04 & -0.14 & +0.03 & +0.32 & -0.18 & -0.37 & +0.33 \\
 & Human (80+) & -0.23 & +0.03 & +0.53 & -0.38 & -0.08 & +0.31 & -0.18 & -0.52 & +0.46 & +0.07 \\
 & PVQ-40 Likert (80+) & -1.49 & +0.28 & +0.83 & -0.53 & -0.27 & +1.33 & -0.24 & -1.27 & +1.46 & -0.09 \\
 & PVQ-21 Likert (80+) & -1.56 & +0.76 & +0.62 & -0.38 & +0.12 & +1.48 & -0.31 & -1.67 & +1.33 & -0.39 \\
 & VP Gen-Prob (80+) & -0.42 & +0.17 & -0.00 & -0.15 & -0.18 & +0.14 & +0.63 & -0.19 & -0.37 & +0.38 \\
\midrule
\multirow{8}{*}{Political} & Human (Right-wing) & +0.01 & -0.05 & +0.09 & +0.01 & +0.06 & -0.02 & -0.03 & +0.00 & +0.06 & -0.12 \\
 & PVQ-40 Likert (Right-wing) & +1.27 & -1.53 & +0.58 & -0.99 & +2.13 & +1.55 & -0.26 & -0.87 & +0.60 & -2.47 \\
 & PVQ-21 Likert (Right-wing) & +0.45 & -1.27 & +1.23 & -0.77 & +2.30 & +2.30 & -0.27 & -1.20 & +0.13 & -2.92 \\
 & VP Gen-Prob (Right-wing) & -0.16 & +0.48 & -0.19 & +0.04 & -0.11 & -0.48 & +0.39 & -0.07 & -0.33 & +0.44 \\
 & Human (Left-wing) & -0.03 & +0.09 & -0.14 & +0.06 & -0.10 & -0.09 & +0.10 & +0.06 & -0.13 & +0.17 \\
 & PVQ-40 Likert (Left-wing) & -0.95 & +1.05 & -0.51 & +0.18 & -0.13 & -0.83 & +0.44 & +0.53 & -0.81 & +1.03 \\
 & PVQ-21 Likert (Left-wing) & -1.57 & +1.29 & -0.64 & +0.08 & +0.04 & -0.92 & +0.76 & +0.29 & -0.60 & +1.27 \\
 & VP Gen-Prob (Left-wing) & -0.12 & +0.45 & -0.04 & +0.11 & -0.04 & -0.59 & +0.49 & -0.26 & -0.24 & +0.23 \\
\midrule
\multirow{8}{*}{Education} & Human (Below Univ.) & -0.02 & -0.02 & +0.06 & -0.02 & +0.01 & +0.06 & -0.07 & -0.05 & +0.08 & -0.05 \\
 & PVQ-40 Likert (Below Univ.) & -0.19 & -0.14 & +0.26 & +0.05 & +0.15 & +0.01 & -0.17 & +0.10 & +0.03 & -0.10 \\
 & PVQ-21 Likert (Below Univ.) & -0.71 & -0.14 & +0.33 & +0.22 & +0.33 & +0.50 & -0.28 & -0.07 & +0.22 & -0.39 \\
 & VP Gen-Prob (Below Univ.) & -0.33 & +0.48 & +0.15 & -0.28 & -0.13 & -0.14 & +0.13 & -0.05 & -0.42 & +0.57 \\
 & Human (Univ.+) & +0.06 & +0.04 & -0.17 & +0.05 & -0.04 & -0.17 & +0.18 & +0.13 & -0.22 & +0.13 \\
 & PVQ-40 Likert (Univ.+) & +0.15 & +0.00 & +0.13 & -0.09 & -0.05 & +0.17 & -0.10 & +0.19 & -0.32 & -0.08 \\
 & PVQ-21 Likert (Univ.+) & -0.01 & -0.04 & -0.04 & -0.11 & +0.10 & +0.28 & -0.04 & -0.11 & -0.04 & +0.02 \\
 & VP Gen-Prob (Univ.+) & -0.36 & +0.54 & +0.13 & -0.35 & -0.19 & -0.11 & +0.46 & -0.13 & -0.29 & +0.29 \\
\bottomrule
\end{tabular}}
\caption{Centered-mean deltas ($\Delta$): Human demographic effect vs.\ LLM persona shift (PVQ-40 Likert, PVQ-21 Likert, and VP generation-probability). Human, PVQ-40, and PVQ-21 rows are raw centered-mean deltas on their native scales; VP Gen-Prob rows are L2-normalized to unit length because raw log-probability magnitudes are not commensurable with Likert scales. Cross-row magnitudes are therefore not directly comparable; within-row signs and relative ordering are.}
\label{tab:app_rq4_delta_with_vp}
\end{table*}

\subsection{Per-Model Delta Tables}
\label{sec:app_rq4_per_model}

Tables~\ref{tab:app_rq4_per_model_pvq} and~\ref{tab:app_rq4_per_model_PVQ-21} show the per-model deltas for PVQ-40 and PVQ-21 respectively, allowing inspection of individual model behavior. Throughout this section and the cross-model variance table that follows, model names are abbreviated for display: ``\texttt{Qwen2.5-7B-Inst.}'' and ``\texttt{Qwen2.5-72B-Inst.}'' correspond to the official \texttt{-Instruct} releases, and ``\texttt{Qwen3-30B-A3B-Inst.}'' and ``\texttt{Qwen3-235B-A22B-Inst.}'' correspond to the \texttt{-Instruct-2507} (and, for the 235B model, \texttt{-FP8}) checkpoints used throughout the paper.

\begin{table*}[htbp]
\centering
\resizebox{\textwidth}{!}{%
\begin{tabular}{llrrrrrrrrrr}
\toprule
Condition & Model & Ach & Ben & Con & Hed & Pow & Sec & SD & Sti & Tra & Uni \\
\midrule
\multirow{7}{*}{Male} & gemma-3-27b-it & +0.28 & -0.47 & +0.53 & +0.11 & +1.45 & +0.58 & -0.22 & +0.11 & -0.97 & -1.39 \\
 & gpt-oss-20b & -0.53 & -0.15 & -1.03 & +0.60 & +0.26 & -0.20 & +0.35 & +1.10 & -0.65 & +0.26 \\
 & gpt-oss-120b & -0.03 & -1.03 & -0.78 & +0.31 & -0.36 & +0.27 & +0.10 & +2.14 & +0.72 & -1.36 \\
 & Qwen2.5-7B-Inst. & -0.09 & -0.34 & -0.09 & +0.16 & +0.16 & +0.06 & +0.53 & +0.16 & -0.09 & -0.43 \\
 & Qwen2.5-72B-Inst. & +0.87 & -0.63 & -0.38 & +0.00 & +0.83 & +0.00 & -0.38 & +0.66 & +0.12 & -1.09 \\
 & Qwen3-30B-A3B-Inst. & -0.08 & -0.21 & -0.21 & -0.08 & -0.25 & +0.12 & -0.08 & -0.08 & +0.54 & +0.33 \\
 & Qwen3-235B-A22B-Inst. & +0.81 & -0.44 & -0.31 & +0.40 & +0.06 & +1.06 & -0.44 & +0.56 & -1.69 & -0.02 \\
\midrule
\multirow{7}{*}{Female} & gemma-3-27b-it & -0.37 & +0.63 & +0.01 & +0.55 & -0.62 & +0.48 & -0.12 & +0.05 & -0.49 & -0.12 \\
 & gpt-oss-20b & -0.71 & -0.21 & -0.46 & +0.87 & -0.29 & -0.36 & +0.67 & +0.71 & -1.08 & +0.87 \\
 & gpt-oss-120b & +0.12 & -0.38 & +0.37 & +0.58 & -0.42 & -0.25 & +0.50 & +0.42 & -0.13 & -0.83 \\
 & Qwen2.5-7B-Inst. & -0.25 & +1.00 & -0.25 & -0.34 & +0.00 & -0.10 & +0.62 & +0.00 & -0.25 & -0.42 \\
 & Qwen2.5-72B-Inst. & -0.12 & +0.13 & +0.00 & -0.12 & -0.12 & +0.28 & -0.37 & +0.21 & +0.25 & -0.12 \\
 & Qwen3-30B-A3B-Inst. & -0.05 & -0.17 & -0.05 & -0.05 & -0.88 & +0.15 & -0.05 & -0.05 & +0.45 & +0.70 \\
 & Qwen3-235B-A22B-Inst. & -0.11 & -0.49 & +0.27 & +0.35 & -0.49 & +0.12 & -0.49 & +0.52 & +0.39 & -0.07 \\
\midrule
\multirow{7}{*}{20--39 years old} & gemma-3-27b-it & +0.07 & -0.19 & -0.31 & +0.36 & +1.19 & -0.21 & -0.31 & +0.36 & -0.31 & -0.64 \\
 & gpt-oss-20b & -0.10 & -0.22 & -1.10 & +0.61 & +0.28 & -0.32 & +0.65 & +1.11 & -1.10 & +0.19 \\
 & gpt-oss-120b & +0.75 & -1.50 & -0.88 & +1.54 & -0.29 & +0.77 & -0.25 & +1.54 & -0.13 & -1.54 \\
 & Qwen2.5-7B-Inst. & -0.25 & +0.75 & -0.38 & -0.34 & +0.00 & -0.30 & +0.50 & +0.00 & -0.38 & +0.41 \\
 & Qwen2.5-72B-Inst. & +1.24 & -0.63 & -0.38 & +0.32 & +0.32 & -0.01 & -0.01 & +0.66 & -1.01 & -0.51 \\
 & Qwen3-30B-A3B-Inst. & +0.00 & +0.00 & -0.25 & +0.00 & -0.17 & -0.20 & +0.00 & +0.00 & -0.13 & +0.75 \\
 & Qwen3-235B-A22B-Inst. & +1.33 & -0.55 & -0.17 & +0.29 & -0.05 & +0.16 & -0.55 & +0.46 & -0.80 & -0.13 \\
\midrule
\multirow{7}{*}{80+ years old} & gemma-3-27b-it & -1.75 & +0.88 & +1.63 & -0.37 & +0.13 & +1.63 & -0.75 & -2.37 & +1.50 & -0.54 \\
 & gpt-oss-20b & -1.78 & +1.09 & +0.34 & -0.70 & -0.53 & +1.17 & +0.59 & -0.87 & +0.22 & +0.47 \\
 & gpt-oss-120b & -0.57 & -0.57 & -0.32 & +0.39 & -0.61 & +2.15 & -0.45 & +0.22 & +0.80 & -1.03 \\
 & Qwen2.5-7B-Inst. & +0.24 & -0.51 & +0.24 & +0.12 & +0.62 & +0.32 & -0.76 & +0.28 & +0.24 & -0.80 \\
 & Qwen2.5-72B-Inst. & -1.32 & +0.56 & +1.18 & -1.15 & -0.32 & +1.38 & -0.19 & -1.82 & +2.06 & -0.40 \\
 & Qwen3-30B-A3B-Inst. & -3.95 & +0.55 & +0.55 & +0.55 & -1.11 & +1.25 & +0.55 & -1.78 & +2.05 & +1.30 \\
 & Qwen3-235B-A22B-Inst. & -1.32 & -0.07 & +2.18 & -2.57 & -0.07 & +1.43 & -0.69 & -2.57 & +3.31 & +0.35 \\
\midrule
\multirow{7}{*}{Right-wing} & gemma-3-27b-it & +0.54 & -0.58 & +1.04 & -1.29 & +2.21 & +1.44 & -0.46 & -1.46 & +0.92 & -2.37 \\
 & gpt-oss-20b & +0.57 & -1.18 & -0.43 & -0.77 & +2.40 & +0.37 & +0.82 & -0.27 & -0.31 & -1.18 \\
 & gpt-oss-120b & +0.51 & -1.49 & +0.39 & +0.39 & -0.61 & +2.39 & +0.39 & +0.05 & +0.76 & -2.78 \\
 & Qwen2.5-7B-Inst. & +0.31 & -1.69 & +0.06 & -0.90 & +2.77 & +1.93 & +0.18 & -0.73 & +0.31 & -2.23 \\
 & Qwen2.5-72B-Inst. & +2.94 & -2.43 & +0.94 & -1.14 & +2.69 & +1.79 & -0.68 & -0.97 & +0.32 & -3.47 \\
 & Qwen3-30B-A3B-Inst. & -0.14 & -0.64 & +0.99 & -0.14 & +2.70 & +0.56 & -0.14 & -1.64 & +0.24 & -1.80 \\
 & Qwen3-235B-A22B-Inst. & +4.19 & -2.69 & +1.06 & -3.11 & +2.73 & +2.36 & -1.94 & -1.10 & +1.94 & -3.44 \\
\midrule
\multirow{7}{*}{Left-wing} & gemma-3-27b-it & -0.94 & +1.93 & -0.69 & -0.15 & -0.32 & -0.22 & +1.18 & +0.35 & -1.82 & +0.68 \\
 & gpt-oss-20b & -0.86 & +1.14 & -0.61 & +0.10 & -0.57 & -0.53 & +0.89 & +0.27 & -1.61 & +1.77 \\
 & gpt-oss-120b & +0.06 & -0.06 & +0.69 & -0.48 & -0.31 & -0.41 & +0.06 & +0.36 & +0.06 & +0.02 \\
 & Qwen2.5-7B-Inst. & -0.47 & +1.78 & -0.59 & -0.55 & -0.22 & -0.52 & +0.03 & -0.22 & -0.59 & +1.36 \\
 & Qwen2.5-72B-Inst. & -0.39 & +0.86 & -0.89 & +0.48 & +0.48 & -1.22 & -0.14 & +0.65 & -0.39 & +0.57 \\
 & Qwen3-30B-A3B-Inst. & -3.44 & +1.06 & -0.31 & +1.06 & -0.60 & -1.04 & +1.06 & +1.06 & -0.69 & +1.81 \\
 & Qwen3-235B-A22B-Inst. & -0.64 & +0.61 & -1.14 & +0.78 & +0.61 & -1.89 & -0.01 & +1.28 & -0.64 & +1.03 \\
\midrule
\multirow{7}{*}{Below university} & gemma-3-27b-it & -0.24 & -0.12 & +1.38 & +0.34 & +0.34 & +0.51 & -0.99 & -0.66 & +0.51 & -1.08 \\
 & gpt-oss-20b & -0.76 & -0.26 & -0.51 & +0.07 & -0.09 & -0.16 & +0.99 & +0.91 & -0.51 & +0.32 \\
 & gpt-oss-120b & +0.09 & -0.79 & +0.46 & -0.29 & -0.12 & -0.19 & +0.09 & +1.54 & -0.16 & -0.62 \\
 & Qwen2.5-7B-Inst. & +0.18 & +0.31 & -0.07 & +0.01 & +0.18 & +0.08 & -0.57 & +0.18 & -0.07 & -0.24 \\
 & Qwen2.5-72B-Inst. & +0.04 & -0.33 & +0.04 & +0.29 & +0.13 & +0.29 & -0.46 & +0.12 & +0.42 & -0.54 \\
 & Qwen3-30B-A3B-Inst. & +0.03 & -0.35 & -0.22 & +0.03 & +0.03 & -0.27 & +0.03 & +0.03 & +0.28 & +0.44 \\
 & Qwen3-235B-A22B-Inst. & -0.66 & +0.59 & +0.72 & -0.07 & +0.59 & -0.21 & -0.28 & -1.41 & -0.28 & +1.01 \\
\midrule
\multirow{7}{*}{University or above} & gemma-3-27b-it & -0.12 & +0.01 & +0.01 & -0.32 & +0.01 & +0.41 & +0.01 & +0.34 & -0.37 & +0.01 \\
 & gpt-oss-20b & +0.32 & +0.45 & +0.07 & +0.20 & -0.30 & -0.10 & +0.20 & +0.03 & -1.55 & +0.70 \\
 & gpt-oss-120b & +0.21 & -0.54 & +1.21 & -0.38 & +0.12 & -0.24 & -0.42 & +0.79 & +0.46 & -1.21 \\
 & Qwen2.5-7B-Inst. & +0.00 & +0.38 & -0.25 & +0.00 & +0.00 & +0.00 & +0.00 & +0.00 & -0.13 & +0.00 \\
 & Qwen2.5-72B-Inst. & -0.19 & +0.31 & -0.06 & -0.06 & -0.23 & +0.44 & +0.06 & +0.10 & -0.06 & -0.31 \\
 & Qwen3-30B-A3B-Inst. & -0.14 & -0.14 & -0.02 & -0.14 & +0.53 & +0.06 & -0.14 & -0.14 & -0.14 & +0.28 \\
 & Qwen3-235B-A22B-Inst. & +0.93 & -0.44 & -0.07 & +0.06 & -0.44 & +0.66 & -0.44 & +0.22 & -0.44 & -0.03 \\
\bottomrule
\end{tabular}}
\caption{Per-model centered-mean deltas ($\Delta$) for PVQ-40: individual model shifts under persona prompting. All cell values are rounded to two decimals using round-half-up.}
\label{tab:app_rq4_per_model_pvq}
\end{table*}

\begin{table*}[htbp]
\centering
\resizebox{\textwidth}{!}{%
\begin{tabular}{llrrrrrrrrrr}
\toprule
Condition & Model & Ach & Ben & Con & Hed & Pow & Sec & SD & Sti & Tra & Uni \\
\midrule
\multirow{7}{*}{Male} & gemma-3-27b-it & +0.53 & -0.23 & -0.73 & +0.03 & +1.28 & +0.78 & -0.23 & +0.28 & -0.48 & -1.23 \\
 & gpt-oss-20b & -0.79 & -0.04 & -0.79 & +0.21 & +0.71 & +0.71 & -0.54 & -0.04 & -0.04 & +0.63 \\
 & gpt-oss-120b & -1.01 & +0.24 & -1.26 & +1.74 & -0.76 & -0.26 & +1.24 & +1.49 & -0.01 & -1.43 \\
 & Qwen2.5-7B-Inst. & +0.02 & +1.02 & +0.02 & +0.02 & +0.02 & -0.48 & +0.02 & +0.02 & +0.02 & -0.65 \\
 & Qwen2.5-72B-Inst. & +0.87 & -0.63 & +0.12 & +0.12 & +0.62 & +0.12 & -0.88 & +0.12 & +0.12 & -0.55 \\
 & Qwen3-30B-A3B-Inst. & +0.00 & +0.00 & -0.25 & +0.00 & +0.25 & +0.00 & +0.00 & +0.00 & +0.00 & +0.00 \\
 & Qwen3-235B-A22B-Inst. & +0.03 & -0.23 & -0.73 & +0.78 & -0.23 & +1.28 & -0.23 & +1.03 & -1.48 & -0.23 \\
\midrule
\multirow{7}{*}{Female} & gemma-3-27b-it & -0.18 & +0.33 & +0.33 & +0.08 & -0.18 & +0.83 & -0.68 & +0.33 & -0.68 & -0.18 \\
 & gpt-oss-20b & -0.39 & +0.86 & -0.64 & -0.14 & +0.36 & +0.86 & +0.36 & -0.39 & -0.89 & +0.03 \\
 & gpt-oss-120b & -0.82 & -0.07 & -0.82 & +0.68 & -0.32 & -0.57 & -0.32 & +1.43 & +0.93 & -0.15 \\
 & Qwen2.5-7B-Inst. & -0.14 & +0.86 & +0.11 & -0.14 & -0.14 & -0.64 & +1.11 & -0.39 & -0.14 & -0.48 \\
 & Qwen2.5-72B-Inst. & -0.08 & +0.18 & -0.08 & -0.08 & -0.08 & -0.08 & -0.33 & -0.08 & +0.68 & -0.08 \\
 & Qwen3-30B-A3B-Inst. & -0.03 & -0.03 & -0.28 & -0.03 & -0.03 & -0.03 & -0.03 & -0.03 & +0.48 & -0.03 \\
 & Qwen3-235B-A22B-Inst. & -0.48 & -0.23 & -0.73 & +0.78 & -0.23 & +0.53 & -0.23 & +1.03 & -0.23 & -0.23 \\
\midrule
\multirow{7}{*}{20--39 years old} & gemma-3-27b-it & +0.07 & -0.18 & -0.68 & +0.32 & +0.82 & +0.57 & -0.18 & +0.32 & -0.18 & -0.85 \\
 & gpt-oss-20b & -0.03 & +0.23 & -1.78 & +0.48 & +0.23 & +0.73 & -0.03 & +0.23 & -0.53 & +0.48 \\
 & gpt-oss-120b & -0.37 & +0.38 & -1.87 & +1.38 & -1.37 & +0.63 & +0.88 & +0.88 & +0.63 & -1.20 \\
 & Qwen2.5-7B-Inst. & -0.23 & +1.28 & -0.23 & -0.23 & -0.23 & -0.48 & +1.03 & -0.48 & -0.23 & -0.23 \\
 & Qwen2.5-72B-Inst. & +0.64 & -0.11 & +0.14 & -0.11 & +0.14 & -0.11 & -0.36 & +0.39 & -0.36 & -0.28 \\
 & Qwen3-30B-A3B-Inst. & -0.08 & -0.08 & -0.08 & -0.08 & +1.18 & -0.08 & -0.08 & -0.08 & -0.58 & -0.08 \\
 & Qwen3-235B-A22B-Inst. & -0.43 & -0.18 & -0.43 & +0.83 & -0.18 & +0.83 & -0.18 & +0.58 & -0.68 & -0.18 \\
\midrule
\multirow{7}{*}{80+ years old} & gemma-3-27b-it & -1.40 & +0.10 & +2.60 & -0.40 & +0.60 & +2.10 & -1.15 & -2.90 & +1.35 & -0.90 \\
 & gpt-oss-20b & -1.48 & +1.77 & -0.23 & -1.23 & +0.02 & +2.52 & -1.23 & -1.48 & +1.27 & +0.10 \\
 & gpt-oss-120b & -0.95 & +1.30 & -1.45 & +0.05 & -0.45 & +1.80 & +0.30 & -1.20 & +2.05 & -1.45 \\
 & Qwen2.5-7B-Inst. & +0.25 & +0.25 & +0.25 & +0.25 & +0.25 & -0.50 & +0.50 & +0.00 & +0.25 & -1.50 \\
 & Qwen2.5-72B-Inst. & -1.23 & +1.02 & +1.27 & -2.23 & +0.52 & +1.77 & -1.48 & -2.23 & +2.52 & +0.10 \\
 & Qwen3-30B-A3B-Inst. & -1.85 & +0.40 & +0.15 & +0.40 & -0.60 & +0.40 & +0.40 & -0.60 & +0.90 & +0.40 \\
 & Qwen3-235B-A22B-Inst. & -4.25 & +0.50 & +1.75 & +0.50 & +0.50 & +2.25 & +0.50 & -3.25 & +1.00 & +0.50 \\
\midrule
\multirow{7}{*}{Right-wing} & gemma-3-27b-it & +1.03 & -1.23 & +2.53 & -1.48 & +2.03 & +2.53 & -0.73 & -2.23 & +0.03 & -2.48 \\
 & gpt-oss-20b & +1.08 & -0.68 & -0.93 & -0.68 & +3.58 & +2.58 & -1.18 & -0.68 & -0.43 & -2.68 \\
 & gpt-oss-120b & -0.16 & -1.91 & -0.91 & +0.09 & +1.09 & +2.09 & +1.34 & -0.66 & +1.59 & -2.58 \\
 & Qwen2.5-7B-Inst. & -0.88 & -1.13 & +2.87 & -0.88 & -0.88 & +3.12 & +1.12 & -1.13 & +0.87 & -3.05 \\
 & Qwen2.5-72B-Inst. & +1.94 & -1.81 & +1.94 & -2.06 & +3.19 & +2.44 & -1.56 & -1.81 & +1.19 & -3.47 \\
 & Qwen3-30B-A3B-Inst. & +0.22 & +0.22 & +0.97 & +0.22 & +2.47 & +0.22 & +0.22 & -0.78 & -1.28 & -2.45 \\
 & Qwen3-235B-A22B-Inst. & -0.08 & -2.33 & +2.17 & -0.58 & +4.67 & +3.17 & -1.08 & -1.08 & -1.08 & -3.75 \\
\midrule
\multirow{7}{*}{Left-wing} & gemma-3-27b-it & -1.00 & +1.00 & -0.25 & -0.25 & -0.25 & -0.50 & +1.00 & +0.00 & -0.75 & +1.00 \\
 & gpt-oss-20b & -1.95 & +1.05 & -1.20 & -0.45 & +0.05 & +0.80 & +0.55 & -0.45 & -0.45 & +2.05 \\
 & gpt-oss-120b & -0.26 & -0.01 & -0.76 & -0.01 & -0.26 & -0.26 & +0.24 & +0.99 & -0.51 & +0.82 \\
 & Qwen2.5-7B-Inst. & -0.57 & +2.68 & -0.57 & -0.57 & -0.57 & -1.07 & +1.18 & -0.82 & -0.57 & +0.92 \\
 & Qwen2.5-72B-Inst. & -1.05 & +1.45 & -0.55 & -0.30 & +0.70 & -1.55 & -0.55 & +0.70 & -0.05 & +1.20 \\
 & Qwen3-30B-A3B-Inst. & -2.88 & +1.38 & -0.13 & +1.38 & -0.88 & -2.88 & +1.38 & +1.38 & -0.13 & +1.38 \\
 & Qwen3-235B-A22B-Inst. & -3.25 & +1.50 & -1.00 & +0.75 & +1.50 & -1.00 & +1.50 & +0.25 & -1.75 & +1.50 \\
\midrule
\multirow{7}{*}{Below university} & gemma-3-27b-it & -0.56 & -0.81 & +1.94 & -0.06 & +0.94 & +1.19 & -0.81 & -1.06 & +0.69 & -1.48 \\
 & gpt-oss-20b & -1.63 & +0.38 & -0.13 & -0.13 & +0.88 & +1.38 & -0.38 & -0.63 & -0.13 & +0.38 \\
 & gpt-oss-120b & -0.13 & -0.63 & +0.37 & -0.13 & -0.13 & -0.13 & -0.13 & +0.37 & +0.62 & -0.05 \\
 & Qwen2.5-7B-Inst. & +0.28 & +0.03 & +0.28 & +0.28 & +0.28 & -0.48 & +0.28 & +0.03 & +0.28 & -1.23 \\
 & Qwen2.5-72B-Inst. & -0.41 & -0.41 & +0.09 & +0.34 & +0.34 & +0.34 & -0.91 & +0.34 & +1.09 & -0.83 \\
 & Qwen3-30B-A3B-Inst. & +0.18 & +0.18 & -0.08 & +0.18 & -0.33 & +0.18 & +0.18 & +0.18 & -0.83 & +0.18 \\
 & Qwen3-235B-A22B-Inst. & -2.70 & +0.30 & -0.20 & +1.05 & +0.30 & +1.05 & -0.20 & +0.30 & -0.20 & +0.30 \\
\midrule
\multirow{7}{*}{University or above} & gemma-3-27b-it & -0.45 & -0.45 & +0.30 & -0.45 & +0.30 & +0.55 & +0.05 & +0.05 & +0.05 & +0.05 \\
 & gpt-oss-20b & +0.38 & +0.13 & -0.12 & -1.12 & +0.88 & +1.38 & -0.62 & -1.12 & -1.12 & +1.30 \\
 & gpt-oss-120b & -0.19 & -0.19 & +0.31 & +0.06 & -0.19 & -0.19 & +0.56 & -0.19 & -0.19 & +0.22 \\
 & Qwen2.5-7B-Inst. & +0.12 & +0.37 & +0.37 & +0.12 & +0.12 & -0.38 & +0.12 & +0.12 & +0.12 & -1.05 \\
 & Qwen2.5-72B-Inst. & +0.58 & +0.08 & -0.18 & -0.18 & -0.18 & -0.18 & -0.18 & -0.18 & +0.58 & -0.18 \\
 & Qwen3-30B-A3B-Inst. & +0.03 & +0.03 & -0.23 & +0.03 & +0.03 & +0.03 & +0.03 & +0.03 & +0.03 & +0.03 \\
 & Qwen3-235B-A22B-Inst. & -0.50 & -0.25 & -0.75 & +0.75 & -0.25 & +0.75 & -0.25 & +0.50 & +0.25 & -0.25 \\
\bottomrule
\end{tabular}}
\caption{Per-model centered-mean deltas ($\Delta$) for PVQ-21: individual model shifts under persona prompting. All cell values are rounded to two decimals using round-half-up.}
\label{tab:app_rq4_per_model_PVQ-21}
\end{table*}

\subsection{Cross-Model Variance}
\label{sec:app_rq4_model_var}

Table~\ref{tab:app_rq4_model_variance} shows the population standard deviation (\texttt{numpy.std} with \texttt{ddof=0}) of PVQ-40 centered-mean deltas across the seven models for each condition. Higher values indicate greater inter-model disagreement. Political conditions exhibit the highest variance, while Gender and Education conditions are more stable.

\begin{table*}[htbp]
\centering
\begin{tabular}{lrrrrrrrrrr}
\toprule
Condition & Ach & Ben & Con & Hed & Pow & Sec & SD & Sti & Tra & Uni \\
\midrule
Male & 0.47 & 0.27 & 0.46 & 0.22 & 0.59 & 0.39 & 0.34 & 0.71 & 0.80 & 0.69 \\
Female & 0.25 & 0.51 & 0.26 & 0.41 & 0.28 & 0.28 & 0.45 & 0.27 & 0.51 & 0.56 \\
20–39 years old & 0.61 & 0.63 & 0.32 & 0.54 & 0.46 & 0.36 & 0.40 & 0.53 & 0.38 & 0.71 \\
80+ years old & 1.20 & 0.61 & 0.81 & 1.00 & 0.52 & 0.51 & 0.55 & 1.09 & 1.04 & 0.77 \\
Right-wing & 1.50 & 0.76 & 0.55 & 1.02 & 1.13 & 0.75 & 0.83 & 0.56 & 0.66 & 0.77 \\
Left-wing & 1.06 & 0.63 & 0.54 & 0.58 & 0.45 & 0.54 & 0.53 & 0.47 & 0.62 & 0.61 \\
Below university & 0.35 & 0.42 & 0.60 & 0.20 & 0.23 & 0.27 & 0.58 & 0.89 & 0.35 & 0.67 \\
University or above & 0.37 & 0.37 & 0.45 & 0.19 & 0.30 & 0.31 & 0.23 & 0.28 & 0.57 & 0.54 \\
\bottomrule
\end{tabular}
\caption{Cross-model standard deviation of PVQ-40 centered-mean deltas for each condition, computed across the seven RQ4 models using the population estimator (\texttt{numpy.std}, \texttt{ddof=0}). Higher values indicate greater disagreement among models about the direction or magnitude of persona-induced shift.}
\label{tab:app_rq4_model_variance}
\end{table*}

\end{document}